\documentclass[11pt, oneside]{article}

\usepackage[utf8]{inputenc}
\usepackage[english]{babel}
\usepackage{geometry}  
\usepackage{graphicx}      
\usepackage{tabularx}      
\usepackage{amssymb}
\usepackage{amsmath}

\usepackage[hyphens]{url}
\usepackage[hidelinks]{hyperref}
\hypersetup{breaklinks=true}
\usepackage[backend=biber,style=chem-acs, doi=true, articletitle=true, isbn=false, date=year, url=true, autopunct=true]{biblatex}
\makeatletter
\renewbibmacro*{textcite}{%
  \iffieldequals{namehash}{\cbx@lasthash}
    {\usebibmacro{cite:comp}}
    {\usebibmacro{cite:dump}%
     \ifbool{cbx:parens}
       {\bibclosebracket\global\boolfalse{cbx:parens}}
       {}%
     \iffirstcitekey
       {}
       {\textcitedelim}%
     \usebibmacro{cite:init}%
     \ifnameundef{labelname}
       {\printfield[citetitle]{labeltitle}}
       {\printnames{labelname}}%
     \addspace
     \ifnumequal{\value{citecount}}{1}
       {\usebibmacro{prenote}}
       {}%
     \mkbibsuperscript{\usebibmacro{cite:comp}}%
     \stepcounter{textcitecount}%
     \savefield{namehash}{\cbx@lasthash}}}
\makeatother

\usepackage[version=4]{mhchem}

\usepackage{csquotes}
\usepackage{cleveref}
\usepackage{siunitx}
\DeclareSIUnit\molar{M}
\usepackage{geometry}
\usepackage{multirow}
\usepackage[dvipsnames,table,xcdraw]{xcolor}
\definecolor{tumblue}{rgb}{0.0980392157,0.2901960784,0.5058823529}

\usepackage{xspace}
\usepackage{caption}
\usepackage{subcaption}
\usepackage{authblk}

\usepackage{marvosym}

\usepackage{sectsty} 

\sectionfont{\color{tumblue}\selectfont\sffamily}
\subsectionfont{\color{tumblue}\selectfont\sffamily}
\subsubsectionfont{\color{tumblue}\selectfont\sffamily}
\paragraphfont{\selectfont\sffamily}
\subparagraphfont{\color{gray}\selectfont\sffamily}

\definecolor{halfgray}{gray}{0.55}

\IfFileExists{./fonts/Heliotrope-Bold.otf}{
    \usepackage{fontspec}
    \setmainfont{Heliotrope}[Extension = .otf, UprightFont = *-Regular, ItalicFont = *-Italic, BoldFont=*-Bold, Path = ./fonts/]
}{
    \usepackage{biolinum}
    
}

\usepackage{lipsum}
\usepackage[labelfont=bf]{caption}
\usepackage[nonumberlist,acronym]{glossaries-extra}

\newglossaryentry{alignmentg}{
  name={alignment},
  description={The process of ensuring that a machine learning model's behavior is consistent with human intentions, values, or task objectives. In the context of \glspl{llm}, alignment often involves \gls{finetuningg} with \glslink{rlhfg}{human feedback} to produce helpful, honest, and harmless outputs. Alignment can also refer to the process of aligning representations or \glspl{embeddingg} across different \glslink{modalityg}{modalities}---such as text and images—--in \glslink{modalityg}{multi-modal} systems, enabling meaningful cross-modal reasoning and retrieval.}
}

\newglossaryentry{agentg}{
  name={agent},
  description={An entity that interacts with an environment by taking actions based on observations to achieve a goal. In \gls{rlg}, agents learn from feedback to optimize behavior over time. In the context of \glspl{llm}, agents can use tools, retrieve external information, and perform multi-step reasoning to complete complex, goal-oriented tasks.}
}

\newglossaryentry{apig}{
  name={application programming interface},
  description={A set of rules and protocols that allow different software systems to communicate with each other. APIs enable developers to access the functionality of external services or applications programmatically, often over a network. For example, the PubChem API \autocite{pubchemAPI} allows programmatic access to chemical compound data, enabling automated retrieval of molecular structures and properties.}
}

\newglossaryentry{benchmarkg}{
  name={benchmark},
  description={A standardized task or dataset used to evaluate and compare the performance of models or methods. Benchmarks help assess progress and identify strengths and weaknesses of different approaches.}
}

\newglossaryentry{bfsg}{
  name={breadth-first search},
  description={A graph traversal algorithm that explores all neighbors of a node before moving to the next level of nodes. It is often used to find the shortest path in unweighted graphs.}
}

\newglossaryentry{bleug}{
  name={bilingual evaluation understudy},
  description={A metric for evaluating the quality of text generated by a model by comparing it to one or more reference texts. BLEU measures \gls{ngramg} overlap between the generated output and the reference. Higher BLEU scores indicate closer matches.}
}

\newglossaryentry{bog}{
  name={Bayesian optimization},
  description={A global optimization strategy for expensive black-box functions. It does not require access to derivatives of the objective function. BO builds a probabilistic surrogate model, typically a \gls{gpg}, to guide the selection of query points by balancing exploration and exploitation.\autocite{frazier2018tutorial}}
}

\newglossaryentry{cnng}{
  name={convolutional neural network},
  description={A type of \gls{nng} designed to process data with a grid-like structure, such as images. CNNs use convolutional layers to automatically learn spatial hierarchies of features from the input.}
}

\newglossaryentry{cotg}{
  name={chain-of-thought}, 
  description={A \gls{promptingg} strategy that involves a series of intermediate natural-language reasoning steps that lead to the final output.\autocite{wei2022chain} CoT encourages an \gls{llm} to explain its reasoning step by step (e.g., by \gls{promptingg} it to \enquote{think step by step}), and it is intended to improve the ability of \glspl{llm} to perform complex reasoning.}
}

\newglossaryentry{compositionalityg}{
  name={compositionality},
  description={When a complex structure can be constructed by combining simpler components, it is said to have the compositional property. In the context of \gls{mlg}, a model or function $f(x,y)$ exhibits compositionality if it can be constructed from $h(x)$ and $k(y)$ as its components: $f(x,y) = g[h(x),k(y)]$, where $h$, $k$, and $g$ are learned functions. This structure allows modularity, parameter sharing, and generalization to novel input combinations by leveraging the learned behavior of the components. In many \gls{dlg} architectures, compositionality is achieved by the hierarchical application of simpler transformations, such as \glspl{fnng}.}
}

\newglossaryentry{comptscalingg}{
  name={computational scaling},
  description={The way in which the computational cost of an algorithm or model increases with problem size (e.g., the number of atoms, data points, or parameters). Understanding scaling behavior is important for assessing feasibility and performance at larger scales. For example, traditional implementations of \gls{dft} scale as $\mathcal{O}(N^3)$ with the number of basis functions, making large systems computationally expensive.}
}

\newglossaryentry{contextwindowg}{
  name={context window},
  description={The span of input \glspl{tokeng} that a model considers at once when generating predictions. In \glspl{llm}, the context window defines how much preceding (and possibly surrounding) text the model can attend to. Larger context windows allow the model to capture longer dependencies but increase computational cost.}
}

\newglossaryentry{damdg}{
  name={dry age-related macular degeneration},
  description={A chronic eye disease that causes gradual loss of central vision due to the thinning of the macula. dAMD is the more common and less severe form of age-related macular degeneration, typically progressing slowly over time.}
}

\newglossaryentry{dataaugmentationg}{
  name={data augmentation},
  description={A technique used to artificially increase the size and diversity of a training dataset by applying transformations or generating variations of the original data. Data augmentation helps improve model generalization and robustness.}
}

\newglossaryentry{datachunktg}{
  name={data chunk},
  description={A contiguous block or segment of data treated as a unit for processing, storage, or transmission. Data chunks are used to divide large datasets into manageable parts---for example, when streaming text or segmenting long sequences. Unlike a \emph{batch}, which typically refers to a set of independent samples processed together in training, a data chunk often preserves sequential or structural continuity within a single sample.}
}

\newglossaryentry{denovodesigng}{
  name={\textit{de novo} design},
  sort={de novo design},
  description={The process of generating novel molecules or materials from scratch, guided by desired properties or objectives rather than by modifying known structures.}
}

\newglossaryentry{densesparseg}{
  name={dense/sparse vector},
  description={Two categories of vector representations used in \gls{mlg} and data processing. \emph{Dense vectors} have most or all elements nonzero and are typically used for learned \glspl{embeddingg}. \emph{Sparse vectors} contain mostly zero values and are common in high-dimensional representations such as \glspl{oheg}. Sparse vectors are more memory-efficient when stored in specialized formats, while dense vectors are preferred for \glslink{nng}{neural computation}, as they contain richer information.}
}

\newglossaryentry{dfsg}{
  name={depth-first search},
  description={A graph traversal algorithm that explores as far as possible along each branch before backtracking. DFS is often used for pathfinding, cycle detection, and analyzing graph structures.}
}

\newglossaryentry{dlg}{
  name={deep learning},
  description={This subfield of \gls{mlg} uses \glspl{nng} with many layers to model complex patterns in data. Deep learning has enabled major advances in image recognition, natural language processing, and molecular modeling.}
}

\newglossaryentry{dorag}{
  name={weight-decomposed low-rank adaptation},
  description={An extension of \gls{lorag} that separates weight adaptation into magnitude and direction components.\autocite{liu2024dora} DoRA achieves efficient \gls{finetuningg} of \glspl{llm} by modifying only the direction of weights while preserving the pre-trained magnitude, improving stability and performance.}
}

\newglossaryentry{dpog}{
  name={direct preference optimization},
  description={A method for \gls{finetuningg} \glspl{llm} based on human preference data without using reinforcement learning. DPO directly optimizes the model (\gls{policyg}) to prefer outputs ranked higher by human annotators, simplifying the \gls{rlhf} pipeline.}
}

\newglossaryentry{dslg}{
  name={domain-specific language},
  description={A programming language tailored to a particular application domain. DSLs offer specialized syntax and abstractions that make it easier to express solutions within that domain, such as chemical synthesis.}
}

\newglossaryentry{eag}{
  name={evolutionary algorithm},
  description={This family of optimization algorithms is inspired by natural selection. EAs evolve a population of candidate solutions over multiple generations using operations such as mutation, crossover, and score-based selection to find high-performing solutions.}
}

\newglossaryentry{ecfpg}{
  name={extended connectivity fingerprint},
  description={A type of molecular fingerprint that represents chemical structures as binary or count vectors based on the local atomic environments. ECFPs are generated by iteratively hashing the neighborhoods of each atom up to a given radius, capturing information about connectivity and substructures. These fingerprints are invariant to atom indexing and are commonly used in \gls{mlg} pipelines for tasks such as virtual screening, molecular similarity, and property modeling. A widely used variant is ECFP4, which uses a radius of 2.}
}

\newglossaryentry{embeddingg}{
  name={embedding},
  description={A representation of discrete or high-dimensional data in a continuous vector space that preserves relevant relationships or structure. Embeddings are commonly used for words, molecules, graphs, and other symbolic data.}
}

\newglossaryentry{erag}{
  name={energy rank alignment},
  description={A method for aligning the training of generative models with energy-based evaluations. ERA encourages the model to assign higher probabilities to lower-energy (more favorable) configurations by matching the model's ranking of samples to their energy scores.}
}

\newglossaryentry{f1g}{
  name={F$_1$ score},
  description={This performance metric for classification tasks is aimed at balancing precision and recall. It is defined as the harmonic mean of precision and recall: $ F_1 = 2 \cdot \frac{\text{precision} \; \times \; \text{recall}}{\text{precision} \; + \; \text{recall}} $. The F$_1$ score ranges from $0$ to $1$. Since the harmonic mean is dominated by the smaller of the two numbers, a high F$_1$ score means that both precision and recall are high, making this metric useful when classes are imbalanced or when both false positives and false negatives are important.}
}

\newglossaryentry{finetuningg}{
  name={fine-tuning},
  description={The process of taking a pre-trained model and continuing its training on a smaller, task-specific dataset to adapt it to a particular application. Fine-tuning updates the model parameters to specialize its behavior while retaining the general knowledge learned during pre-training.}
}

\newglossaryentry{fnng}{
  name={feed-forward neural network},
  description={The simplest class of \glspl{nng} in which information flows in one direction from input to output through a series of connected layers, without cycles or feedback connections.}
}

\newglossaryentry{fusedrepresentationg}{
  name={fused representation},
  description={A joint representation that integrates information from multiple \glslink{modalityg}{modalities} into a single \glslink{latentspaceg}{embedding space}. See \glslink{latentfusiong}{\textbf{latent fusion}}.}
}

\newglossaryentry{gatingg}{
  name={gating mechanism},
  description={A \gls{nng} component that controls the flow of information by modulating one signal using another, typically via element-wise multiplication. After computing a gate vector $g(x)$, the gating mechanism applies it to an input signal $z$ as $z' = g(x) \odot z$, where $\odot$ denotes element-wise multiplication. This allows the model to selectively suppress or pass through different components of $z$ based on the learned gating function $g$. Gating mechanisms are central to architectures like \glspl{lstm}, where they regulate memory updates, retention, and output generation.}
}

\newglossaryentry{gag}{
  name={genetic algorithm},
  description={This subset of \glspl{eag} model candidate solutions as individuals represented by strings of \enquote{genes} (e.g., binary digits or symbols). Unlike broader \glspl{eag}, GAs focus heavily on genetic representations and recombination to explore the search space.}
}

\newglossaryentry{gnng}{
  name={graph neural network},
  description={A type of \gls{nng} architecture designed to operate on graph-structured data. GNNs learn representations by passing and aggregating information between neighboring nodes, making them well-suited for tasks involving chemical structures, such as molecules and crystals.}
}

\newglossaryentry{gpmg}{
  name={general-purpose model},
  description={A model that is designed to generalize across a wide range of tasks and domains with minimal task-specific modifications. General-purpose models are typically pre-trained on vast, diverse datasets using \glslink{ssl}{self-supervised} objectives. They can be efficiently adapted to new tasks via \gls{promptingg} or \gls{finetuningg}. Examples include architectures such as \glspl{llm} and \glspl{vlm}.}
}

\newglossaryentry{gpg}{
  name={Gaussian process},
  description={A nonparametric probabilistic model used to define a distribution over functions. It is commonly used in \gls{bog} and regression to make predictions with uncertainty estimates.}
}

\newglossaryentry{hallucinationg}{
  name={hallucination},
  description={The phenomenon where a model generates output that seems plausible but factually incorrect or unsupported by the input or training data. Hallucinations are common in \glspl{llm} and pose challenges for applications requiring reliability and factual accuracy.}
}

\newglossaryentry{hnswg}{
  name={hierarchical navigable small world},
  description={An efficient algorithm for approximate nearest neighbor search in high-dimensional spaces. It builds a graph-based data structure with multiple layers of navigable small-world graphs, where most nodes can be reached in a few steps through well-connected hubs. This structure allows fast and scalable similarity search. In chemistry, HNSW is often used for rapid retrieval of structurally similar molecules in large compound libraries.}
}

\newglossaryentry{iclg}{
  name={in-context learning},
  description={In this method, \glspl{llm} are adapted to perform new tasks at inference time by conditioning them on examples provided directly in the input prompt, without updating their parameters. The model uses patterns inferred from these examples to predict or generate appropriate outputs for new inputs.}
}

\newglossaryentry{inchig}{
  name={\gls{inchi}},
  sort={inchi},
  description={A textual representation for chemical substances developed by \gls{iupac}, designed to provide a standard and machine-readable representation of molecular structures. InChI encodes information such as connectivity, hydrogen atoms, stereochemistry, and isotopes in a structured sequence of layers. The general format is \texttt{InChI=version/layers}. }
}

\newglossaryentry{inductivebiasg}{
  name={inductive bias},
  description={The set of assumptions a learning algorithm uses to generalize from limited training data to unseen examples. Inductive bias guides which solutions a model is likely to prefer and can arise from model architecture, input representations, or training objectives. \emph{Hard} inductive biases are built into the structure of the model and define the solution space, while \emph{soft} inductive biases, nudging the model to different parts of the solution space, are encouraged by training choices or priors but can be overridden by the data.}
}

\newglossaryentry{inferenceg}{
  name={inference},
  description={In the context of \glspl{llm}, inference is the computational process by which a trained model produces output \glspl{tokeng} given an input sequence. Inference involves executing a forward pass through the model to evaluate the conditional probability distribution over possible next tokens, and then sampling from that distribution using a decoding strategy. Unlike training, inference does not involve gradient computation or parameter updates, and is often optimized for speed and throughput.}
}

\newglossaryentry{irtg}{
  name={item response theory},
  description={This statistical framework was originally developed in psychometrics to model the relationship between a person’s latent ability and their probability of correctly answering test items. In \gls{mlg}, IRT is used to analyze model performance by treating test examples as such items and estimating their difficulty, allowing for more nuanced evaluation than aggregate accuracy.}
}

\newglossaryentry{latentfusiong}{
  name={latent fusion},
  description={A method for integrating information from multiple \glslink{modalityg}{modalities} by combining their representations in a shared \gls{latentspaceg}. Latent fusion enables models to reason jointly over heterogeneous data, such as combining text and molecular structure embeddings for property prediction or generation tasks.}
}

\newglossaryentry{latentspaceg}{
  name={latent space},
  description={An abstract, typically lower-dimensional space where input data is represented after being transformed by a model. Latent spaces often aim to capture meaningful features or structures that are not explicitly present in the original data.}
}

\newglossaryentry{latentstateg}{
  name={latent state},
  description={An internal, unobserved representation maintained by a model to capture relevant information about the input or sequence history. Latent states are used in models like \glspl{rnn} and \glspl{ssm}.}
}

\newglossaryentry{liftg}{
  name={language-interfaced fine-tuning},
  description={A method for fine-tuning models by framing structured tasks such as classification or regression as natural-language \glslink{promptingg}{prompts}.\autocite{dinh2022lift} LIFT enables the use of \glspl{llm} for supervised tasks without modifying the model architecture, leveraging prompt-based interfaces to adapt to diverse formats.}
}

\newglossaryentry{llmg}{
  name={large language model}, 
  description={A \gls{lmg} that has a large number of parameters. There is no agreed-upon rule for the number of parameters that makes a language model large enough to be called a large language model, but this number is usually on the scale of billions. For example, Llama 3, GPT-3, and GPT-4 contain 70B, 175B, and 1.76T parameters, respectively, while Claude 3 Opus is estimated to have 2 trillion parameters. Most current \glspl{llm} are based on the transformer architecture. \glspl{llm} can perform many types of language tasks, such as generating human-like text, understanding context, translation, summarization, and question-answering.}
}

\newglossaryentry{lmg}{
  name={language model}, 
  description={A model that estimates the probability of a \gls{tokeng} or sequence of \glspl{tokeng} occurring in a longer sequence of \glspl{tokeng}. This probability is used to predict the most likely next \gls{tokeng} based on the previous sequence. Language models are trained on large datasets of text, learning the patterns and structures of language to understand, interpret, and generate natural language.}
}

\newglossaryentry{localenvg}{
  name={Local-Env},
  description={A textual representation of crystal structures, inspired by Pauling's rule of parsimony\autocite{pauling_parsimony} and designed to leverage the structural redundancy often found in the local atomic arrangement of crystals. It begins with the crystal's space group, followed by a list of distinct coordination environments, each specified by its Wyckoff label and a corresponding \gls{smilesg} string.}
}

\newglossaryentry{lorag}{
  name={low-rank adaptation}, 
  description={A \gls{peft} technique that freezes the pre-trained model weights and decomposes the update matrix into two lower-rank matrices that contain a reduced number of trainable parameters to be optimized during \gls{finetuningg}. \autocite{hu2022lora}}
}

\newglossaryentry{lossfunctiong}{
  name={loss function},
  description={Optimization processes often try to minimize a loss function, which is a mathematical function that quantifies the difference between a model's prediction and the true target. It guides the optimization process during training by indicating how well the model is performing.}
}

\newglossaryentry{lstmg}{
  name={long short-term memory},
  description={A type of \gls{rnng} architecture designed to capture long-range dependencies in sequential data. LSTMs use \glspl{gatingg} to control the flow of information, making them effective for tasks like language modeling, time-series prediction, and speech recognition.}
}

\newglossaryentry{mcpg}{
  name={model context protocol},
  description={An open standard for wiring AI systems to external tools and data. It defines a simple client–server setup where servers (also called MCP-servers) expose capabilities---like tools and resources---and clients use them in a consistent, model-agnostic way.}
}

\newglossaryentry{mctsg}{
  name={Monte Carlo tree search},
  description={A heuristic search algorithm used for decision-making in sequential environments, particularly in games and planning problems. MCTS incrementally builds a search tree by simulating many random playouts from different states to estimate action values.\autocite{coulom2006mcst} MCTS typically proceeds through four phases: selection, expansion, simulation, and backpropagation. It has been notably used in systems like AlphaGo \autocite{silver2016alphago} for planning in high-dimensional, sparse-reward environments.}
}

\newglossaryentry{mlg}{
  name={machine learning},
  description={This field of computer science is focused on developing algorithms that enable systems to learn patterns from data and make predictions or decisions without being explicitly programmed.}
}

\newglossaryentry{mlipg}{
  name={machine-learning interatomic potential},
  description={A model that approximates the potential energy surface of atomic systems using \gls{mlg} techniques. MLIPs are typically trained on reference data from quantum mechanical calculations such as \gls{dft}, learning to predict both energies and forces. They serve as a data-driven alternative to classical force fields, enabling more accurate and transferable atomistic simulations at a fraction of the cost of ab initio methods.}
}

\newglossaryentry{modalityg}{
  name={modality},
  description={A form of data characterized by a particular sensory or representational channel, such as text, images, audio, or molecular graphs. In \gls{mlg}, handling multiple modalities enables models to integrate and reason across diverse data sources.}
}

\newglossaryentry{modeltemperatureg}{
  name={model temperature},
  description={The term originates from statistical physics, where temperature controls the entropy of a system. In the context of \gls{mlg}, temperature is a parameter used during sampling from a probabilistic model to control the randomness of the output distribution. Lower temperatures sharpen the distribution, making outputs more deterministic, while higher temperatures flatten it, increasing diversity.  In \glspl{llm}, temperature is commonly tuned during text generation to balance creativity and coherence.}
}

\newglossaryentry{moeg}{
  name={mixture of experts},
  description={A general modeling framework in which multiple specialized submodels, or \emph{experts}, are combined using a function that selects or weights their contributions based on the input.}
}

\newglossaryentry{mpnng}{
  name={message-passing neural network},
  description={A class of \glspl{gnng} that operate on graphs by iteratively updating node representations through the exchange of messages with neighboring nodes. MPNNs are widely used in molecular property prediction and other graph-structured tasks.}
}

\newglossaryentry{ngramg}{
  name={n-gram},
  description={A contiguous sequence of \(n\) \glspl{tokeng} from a given text. N-grams are commonly used in language modeling and text analysis.}
}

\newglossaryentry{nlpg}{
  name={natural language processing}, 
  description={A subfield of computer science that uses \gls{mlg} to enable computers to process and generate human language. The primary tasks in NLP include speech recognition, text classification, natural language understanding, and natural language generation.}
}

\newglossaryentry{nng}{
  name={neural network},
  description={One of the most prevalent model components used in \gls{dlg}, inspired by the structure of the brain. Neural networks are made up of layers of connected units called neurons. Each layer of neurons takes a vector $x$ as input, performs a simple calculation $f(x)$, and passes the result to the next layer. A typical layer can be mathematically represented as $f(x) = \sigma(Wx + b)$, where $W$ is a matrix of learned weights, $b$ is a learned bias term, and $\sigma$ is a non-linear function called the activation function. By stacking many such layers, neural networks can learn to approximate complex functions. See also: \textbf{\gls{compositionalityg}}.}
}

\newglossaryentry{ocrg}{
  name={optical character recognition}, 
  description={A technique used to identify and convert images of printed or handwritten text into a machine-readable text format. This involves segmentation of text regions, character recognition, and post-processing to correct errors and enhance accuracy.}
}

\newglossaryentry{oheg}{
  name={one-hot encoding},
  description={A method for representing categorical variables as binary vectors, where each category is assigned a unique position set to 1, and all others are 0. One-hot encoding is commonly used to input discrete features into machine learning models that require numerical input.}
}

\newglossaryentry{osg}{
  name={operating system},
  description={Software that manages computer resources, providing an interface between hardware and software. The operating system handles tasks such as memory management, process scheduling, and device control.}
}

\newglossaryentry{pddlg}{
  name={planning domain definition Language},
  description={A formal language used to specify planning problems in automated planning. PDDL defines the initial state, goal conditions, and available actions with their preconditions and effects, enabling planners to generate sequences of actions to achieve the desired outcomes. It has been applied to domains such as robotic control and retrosynthetic planning.}
}

\newglossaryentry{peftg}{
  name={parameter-efficient fine-tuning}, 
  description={A methodology to efficiently \glslink{finetuningg}{fine-tune} large pre-trained models without modifying their original parameters. PEFT strategies involve adjusting only a small number of additional model parameters during \gls{finetuningg} on a new, smaller training dataset. This significantly reduces the computational and storage costs while achieving comparable performance to full \gls{finetuningg}. Common PEFT methods include \gls{lora}, \gls{qlora}, and \gls{dora}.}
}

\newglossaryentry{policyg}{
  name={policy},
  description={In the context of \gls{rlg}, a policy defines the agent's behavior by mapping states to actions. It is typically denoted as $\pi(a \mid s)$, representing the conditional probability of taking action $a$ given state $s$. A policy can be deterministic or stochastic, and may be represented by a parameterized function such as a \gls{nng}. The objective of \gls{rl} is to learn a policy that maximizes expected cumulative reward.}
}

\newglossaryentry{ppog}{
  name={proximal policy optimization}, 
  description={A \gls{rl} algorithm used to train \glspl{llm} for alignment. PPO achieves this by adjusting the \gls{policyg} parameters in a way that keeps changes within a predefined safe range to maintain stability and improve learning efficiency. PPO is often used as part of \gls{rlhf}.}
}

\newglossaryentry{promptingg}{
  name={prompting},
  description={The practice of providing input to a \gls{llm} in the form of natural-language instructions, questions, or examples to guide its behavior. Prompting can influence the model’s responses without changing its parameters.}
}

\newglossaryentry{qsprg}{
  name={quantitative structure–property relationship},
description={statistical or machine‑learning approaches that predict a compound’s physicochemical, biological, or environmental properties from its molecular structure by correlating structural descriptors with experimentally measured values.}
}

\newglossaryentry{ragg}{
  name={retrieval augmented generation}, 
  description={A technique for improving the quality of text generation by providing \glspl{llm} with access to information retrieved from external knowledge sources. In practice, this means that relevant retrieved text snippets are added to the prompt.}
}

\newglossaryentry{regexg}{
  name={regular expression},
  description={Often abbreviated as \emph{regex}, it is a sequence of characters that defines a search pattern for matching text. Regular expressions are widely used for tasks such as string validation, extraction, and substitution.}
}

\newglossaryentry{rlg}{
  name={reinforcement learning},
  description={In this \gls{mlg} paradigm, an \gls{agentg} learns to take actions in an environment to maximize cumulative reward through trial and error. See also: \textbf{\gls{policyg}}}.
}

\newglossaryentry{rlhfg}{
  name={reinforcement learning from human feedback}, 
  description={This mechanism uses \gls{rl} to align \glspl{llm} with user preferences by \gls{finetuningg} on human feedback. Users are asked to rate the quality of a model's response. Based on this, a preference model is trained and then used in a reinforcement learning setup to optimize the generations of the \gls{llm}.}
}

\newglossaryentry{rnng}{
  name={recurrent neural network},
  description={A type of \gls{nng} designed for processing sequential data by maintaining a \glslink{latentstateg}{hidden state} that captures information from previous time steps. Without special mechanisms, RNNs are limited in capturing long-range dependencies.}
}

\newglossaryentry{rocaucg}{
  name={receiver operating characteristic---area under the curve},
  description={This performance metric is used for binary classifiers and measures the area under the ROC curve, which plots the true positive rate against the false positive rate at various thresholds. A higher ROC-AUC indicates better ability to distinguish between classes, with 1.0 being perfect and 0.5 representing random guessing.}
}

\newglossaryentry{semanticsearchg}{
  name={semantic search},
  description={Rather than retrieving text based on exact keyword matching, semantic search is a technique that retrieves information based on meaning. Semantic search uses vector representations (see \glslink{embeddingg} {\textbf{embedding}}) of queries and documents to capture contextual similarity, enabling more accurate retrieval of relevant results even when different words are used.}
}

\newglossaryentry{selfiesg}{
  name={\gls{selfies}},
  sort={selfies},
  description={A textual representation of molecular structures designed to be 100\% robust, meaning every SELFIES string maps to a valid molecule. Unlike \gls{smilesg}, SELFIES uses a grammar-based system to encode molecular structures in a way that prevents syntactic invalidity, making it especially useful for generative models.}
}

\newglossaryentry{slicesg}{
  name={\gls{slices}},
  sort={slices},
  description={A string-based representation of crystal structures in a human-readable and machine-interpretable form. SLICES aims to facilitate generative modeling in materials science by representing crystalline structures as linear sequences amenable to sequence-based models.}
}

\newglossaryentry{sftg}{
  name={supervised fine-tuning}, 
  description={One of the simplest forms of the \gls{finetuningg} process, in which a pre-trained \gls{llm} is \glslink{finetuningg}{fine-tuned} on a smaller, labeled dataset for a specific task.}
}

\newglossaryentry{smilesg}{
  name={\gls{smiles}},
  sort={SMILES},
  description={A non-unique textual representation of molecular structures, often small organic molecules, using a sequence of ASCII characters. SMILES encodes atoms and bonds in a linear form suitable for storage, search, and \gls{mlg} applications. The term \emph{canonical} smiles refers to a smiles string that is generated using a deterministic algorithm (e.g., by \modelname{RDKit}) to ensure consistency.}
}

\newglossaryentry{sotag}{
  name={state-of-the-art}, 
  description={In the context of \gls{mlg}, this term is used to describe the most advanced models or techniques that represent the highest level of performance achievable today. They are typically the result of extensive research and development and serve as benchmarks for researchers and developers.}
}

\newglossaryentry{statespaceg}{
  name={state space},
  description={The set of all possible internal configurations (states) a system or model can occupy. In \gls{mlg}, the state space defines how the system evolves over time and is used to model sequential behavior.}
}

\newglossaryentry{supervisedg}{
  name={supervised learning},
  description={In this \gls{mlg} paradigm, the model is trained on labeled data, learning to map inputs to known outputs.}
}

\newglossaryentry{sslg}{
  name={self-supervised learning}, 
  description={A \gls{mlg} technique that involves generating labels from the input data itself instead of relying on external labeled data. It has been foundational for the success of \glspl{llm}, as their pre-training task (next word prediction or filling in of masked words) is a self-supervised task.}
}

\newglossaryentry{ssmg}{
  name={selective state space model},
  description={A \glslink{nng}{neural architecture} that models sequential data using \glspl{latentstateg} and learned transitions, while \glslink{gatingg}{selectively controlling} which components of the state are updated at each step. SSMs aim to improve long-range reasoning and efficiency, and are used as alternatives to attention mechanisms in tasks like language modeling.}
}

\newglossaryentry{tokeng}{
  name={token},
  description={A basic unit of text used by \glspl{lmg}, typically corresponding to a word, subword, or character depending on the tokenizer. Models process and generate text as sequences of tokens.}
}

\newglossaryentry{transferabilityg}{
  name={transferability},
  description={The degree to which knowledge learned in one setting---such as a task, domain, or data distribution---can be effectively reused or adapted in another. High transferability indicates that representations or models generalize well to new contexts, and it is a key objective in \gls{transferlearningg}, foundation models, and general-purpose architectures.}
}

\newglossaryentry{transferlearningg}{
  name={transfer learning},
  description={In this \gls{mlg} paradigm, knowledge gained from one task or domain is reused to improve performance on a different, often related, task. Transfer learning typically involves pre-training a model on a large dataset and then fine-tuning it on a smaller, task-specific dataset, making it particularly valuable in low-data settings.}
}

\newglossaryentry{totg}{
  name={tree-of-thought},
  description={This \gls{promptingg} and reasoning framework extends \gls{cotg} by exploring multiple reasoning paths in a tree structure. ToT allows a model to evaluate and revise intermediate steps, enabling more deliberate decision-making in complex tasks.}
}

\newglossaryentry{unsupervisedg}{
  name={unsupervised learning},
  description={In this \gls{mlg} paradigm, the model is trained on unlabeled data to discover hidden patterns or structure, such as clusters or latent variables.}
}

\newglossaryentry{vlmg}{
  name={vision language model},
  description={A \glslink{modalityg}{multi-modal} model that can simultaneously learn from images and texts, generating text outputs.}
}
\makenoidxglossaries 

\setabbreviationstyle[acronym]{long-short}
\newacronym{acl}{ACL}{association for computational linguistics}
\newacronym{ai}{AI}{artificial intelligence}
\newacronym{afm}{AFM}{atomic force microscopy}
\newacronym{agi}{AGI}{artificial general intelligence}
\newacronym[long=application programming interface]{api}{API}{\gls{apig}}
\newacronym{ar}{AR}{augmented reality}
\newacronym[long=bilingual evaluation understudy]{bleu}{BLEU}{\gls{bleug}}
\newacronym[long=breadth-first search]{bfs}{BFS}{\gls{bfsg}}
\newacronym[long=Bayesian optimization]{bo}{BO}{\gls{bog}}
\newacronym{camp}{CAMP}{context-aware molecule prediction}
\newacronym{chasm}{ChASM}{chemical assembly language}
\newacronym{cas}{CAS}{chemical abstracts service}
\newacronym{casp}{CASP}{computer-aided synthesis planning}
\newacronym{caspr}{CASP}{critical assessment of techniques for protein structure prediction}
\newacronym{cern}{CERN}{The European Organization for Nuclear Research}
\newacronym[long=chain-of-thought]{cot}{CoT}{\gls{cotg}}
\newacronym[long=convolutional neural network]{cnn}{CNN}{\gls{cnng}}
\newacronym{chidl}{$\chi$DL}{chemical description language}
\newacronym{cif}{CIF}{crystallographic information file}
\newacronym{cpu}{CPU}{central processing unit}
\newacronym{deet}{DEET}{N,N-diethyl-meta-toluamide}
\newacronym[long=depth-first search]{dfs}{DFS}{\gls{dfsg}}
\newacronym{dft}{DFT}{density functional theory}
\newacronym[long=deep learning]{dl}{DL}{\gls{dlg}}
\newacronym[long=direct preference optimization]{dpo}{DPO}{\gls{dpog}}
\newacronym[long=dry age-related macular degeneration]{damd}{dAMD}{\gls{damdg}}
\newacronym[long=weight-decomposed low-rank adaptation]{dora}{DoRA}{\gls{dorag}}
\newacronym[long=domain-specific language]{dsl}{DSL}{\gls{dslg}}
\newacronym[long=evolutionary algorithm]{ea}{EA}{\gls{eag}}
\newacronym[long=extended connectivity fingerprint]{ecfp}{ECFP}{\gls{ecfpg}}
\newacronym[long=energy ranking alignment]{era}{ERA}{\gls{erag}}
\newacronym{eu}{EU}{European Union}
\newacronym{fm}{FM}{foundation model}
\newacronym[long=feed-forward neural network]{fnn}{FNN}{\gls{fnng}}
\newacronym[long=genetic algorithm]{ga}{GA}{\gls{gag}}
\newacronym[long=graph neural network]{gnn}{GNN}{\gls{gnng}}
\newacronym{go}{GO}{gene ontology}
\newacronym[long=Gaussian process]{gp}{GP}{\gls{gpg}}
\newacronym[long=general-purpose model]{gpm}{GPM}{\gls{gpmg}}
\newacronym{gpr}{GPR}{Gaussian process regression}
\newacronym{grpo}{GRPO}{group-relative policy optimization}
\newacronym{homo}{HOMO}{highest occupied molecular orbital}
\newacronym[long=hierarchical navigable small world]{hnsw}{HNSW}{\gls{hnswg}}
\newacronym{iaio}{IAIO}{International Artificial Intelligence Oversight organization}
\newacronym[long=in-context learning]{icl}{ICL}{\gls{iclg}}
\newacronym{iclr}{ICLR}{International Conference on Learning Representations}
\newacronym{icma}{ICMA}{in-context molecule adaptation}
\newacronym[long=international chemical identifier]{inchi}{InChI}{international chemical identifier (See glossary: \gls{inchig})}
\newacronym{ip}{IP}{intellectual property}
\newacronym{ir}{IR}{infrared spectroscopy}
\newacronym[long=item response theory]{irt}{IRT}{\gls{irtg}}
\newacronym{iupac}{IUPAC}{International Union of Pure and Applied Chemistry}
\newacronym[long=language-interfaced finetuning]{lift}{LIFT}{\gls{liftg}}
\newacronym[long=large language model]{llm}{LLM}{\gls{llmg}}
\newacronym[long=low-rank adaptation]{lora}{LoRA}{\gls{lorag}}
\newacronym{lhasa}{LHASA}{logic and heuristics applied to synthetic analysis}
\newacronym{lime}{LIME}{local interpretable model-agnostic explanations}
\newacronym[long=language model]{lm}{LM}{\gls{lmg}}
\newacronym[long=local-environment]{localenv}{Local-Env}{\glslink{localenvg}{local-environment}}
\newacronym[long=long short-term memory]{lstm}{LSTM}{\gls{lstmg}}
\newacronym{lumo}{LUMO}{lowest unoccupied molecular orbital}
\newacronym{json}{JSON}{JavaScript object notation}
\newacronym{mae}{MAE}{mean absolute error}
\newacronym[long=model context protocol]{mcp}{MCP}{\gls{mcpg}}
\newacronym{mcq}{MCQ}{multiple-choice question}
\newacronym[long=Monte Carlo tree search]{mcts}{MCTS}{\gls{mctsg}}
\newacronym{md}{MD}{molecular dynamics}
\newacronym[long=machine learning]{ml}{ML}{\gls{mlg}}
\newacronym[long=machine-learning interatomic potential]{mlip}{MLIP}{\gls{mlipg}}
\newacronym{mlm}{MLM}{masked language model}
\newacronym[long=mixture of experts]{moe}{MoE}{\gls{moeg}}
\newacronym{mof}{MOF}{metal-organic framework}
\newacronym[long=message-passing neural network]{mpnn}{MPNN}{\gls{mpnng}}
\newacronym{ms}{MS}{mass spectrometry}
\newacronym{neurips}{NeurIPS}{Conference on Neural Information Processing Systems}
\newacronym{nih}{NIH}{National Institutes of Health}
\newacronym[long=natural-language processing]{nlp}{NLP}{\gls{nlpg}}
\newacronym{nmr}{NMR}{nuclear magnetic resonance}
\newacronym[long=One-hot encoding]{ohe}{OHE}{\gls{oheg}}
\newacronym[long=optical character recognition]{ocr}{OCR}{\gls{ocrg}}
\newacronym{ord}{ORD}{Open Reaction Database}
\newacronym[long=operating system]{os}{OS}{\gls{osg}}
\newacronym{pddl}{PDDL}{planning domain definition language}
\newacronym{pflop}{PFLOP}{peta floating point operations per second}
\newacronym{pmo}{PMO}{practical molecular optimization}
\newacronym[long=proximal policy optimization]{ppo}{PPO}{\gls{ppog}}
\newacronym[long=parameter-efficient fine-tuning]{peft}{PEFT}{\gls{peftg}}
\newacronym{petn}{PETN}{pentaerythritol tetranitrate}
\newacronym{qa}{Q\&A}{questions \& answers}
\newacronym{qlora}{QLoRA}{quantized low-rank adaptation}
\newacronym{qmmm}{QM/MM}{quantum mechanics/molecular mechanics}
\newacronym[long=quantitative structure-property relationships]{qspr}{QSPR}{\gls{qsprg}}
\newacronym[long=retrieval-augmented generation]{rag}{RAG}{\gls{ragg}}
\newacronym{rap}{RAP}{reasoning via planning}
\newacronym{react}{ReAct}{reasoning and acting}
\newacronym{rfm}{RFM}{Riemannian flow-matching}
\newacronym[long=reinforcement learning]{rl}{RL}{\gls{rlg}}
\newacronym[long=reinforcement learning from human feedback]{rlhf}{RLHF}{\gls{rlhfg}}
\newacronym{rmse}{RMSE}{root mean squared error}
\newacronym{rna}{RNA}{ribonucleic acid}
\newacronym[long=recurrent neural network]{rnn}{RNN}{\gls{rnng}}
\newacronym[long=receiver operating characteristic --- area under the curve]{roc_auc}{ROC-AUC}{\gls{rocaucg}}
\newacronym{sdl}{SDL}{self-driving lab}
\newacronym[long=self-referencing embedded strings]{selfies}{SELFIES}{self-referencing embedded strings (See glossary: \gls{selfiesg})}
\newacronym{shap}{SHAP}{shapley additive explanations}
\newacronym[long=supervised fine-tuning]{sft}{SFT}{\gls{sftg}}
\newacronym[long=simplified line-input crystal-encoding system]{slices}{SLICES}{simplified line-input crystal-encoding system (See glossary: \gls{slicesg})}
\newacronym{sm}{SM}{statistical mechanics}
\newacronym[long=simplified molecular input line entry system]{smiles}{SMILES}{simplified molecular input line entry system (See glossary: \gls{smilesg})}
\newacronym[long=state-of-the-art]{sota}{SOTA}{\gls{sotag}}
\newacronym[long=self-supervised learning]{ssl}{SSL}{\gls{sslg}}
\newacronym[long=selective state space model]{ssm}{SSM}{\gls{ssmg}}
\newacronym{tgm-dlm}{TGM-DLM}{text-guided molecule generation with diffusion language modeling}
\newacronym{tmc}{TMC}{transition metal complexes}
\newacronym[long=tree-of-thought]{tot}{ToT}{\gls{totg}}
\newacronym{uk}{UK}{United Kingdom}
\newacronym{uspto}{USPTO}{United States Patent and Trademark Office}
\newacronym{uv-nir}{UV-NIR}{ultra-violet near-infrared}
\newacronym{uv-vis}{UV-Vis}{ultra-violet visible}
\newacronym[long=vision language model]{vlm}{VLM}{\gls{vlmg}}
\newacronym{viads}{VIADS}{visual interactive analytic tool for filtering and summarizing large health data sets}
\newacronym{xps}{XPS}{X-ray photoelectron spectroscopy}
\newacronym{xrd}{XRD}{X-ray diffraction}
\newacronym{xlstm}{xLSTM}{extended long short-term memory}

\makenoidxglossaries

 \usepackage{microtype}
 \usepackage{fontawesome}

\usepackage{tikz}
\usepackage{annotate-equations}

\definecolor{PositiveColor}{RGB}{25, 74, 129}   
\definecolor{NegativeColor}{RGB}{102, 17, 36}   
\definecolor{TemperatureColor}{RGB}{255, 140, 0}   
\definecolor{SimilarityColor}{RGB}{70, 130, 180}   
\definecolor{OverallColor}{RGB}{75, 0, 130}   

\definecolor{StateColor}{RGB}{25, 74, 129} 
\definecolor{ActionColor}{RGB}{102, 17, 36 } 
\definecolor{PolicyColor}{RGB}{179, 28, 43} 
\definecolor{RewardColor}{RGB}{244, 165, 130} 

\definecolor{promptblue}{HTML}{194a81}
\definecolor{promptlightblue}{HTML}{92c4de}
\newcommand{\modelname}[1]{\texttt{#1}}

\usepackage[most]{tcolorbox}
\tcbuselibrary{listings, breakable}
\newcolumntype{Y}{>{\centering\arraybackslash}X}
\newtcolorbox[auto counter]{promptbox}[2][]{
    title = {LLM Prompt Example~\thetcbcounter: #2},
    breakable,
    enhanced,
    colback=white,
    colframe=promptblue,
    colbacktitle=promptblue,
    fonttitle=\bfseries,
    boxrule=0.5mm,
    left=2mm,
    right=2mm,
    top=4mm,
    bottom=4mm,
    middle=10mm,
    #1
}
\crefname{tcb@cnt@promptbox}{prompt example}{prompt examples}
\Crefname{tcb@cnt@promptbox}{Prompt Example}{Prompt Examples}

\definecolor{examplered}{RGB}{100, 127, 188}

\usepackage{float}
\newfloat{example}{htbp}{loe}
\floatname{example}{Example}
\newtcolorbox[auto counter]{examplebox}[2][]{
    title = {Example~\thetcbcounter: #2},
    breakable,
    enhanced,
    colback=white,
    colframe=examplered,
    colbacktitle=examplered,
    fonttitle=\bfseries,
    boxrule=0.5mm,
    left=2mm,
    right=2mm,
    top=4mm,
    bottom=4mm,
    middle=10mm,
    #1
}
\crefname{tcb@cnt@examplebox}{example}{examples}
\Crefname{tcb@cnt@examplebox}{Example}{Examples}
\crefname{example}{example}{examples}
\Crefname{example}{Example}{Examples}

\usepackage{credits}
\usepackage{orcidlink}
\usepackage{hyperref}
\usepackage{multirow}
\usepackage{threeparttable}
\usepackage{arydshln}
\usepackage{xcolor}
\usepackage{booktabs}
\usepackage{tcolorbox}
\usepackage[version=4]{mhchem}
\usepackage{siunitx}
\usepackage{float}

\usepackage{array}
\usepackage{longtable}
\usepackage{booktabs}
\usepackage{xltabular}
\usepackage{makecell}
\usepackage{longtable}

\newcommand{\smi}[1]{\texttt{#1}}

\newcommand{\cellimage}[1]{%
  \raisebox{-0.9\height}{%
    \includegraphics[width=0.30\textwidth,
                     keepaspectratio]{#1}}}

\title{\textsf{General-Purpose Models for the Chemical Sciences: LLMs and Beyond}}

\author[1,$\star$]{Nawaf~Alampara~\orcidlink{0009-0001-7697-7315}}
\author[1,$\star$]{Anagha~Aneesh~\orcidlink{0009-0001-0275-2586}}
\author[1,$\star$]{Martiño~Ríos-García~\orcidlink{0000-0003-1507-4048}}
\author[2,3,$\star$]{Adrian~Mirza~\orcidlink{0000-0003-4033-4235}}
\author[1,$\star$]{Mara~Schilling-Wilhelmi~\orcidlink{0009-0007-4392-5918}}

\author[1]{Ali~Asghar~Aghajani~\orcidlink{0000-0002-1318-9311}}

\author[1,$\dagger$]{Meiling~Sun~\orcidlink{0000-0001-8124-6579}}
\author[2,3,$\dagger$]{Gordan~Prastalo~\orcidlink{0009-0004-5505-568X}}

\author[1,2,4,5 \Letter]{Kevin~Maik~Jablonka~\orcidlink{0000-0003-4894-4660}}

\affil[1]{Laboratory of Organic and Macromolecular Chemistry (IOMC), Friedrich Schiller University Jena, Humboldtstrasse 10, 07743 Jena, Germany}
\affil[2]{HIPOLE Jena (Helmholtz Institute for Polymers in Energy Applications Jena), Lessingstrasse 12-14, 07743 Jena, Germany}

\affil[3]{Helmholtz-Zentrum Berlin für Materialien und Energie GmbH, Hahn-Meitner-Platz 1, 14109
Berlin, Germany}

\affil[4]{Center for Energy and Environmental Chemistry Jena (CEEC Jena), Friedrich Schiller University Jena, Philosophenweg 7a, 07743 Jena, Germany}

\affil[5]{Jena Center for Soft Matter (JCSM), Friedrich Schiller University Jena, Philosophenweg 7, 07743 Jena, Germany}

\affil[$\star$]{These authors had an equivalent impact on this work. Name order was decided at random, and the first position is interchangeable, reflecting their shared contribution.}
\affil[$\dagger$]{These authors had an equivalent impact on this work. Name order was decided at random, and the first position is interchangeable, reflecting their shared contribution.}

\affil[\Letter]{\texttt{mail@kjablonka.com}}

\begin{document}

\maketitle

\begin{abstract}
    \noindent Data-driven techniques have a large potential to transform and accelerate the chemical sciences. 
    However, chemical sciences also pose the unique challenge of very diverse, small, fuzzy datasets that are difficult to leverage in conventional machine learning approaches. 
    A new class of models, which can be summarized under the term general-purpose models (GPMs) such as large language models, has shown the ability to solve tasks they have not been directly trained on, and to flexibly operate with low amounts of data in different formats.
    In this review, we discuss fundamental building principles of GPMs and review recent and emerging applications of those models in the chemical sciences across the entire scientific process.
    While many of these applications are still in the prototype phase, we expect that the increasing interest in GPMs will make many of them mature in the coming years. 
\end{abstract}
\pagebreak
\tableofcontents

\clearpage 
\section{Introduction}

\Gls{ml} shows promise to accelerate the rate of scientific progress.\autocite{jablonka2020big,Butler_2018,yano2022case,Yao_2022,De_Luna_2017,wang2023scientific} 
Recent progress in the field has demonstrated, for example, the ability of \gls{ml} models to make predictions for multiscale systems,\autocite{charalambous2024holistic,yang2020machine, Deringer_2021} to perform experiments by interacting with laboratory equipment \autocite{boiko2023autonomous,coley2019robotic}, to autonomously collect data from scientific literature,\autocite{schilling2025text,zhang2024fine,dagdelen2024structured} and to make predictions with high accuracy.\autocite{jablonka2024leveraging,jablonka2023machine,jung2024automatic, Rupp_2012,Keith_2021,Wu_2024} 

However, the diversity and scale of chemical data create a unique challenge for applying \gls{ml} to the chemical sciences. 
This diversity manifests across temporal, spatial, and representational dimensions. Temporally, chemical processes span femtosecond-scale spectroscopic events to year-long stability studies of pharmaceuticals or batteries, demanding data sampled at resolutions tailored to each time regime. 
Spatially, systems range from the atomic to the industrial scale, requiring models that bridge molecular behavior to macroscopic properties. 
Representationally, even a single observation (e.g., a \ce{^{13}C}-NMR spectrum) can be encoded in chemically equivalent formats: a string \autocite{alberts2024unraveling}, vector \autocite{mirza2024elucidating}, or image\autocite{alberts2024unraveling}. 
However, such representations are not computationally equivalent and have been empirically shown to produce different model outputs.\autocite{atz2024prospective,alampara2024probing,wu2024t,skinnider2024invalid}

Additionally, \gls{ml} for chemistry is challenged by what one can term \enquote{hidden variables}. 
These can be thought of as the parameters in an experiment that remain largely unaccounted for (e.g., their importance is unknown, or they are difficult to control for), but could have a significant impact on experimental outcomes. 
One example is seasonal variations in ambient laboratory conditions that are typically not controlled for and, if at all, only communicated in private accounts.\autocite{Nega_2021}
In addition to that, chemistry is believed to rely on a large amount of \emph{tacit knowledge}, i.e., knowledge that cannot be readily verbalized.\autocite{Taber_2014, Polanyi_2009}  
Tacit chemical knowledge includes the subtle nuances of experimental procedures, troubleshooting techniques, and the ability to anticipate potential problems based on experience.

These factors---the diversity, scale, and tacity---clearly indicate that the full complexity of chemistry cannot be captured using standard approaches with bespoke representations based on structured data.\autocite{jablonka2022making}
Fully addressing the challenges imposed by chemistry requires the development of \gls{ml} systems that can handle diverse, \enquote{fuzzy}, data instances and have transferable capabilities to leverage low amounts of data. \\

\enquote{Foundation model} has become a popular term for large pretrained models that serve as a basis for various downstream tasks. The first comprehensive description of such models was provided by \textcite{bommasani2021opportunities}, who also coined the term \enquote{foundation models}. 
In the chemical literature, this term has different connotations. In many cases, however, the term is used to represent a domain‑specific, state‑of‑the‑art model limited to one input modality (e.g., amino acid sequences, crystal structures). 
Here, we make the distinction between what we term \glspl{gpm}, such as \glspl{llm} \autocite{zhang2024chemllm, guo2025deepseek, openai2023gpt04, anthropic2025system, brown2020language} and domain-specific models with \gls{sota} performance in a subset of tasks, such as machine-learning interatomic potentials.\autocite{batatia2023foundation,Chen_2022,unke2021machine} We adopt the term \glspl{gpm} to avoid the semantic overlap caused by \enquote{foundation model} and to signal the breadth of applicability that we seek to emphasize.

A \gls{gpm} is a model that has been pre‑trained on a broad, heterogeneous corpus spanning multiple data modalities (text, images, graphs) or representations (e.g., common names, 3D coordinates, molecular images).
It can be applied to a wide spectrum of downstream tasks that differ in objective (classification, regression, generation, reasoning), input format, and domain---ranging from natural‑language processing to chemistry and vision---with little or no task‑specific fine‑tuning.
A \gls{gpm} supports zero‑shot, few‑shot, or transfer learning and can serve as the core component of autonomous agents.

\begin{table}[!h]
    \centering
    \caption{\textbf{Illustrative examples of GPMs, domain‑specific foundation models, and specialized chemistry ML pipelines}. The table depicts the definition of \gls{gpm} with examples for such models, as well as comparisons with domain-specific models and chemistry pipelines. Note that a \gls{gpm} does not necessarily output text.}
    \label{tab:gpm}
    \begin{tabularx}{\linewidth}{@{}p{0.22\linewidth} p{0.42\linewidth} X@{}}
        \toprule
        \textbf{Category} & \textbf{Typical Characteristics} & \textbf{Representative Examples} \\
        \midrule
        \addlinespace[0.3em]
        
        \glspl{gpm} &
        Pre‑trained on a large heterogeneous corpus spanning multiple modalities. Supports zero/few‑shot generalization and can be fine‑tuned for diverse chemistry tasks. Capable of autonomous agent behavior, including planning and execution. &
        \textbf{Autoregressive:} GPT‑4,\autocite{openai2023gpt04} LLaMA,\autocite{grattafiori2024llama} Galactica\autocite{taylor2022galactica}
        
        \textbf{Diffusion-based:} Gemini Diffusion,\autocite{deepmind_geminidiffusion_2025} Inception Mercury\autocite{inceptionlabs2025mercury0}
        
        \textbf{Other:} Mamba-based\autocite{gu2023mamba0} models \\
        
        \addlinespace[0.8em]
        \midrule
        \addlinespace[0.3em]
        
        Domain‑Specific Foundation Models &
        Trained on curated, domain‑specific datasets (e.g., protein structures, crystal structures). Achieve state‑of‑the‑art performance in narrow task sets, but are typically not multimodal or generalizable to unrelated chemistry problems. &
        AlphaFold,\autocite{AlphaFold2021} ESM,\autocite{lin2022language} MACE-MP-0\autocite{batatia2023foundation}, MatterSim\autocite{yang2024mattersim}, MolecularTransformer\autocite{schwaller2019molecular} \\
        
        \addlinespace[0.8em]
        \midrule
        \addlinespace[0.3em]
        
        Specialized Chemistry Pipelines &
        Domain‑specific models combined with rule‑based or symbolic components. Often rely on hand‑crafted descriptors with limited transferability beyond the target task. &
        Graph‑based reaction outcome predictors,  \gls{qspr} models using Morgan fingerprints, \gls{gpr} on \gls{nmr} shifts \\
        
        \addlinespace[0.3em]
        \bottomrule
    \end{tabularx}
\end{table}

\Cref{tab:gpm} gives examples of how this definition can be applied. 
By decoupling the notion of \enquote{general‑purpose} from any specific architecture or modality, we aim to foster creative exploration of models that are better aligned with the data characteristics and scientific goals of chemistry.
We hope to contribute to this by addressing chemists and computer scientists, by providing technical background, a consistent terminology, and explaining key technical terminology in a \hyperref[glossary]{glossary} at the end of the manuscript.

\section{The Shape and Structure of Chemical Data} \label{sec:data-section}

\subsection{Shape of Scientific Data}
To understand the successes and failures of \gls{ml} models, it is instructive to explore how the structure of different datasets shapes the learning capabilities of models. 
One useful lens for doing so is to consider how complex a system is (i.e., how many variables are needed to describe it) and what fraction of these variables are explicit. 
One might see the set of variables required to describe a system as the state space. 
A state space encompasses all possible states of a system, similar to concepts in \gls{sm}. 

However, in contrast to many other problems, we often cannot explicitly enumerate all variables and their potential values in relevant chemical systems. 
Commonly, many of the essential factors describing a system are implicit (\enquote{known unknowns} or \enquote{unknown unknowns}). 

\begin{figure}[ht]
    \centering
\includegraphics[width=1\textwidth]{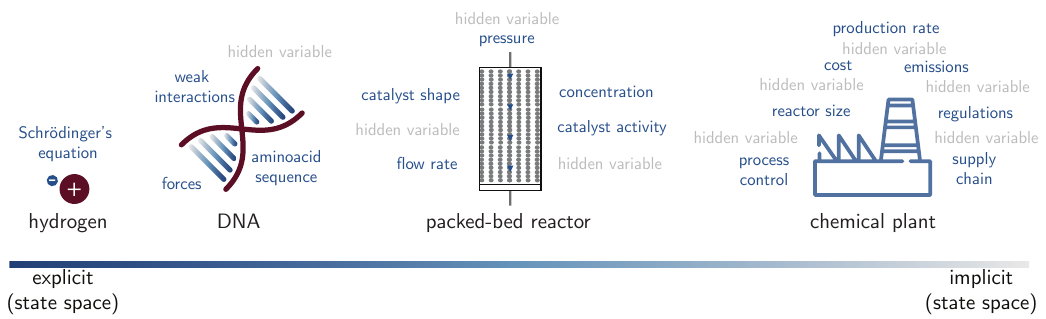}   
    \caption{\textbf{State space description for chemistry at different scales}. We illustrate how the number of hidden variables (gray) is growing with scale and complexity. For simple systems, we can explicitly write down all variables with their values and perfectly describe the system. For more complex systems---closer to practical applications---we can no longer do that. Many more variables cannot be explicitly enumerated.}
    \label{fig:shape_of_data}
\end{figure}

\paragraph{Irreducible Complexity}
\Cref{fig:shape_of_data} illustrates how the state space of chemistry tends to grow more implicit as we move from describing single atoms or small molecules \textit{in vacuo}, to real-world systems. 
For instance, we can completely explain almost all observed phenomena for a hydrogen atom using the position (and atomic numbers) of the hydrogen atom via the Schrödinger equation. 
As we scale up to larger systems such as macromolecular structures or condensed phases, we have to deal with more \enquote{known unknowns} and \enquote{unknown unknowns}.\autocite{martin2022bridging}
For example, it is currently impossible to model a full packed-bed reactor at the atomistic scale because the problem scales with the number of parameters that can be tuned. Often, it becomes infeasible to explicitly label all variables and their values. 
We can describe such complexity as \enquote{irreducible}\autocite{Pietsch_2017}, in contrast to \enquote{emergent} complexity that emerges from systems that can be described with simple equations, such as a double pendulum.

\paragraph{Emergent Complexity} In contrast to irreducible complexity, there is a subset of chemical problems for which all relevant parameters can explicitly be listed, but the complexity emerges from the potentially chaotic interactions among them. A well-known example is the Belousov-Zhabotinsky reaction, \autocite{Cassani2021BZ} which exhibits oscillations and pattern formation as a result of a complex chemical reaction network. 
Individual chemical reactions within the network are simple, but their interactions create a dynamic, self-organizing system with properties not seen in the individual components.
An example of how fast such a parameter space can grow was provided by \textcite{NEURIPS2024_53704142}, who show that a single reaction type and a few hundred molecular building blocks can create tens of thousands of possible solutions. 
When scaling up to only five reaction types, the exploration of the entire space can become intractable, estimated at approximately $10^{22}$ solutions. 

Knowing the ratio between explicit and implicit parameters helps in selecting the appropriate model architecture. 
If most of the variance is caused by explicit factors, these can be incorporated as priors or constraints in the model, thereby increasing data efficiency. 
This strategy can, for instance, be applied in the development of force fields where we know the governing equations and their symmetries, and can use them to enforce such symmetries in the model architecture (as hard restrictions to a family of solutions). \autocite{unke2021machine,Musil_2021}
However, when the variance is dominated by implicit factors, such constraints can no longer be formulated, as the governing relationships are not known. 
In those cases, flexible \glspl{gpm} with soft inductive biases---which guide the model toward preferred solutions without enforcing strict constraints on the solution space\autocite{wilson2025deep}---are more suitable. \glspl{gpm} such as \glspl{llm} fall into this category.

\subsection{Scale of Chemical Data}
Chemistry is an empirical science in which every prediction bears the burden of proof through experimental validation.\autocite{zunger2019beware} 
However, there is often a mismatch between the realities of a chemistry lab and the datasets on which \gls{ml} models for chemistry are trained. 
Much of current data-driven modeling in chemistry focuses on a few large, structured, and highly curated datasets, where most of the variance is explicit (reducible complexity). 
Such datasets, for example \modelname{QM9},\autocite{ramakrishnan2014quantum} often come from quantum-chemical computations.
Experimental chemistry, however, tends to have a significantly higher variance and a greater degree of irreducible complexity. 
In addition, since data generation is often expensive, datasets are small. Because science is about doing new things for the first time, many datasets also contain at least some unique variables.

Considering the largest chemistry text dataset, \modelname{ChemPile},\autocite{mirza2025chempile0} which was produced by curating diverse datasets, we find that the largest dataset is approximately three million times larger than the smallest one (see \Cref{tab:small_large_datasets}).

\begin{table}[!h]
    \centering
    \caption{\textbf{Token counts for the three largest and smallest datasets in the \modelname{ChemPile}\autocite{mirza2025chempile0} collection.} Dominating datasets contribute a large portion of the total token count (a token represents the smallest unit of text that a \gls{ml} model can process), with the small datasets significantly increasing the diversity.}
    \label{tab:dataset-sizes}
    \begin{tabular}{lr}
        \toprule
        \textbf{Dataset} & \textbf{Token count} \\
        \midrule
        \multicolumn{2}{l}{\textit{Three largest ChemPile datasets}} \\
        \midrule
        NOMAD crystal structures\autocite{scheidgen2023nomad} & 5,808,052,794 \\
        \gls{ord}\autocite{Kearnes_2021} reaction prediction & 5,347,195,320 \\
        \modelname{RDKit} molecular features & 5,000,435,822 \\
        \midrule
        \multicolumn{2}{l}{\textit{Three smallest ChemPile datasets}} \\
        \midrule
        Hydrogen storage materials\autocite{hymarcReversibleHydrides} & 1,935 \\
        List of amino acids\autocite{alberts2002molecular} & 6,000 \\
        \gls{ord}\autocite{Kearnes_2021} recipe yield prediction& 8,372 \\
        \bottomrule
    \end{tabular}
    \label{tab:small_large_datasets}
\end{table}

The prevalence of many small, specialized datasets over large ones is commonly referred to as \enquote{the long tail problem}.\autocite{heidorn2008shedding}

\begin{figure}[ht]
    \centering
    \includegraphics{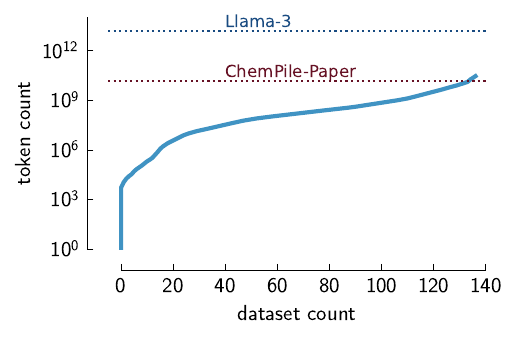}
    \caption{\textbf{Cumulative token count based on the \modelname{ChemPile} tabular datasets \autocite{mirza2025chempile0}}. We compare the approximate token count for three datasets: \modelname{Llama-3} training dataset,\autocite{grattafiori2024llama} openly available chemistry papers in the \modelname{ChemPile-Paper} dataset, and the \modelname{ChemPile-LIFT} dataset. As can be seen, by aggregating the collection of tabular datasets converted to text format in the \modelname{ChemPile-LIFT} subset, we can achieve the same order of magnitude as the collection of open chemistry papers. However, without smaller datasets, we cannot capture the breadth and complexity of chemistry data, which is essential for training \gls{gpm}. The tokenization method for both \modelname{ChemPile} and \modelname{Llama-3} is provided in the respective papers.}
    \label{fig:scale_of_data}
\end{figure}

This can be seen in \Cref{fig:scale_of_data}. 
We show that while a few datasets are large, the majority of the corpus consists of small but collectively significant and chemically diverse datasets.
The actual tail of chemical data is even larger, as \Cref{fig:scale_of_data} only shows the distribution for manually curated tabular datasets and not all data actually created in the chemical sciences.
Given that every dataset in the long tail has its unique characteristics---it is difficult to leverage this long tail with conventional \gls{ml} techniques. However, the promise of \glspl{gpm} is that they can flexibly integrate and jointly model the diversity of small datasets that exist in the chemical sciences.

\subsection{Dataset Creation}
\label{sec:dataset_creation}

Training models requires data. For \glspl{gpm}, the training data must be large and diverse. 
While raw data can be ingested directly, pre-processed data often works better.

\begin{figure}[ht]
    \centering
    \includegraphics{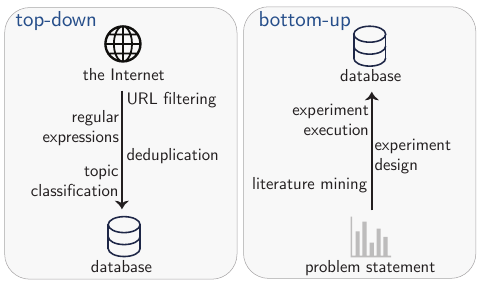}
    \caption{\textbf{Dataset creation protocols}. In \enquote{top-down} approaches, we curate a large corpus of data, which can be used to train \glspl{gpm}. The \enquote{bottom-up} approach starts from a problem definition, and the dataset can be collected via literature mining and experiments. Both approaches can use synthetic data to increase the data size and diversity.}
    \label{fig:data_protocols}
\end{figure}

Strategies for compiling data fall into two groups (see \Cref{fig:data_protocols}).
One can utilize a \enquote{top-down} approach where a large and diverse pool of data---e.g., results from web-crawled resources such as \modelname{CommonCrawl}\autocite{commoncrawl}---is filtered using custom-built procedures (e.g., using regular expressions or classification models). 
This approach is gaining traction in the development of foundation models such as \glspl{llm}.\autocite{penedo2023refinedweb,penedo2024fineweb,guo2025deepseek} 
Alongside large filtered datasets, various data augmentation techniques have further increased the performance of \glspl{gpm}.\autocite{maini2024rephrasing,pieler2024rephrasing}

Alternatively, one can take a \enquote{bottom-up} approach by specifically creating novel datasets for a given problem---an approach which has been very popular in \gls{ml} for chemistry. 

In practice, a combination of both approaches is often used. In most cases, key techniques include filtering and generating synthetic data.

\subsubsection{Filtering}

While initially the focus was on training on maximally large datasets---enabled by the availability of ever-growing computational resources.\autocite{krizhevsky2012imagenet,kaplan2020scaling, hooker2020hardware, dotan2019value0laden}---empirical evidence has shown that smaller, higher-quality datasets can lead to better results.\autocite{gunasekar2023textbooks, marion2023less} 
For example, \textcite{shao2024deepseekmath0} filtered \modelname{CommonCrawl} for mathematical text using a combination of regular expressions and a custom, iteratively trained classification model. 
An alternative approach was pursued by \textcite{thrush2024improving} who introduced a training-free framework. In this method, the pre-training text was chosen by measuring the correlation of each web-domain's perplexity (a metric that measures how well a language model predicts a sequence of text)---as scored by $90$ publicly-available \glspl{llm}---with downstream benchmark accuracy. 

In the chemical domain, \modelname{ChemPile}\autocite{mirza2025chempile0} is an open-source, pre-training scale dataset that underwent several filtering steps. 
For example, a large subset of the papers in \modelname{ChemPile-Paper} comes from the \modelname{Europe PMC} dataset.\autocite{EuropePMC}
To filter for chemistry papers, a custom classification model was trained from scratch using topic-labeled data from the \modelname{CAMEL}\autocite{li2023camel} dataset. 
To evaluate the accuracy of the model, expert-annotated data was used.

\subsubsection{Synthetic Data}
\label{sec:syn-data}

Instead of only relying on existing datasets, one can also generate synthetic data. 
Generation of synthetic data is often required to augment scarce real-world data, but can also be used to achieve the desired model behavior (e.g., invariance in image-based models).

These approaches can be grouped into rule-based and generative methods. 
Rule-based methods apply manually defined transformations---such as rotations and mirroring---to present different representations of the same instance to a model. 
In contrast, generative augmentation creates new data by applying transformations learned through a \gls{ml} model.

\paragraph{Rule-based Augmentation} The transformations applied for generating new data in rule-based approaches vary depending on the modality (e.g., image, text, or audio).  
The most common application of rule-based techniques is on images, via image transformations such as distortion, rotation, blurring, or cropping.\autocite{shorten2019survey} 
In chemistry, tools like \modelname{RanDepict}\autocite{brinkhaus2022randepict} have been used to create enriched datasets of chemical representations. 
These tools generate drawings of chemical structures that mimic the common illustrations found in scientific literature or even in patents (e.g., by applying image templates from different publishers, or emulating the style of older manuscripts).

Rule-based augmentations can also be applied to text. Early approaches involved operations like random word swapping, random synonym replacement, and random deletions or insertions, which are often labeled \enquote{easy augmentation} methods.\autocite{shorten2021text,wei2019eda0}

In chemistry, text templates have been used.\autocite{xie2023darwin,mirza2025chempile0, jablonka2024leveraging, vanherck2025assessment}  
Such templates define a sentence structure with configurable fields, which are then filled using structured tabular data. 
However, it is still unclear how to best construct such templates, as studies have shown that the same data shown in different templates can lead to distinct generalization behavior.\autocite{gonzales2024evaluating} 

We can also apply rule-based augmentation for specific molecular representations (for more details about representations see \Cref{sec:common_representations}). 
For example, the same molecule can be represented with multiple different, yet valid \gls{smiles} strings. 
\textcite{bjerrum2017smiles} used this technique to augment a predictive model, where multiple \gls{smiles} strings were mapped to a single property. 
When averaging the predictions over multiple \gls{smiles} strings, at least a $10\%$ improvement was observed compared to their single \gls{smiles} counterparts. 
Such techniques can be applied to other molecular representations (e.g., \gls{iupac} names or \gls{selfies}), but historically, \gls{smiles} has been used more often.\autocite{kimber2021maxsmi,born2023chemical,arus2019randomized,Tetko_Karpov_VanDeursen_Godin_2020}

A broad array of augmentation techniques has been applied to spectral data---from simple noise addition\autocite{ke2018convolutional,moreno2022application} to physics-informed augmentations (e.g., through DFT simulations).\autocite{oviedo2019fast,gao2020general}

\paragraph{Generative Augmentation}
In some cases, it is not possible to write down augmentation rules. 
For instance, it is not obvious how text can be transformed into different styles using rules alone.
Recent advances in deep learning have facilitated a more flexible approach to synthetic data generation. \autocite{maini2024rephrasing} 
A simple technique is to apply contextual augmentation \autocite{kobayashi2018contextual}, which implies the sampling of synonyms from a probability distribution of a \gls{lm}. 
Another technique is \enquote{back translation},\autocite{edunov2018understanding} a process in which text is translated to another language and then back into the original language to generate semantically similar variants. 
While this technique is typically used within the same language,\autocite{lu2024mathgenie0} it can also be extended to multilingual setups\autocite{hong2024cantonmt0}.
 
Other recent approaches have harnessed auto-formalization\autocite{NEURIPS2022_d0c6bc64}, a \gls{llm}-powered approach that can turn natural-language mathematical proofs into computer-verifiable mathematical languages such as \modelname{Lean}\autocite{de2015lean} or \modelname{Isabelle}\autocite{wenzel2008isabelle}. 
Such datasets have been utilized to advance mathematical capabilities in \glspl{lm}.\autocite{xin2024deepseek,trinh2024solving}

A drawback of generatively augmented data is that its validity is cumbersome to assess at scale, unless it can be verified automatically by a computer program. 
In addition, it was demonstrated that an increasing ratio of synthetic data can facilitate model collapse.\autocite{kazdan2024collapse,shumailov2024ai}

\subsection{Future Directions}

A primary obstacle in the development of \glspl{gpm} for chemistry is the immense scale of data required for pre-training, which reaches into the trillions of tokens. This demand is illustrated by models like \modelname{Llama 3}, trained on 15 trillion tokens. Yet the largest open-source chemistry corpus available contains only approximately 75 billion tokens.\autocite{mirza2025chempile0} 
Beyond its insufficient volume, this dataset is constrained by restrictive licenses and is not ideally suited for the primary pre-training phase. Furthermore, existing data resources lack documentation of negative or failed experiments and reasoning data related to routine laboratory tasks. The absence of such data impedes the development of robust chemistry problem-solving and planning capabilities in \glspl{gpm}. 
This situation stands in contrast to fields like mathematics, where initiatives such as DeepSeek have successfully leveraged large, domain-specific datasets---for instance, 120 billion math tokens---for continual pre-training \autocite{shao2024deepseekmath0}.

Despite the apparent difficulty of amassing diverse data on this scale, we contend that this challenge is accessible through a coordinated community effort.   
\section{Building Principles of GPMs}

\subsection{Taxonomy of Foundation Models}
\label{sec:taxonomy}

In this review, we focus on \glspl{gpm}.
Currently, \glspl{llm} are the most prominent members of the \gls{gpm} family, but many of the principles discussed here are transferable across different types of \glspl{gpm}.

In the following, we discuss the inner workings of such models and the process of building them.

\subsection{Representations}
To interact with any machine, we need to convert the input into numeric values. 
At its core, all information within a computer is represented as bits (zeros and ones). 
Bits are grouped into bytes (8 bits), and meaning is assigned to these sequences through encoding schemes like \texttt{ASCII} or \texttt{UTF-8}. 
Everything---text, a pixel in an image, or even a chemical structure---can be stored as sequences of bytes. 
For example, \enquote{\ce{H2O}} can be translated into the byte sequence, \enquote{H}, \enquote{2}, \enquote{O}. 
However, using raw byte sequences for \gls{ml} presents significant computational inefficiency as representing chemical entities requires long byte sequences, and models would need to learn complex mappings between arbitrary byte patterns and their meanings (as the encoding schemes are not built around chemical principles). 
Furthermore, handling variable-length sequences can pose additional challenges for models, as they may struggle to perform well on unseen inputs. \autocite{zhou2023algorithms,baillargeon2022assessing} 

A more efficient mapping that is built on top of the underlying byte representation is \gls{ohe}. 
Instead of working with variable-length byte sequences, we create a fixed vocabulary (\{\ce{H2O}, \ce{CO2}, \ce{HCl}\}) where each discrete category (in this case, molecule) gets a unique vector: \ce{H2O} becomes [1, 0, 0], \ce{CO2} becomes [0, 1, 0], and so on. 
This provides unambiguous, computationally manageable representations. 
As the number of categories grows, one-hot vectors become increasingly long and sparse, making them computationally inefficient---particularly for large vocabularies, i.e., many categories.
For example, we need a vocabulary of size 118 to model only the unique elements in the periodic table. 
Now, imagine the vocabulary required for all unique compounds---assuming one vocabulary element per compound, the size combinatorially explodes.
More importantly, while \gls{ohe} distinguishes molecules or elements, it still treats them as entirely independent. 
It does not capture any properties of the entity it represents. For example, the ordering of numbers (such as $4<5$) or chemical similarities (such as \ce{Cl} being more similar to \ce{Br} but less similar to \ce{Na}) would not be preserved. \autocite{chuang2018comment}
Embeddings (learned encodings), that we will discuss in \Cref{sec:embeddings}, solve this through learning dense vector representations.

\subsubsection{Common Representations of Molecules and Materials}
\label{sec:common_representations}

Before any chemical entity can be converted into a numerical vector---whether through simple \gls{ohe} or complex learned embeddings---it must first be described in a standardized format (for example, if we are working with materials, it should be able to encode all materials), which is then mapped to encodings. 

For complex entities like molecules, materials, and reactions, this choice of what fundamental units to represent  (\enquote{should we include only atomic numbers?}, \enquote{Should we include something about the coordinates?}, etc.) is thus among the most consequential decisions in building a model. 
It determines the inductive biases---the set of assumptions that guide learning algorithms toward specific patterns over others.\autocite{huang2016understanding} 
The landscape of chemical representations reflects different answers to this question, each making distinct trade-offs between simplicity, expressiveness, and computational efficiency (see \cref{tab:representation_table}).

\begin{longtable}{%
  >{\raggedright\arraybackslash}p{0.20\textwidth}
  >{\raggedright\arraybackslash}p{0.15\textwidth}
  >{\raggedright\arraybackslash}p{0.25\textwidth}
  >{\raggedright\arraybackslash}p{0.32 \textwidth}
}
  \caption{\textbf{Comparison of common molecular representations}. For the encoded information contained by each representation, we followed the criteria used by \textcite{alampara2024mattext}. The examples shown are \textit{aspirin} for elemental composition, \gls{iupac} name, \gls{smiles}, \gls{selfies}, \glslink{inchi}{InChI}, graphs, 3D coordinates; and \textit{silicon} for \glslink{cif}{CIF}, condensed \glslink{cif}{CIF}, \glslink{slices}{SLICES}, \glslink{localenv}{Local-Env}, and natural-language description. Two non-canonical \gls{smiles} are shown to illustrate ambiguity. The examples for 3D coordinates, \glslink{cif}{CIF}, and natural-language description are truncated to fit in the table. For the multimodal representation, only one of the possible modalities is shown ($^{13}$C \glslink{nmr}{NMR} spectrum).}
  \label{tab:molecular-representations} \\
  \toprule
  \textbf{Representation} & \textbf{Encoded information} & \textbf{Description} & \textbf{Example} \\
  \midrule
  \endfirsthead

  \multicolumn{4}{c}%
  {\tablename\ \thetable{} — continued from previous page} \\
  \toprule
  \textbf{Representation} & \textbf{Encoded info} & \textbf{Description} & \textbf{Example} \\
  \midrule
  \endhead

  \midrule
  \multicolumn{4}{r}{Continued on next page} \\
  \endfoot

  \bottomrule
  \endlastfoot
    Elemental composition & Stoichiometry & Always available, but non-unique. & C9H8O4 \\
    \addlinespace
    \gls{iupac} name & Stoichiometry, bonding, geometry & Universally understood, systematic nomenclature, unmanageable for large molecules, and lacks detailed 3D information. & 2-acetyloxybenzoic acid \\
    \addlinespace
    \gls{smiles} \autocite{weininger1988smiles} & Stoichiometry, bonding & Massive public corpora and tooling support, however, there are several valid strings per molecule, and it does not contain spatial information. & \footnotesize \makecell[tl]{%
    \smi{CC(=O)OC1=CC=CC=C1C(=O)O}\\[10pt]      
    \smi{O=C(O)c1ccccc1OC(C)=O}\\[10 pt]
    \emph{etc.}
    } \\
    \addlinespace
    \gls{selfies} \autocite{krenn2020self,cheng2023group} & Stoichiometry, bonding & 100\% syntactic and semantic validity by construction, including meaningful grouping. & \footnotesize \texttt{[C][C][=Branch1][C][=O][O]} \texttt{[C][=C][C][=C][C][=C]} \texttt{[Ring1][=Branch1][C]} \texttt{[=Branch1][C][=O][O]} \\
    \addlinespace
    \gls{inchi} & Stoichiometry, bonding & Canonical one-to-one identifier; encodes stereochemistry layers. & \footnotesize \texttt{InChI=1S/C9H8O4/c1-6(10)13} \texttt{-8-5-3-2-4-7(8)9(11)12/} \texttt{h2-5H,1H3,(H,11,12)} \\
    \addlinespace
    Graphs & Stoichiometry, bonding, geometry & Strong inductive bias that works with \glspl{gnn}. Symmetry-equivariant variants available. Long-range interactions are implicit. & \cellimage{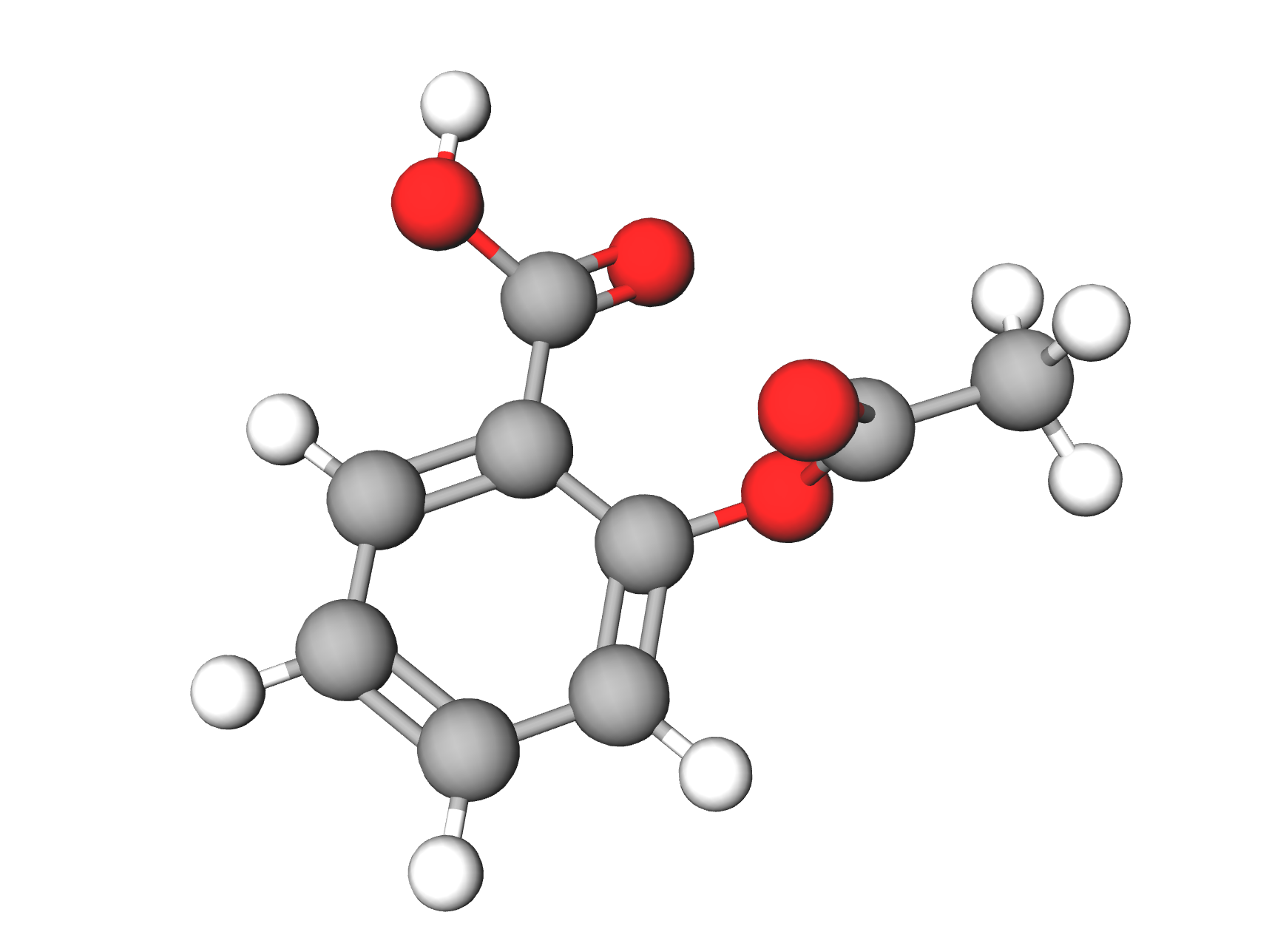} \\
    \addlinespace
    xyz representation & Stoichiometry, geometry & Exact spatial detail. It is high dimensional, and orientation alignment is needed. & \footnotesize 1.2333    0.5540    0.7792 O -0.6952   -2.7148   -0.7502 O 0.7958   -2.1843    0.8685 O 1.7813    0.8105   -1.4821 O -0.0857    0.6088    0.4403 C \ldots \\ 
    \addlinespace
    Multimodal & Stoichiometry, bonding, geometry, symmetry, periodicity, coarse graining & Combines complementary signals; boosts robustness and coverage. It is hard to implement, the complexity scales with the amount of representations, some modalities are data-scarce, and the information encoded totally depends on the modalities included. & \cellimage{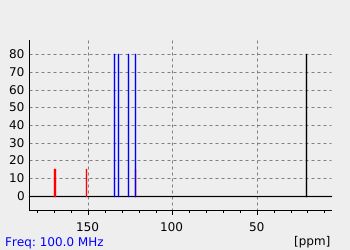} \\
    \addlinespace
    \gls{cif} \autocite{hall1991crystallographic} & Stoichiometry, bonding, geometry, periodicity & Standardized and widely supported, however, it carries heterogeneous keyword sets and parser overhead & \footnotesize \texttt{data\_Si \_symmetry\_space\_group\_name\_H-M   'P 1' \_cell\_length\_a   3.85 \ldots \_cell\_angle\_alpha   60.0 \ldots \_symmetry\_Int\_Tables\_number   1 \_chemical\_formula\_structural   Si \_chemical\_formula\_sum   Si2 \_cell\_volume   40.33 \_cell\_formula\_units\_Z   2 loop\_ \_symmetry\_equiv\_pos\_site\_id  \_symmetry\_equiv\_pos\_as\_xyz   1  'x, y, z' loop\_ \_atom\_type\_symbol \_atom\_type\_oxidation\_number  Si0+  0.0loop\_ \_atom\_site\_type\_symbol \_atom\_site\_label \_atom\_site\_symmetry\_multiplicity \_atom\_site\_fract\_x \ldots \_atom\_site\_occupancy  Si0+  Si0  1  0.75  0.75  0.75  1.0 Si0+  Si1  1  0.0  0.0  0.0  1.0}\\ 
    \addlinespace
    Condensed \gls{cif} \autocite{gruver2024finetuned, antunes2024crystal} & Stoichiometry, geometry, symmetry, periodicity & Good for crystal generation tasks. It omits occupancies and defects, custom tooling is needed, and only works for crystals & \footnotesize \texttt{3.8 3.8 3.8 59 59 59 Si0+ 0.75 0.75 0.75 Si0+ 0.00 0.00 0.00}\\
    \addlinespace
    \glslink{slices}{SLICES} \autocite{Xiao_2023} & Stoichiometry, bonding, periodicity & Invertible, symmetry-invariant and compact for general crystals. However, it carries ambiguity for disordered sites & \footnotesize \texttt{Si Si 0 1 + + + 0 1 + + o 0 1 + o + 0 1 o + +}   \\
    \addlinespace
    \glslink{localenv}{Local-Env}\autocite{alampara2024mattext} & Stoichiometry, bonding, symmetry, coarse graining & Treats each coordination polyhedron as a \enquote{molecule}, it is transferable and compact; but it ignores long-range order and its reconstruction requires post-processing & \footnotesize \texttt{R-3m Si (2c) [Si][Si]([Si])[Si]} \\
    \addlinespace
    Natural-language description \autocite{ganose2019robocrystallographer} & Stoichiometry, bonding, geometry, symmetry, periodicity, coarse graining & It is human-readable and tokenizable in a meaningful way by pretrained \glspl{llm}. However, trying to encode all the information can lead to verbose, ambiguous descriptions. & \enquote{Silicon crystallizes in the diamond-cubic structure, a lattice you can picture as two face-centred-cubic frameworks gently interpenetrating\ldots} \\
\end{longtable}
\label{tab:representation_table}

A common strategy is to represent chemical information as a sequence of characters. This allows us to leverage architectures initially designed for natural language. 
This approach has found success in language modeling for predicting protein structures and functions, where the amino acid sequence, the foundation of a protein's structure and function, is easily represented as text.\autocite{Rives_2021, Elnaggar_2022, Ruffolo_2024}
The most prevalent string representation for molecules in chemistry is \gls{smiles}\autocite{weininger1988smiles}.
\gls{smiles} strings provide a linear textual representation of a molecular graph, including information about atoms, bonds, and rings. 
However, \gls{smiles} representations have limitations. The same molecule can be represented through multiple valid \gls{smiles} strings (so-called non-canonical representations).
Although the existence of non-canonical representations enables data augmentation (see \Cref{sec:syn-data}), it can also confuse models because the same molecule would have different encodings, each one originating from a different \gls{smiles} string. 
In addition, \gls{smiles} imposes a relatively weak inductive bias; the model must still learn the rules of valence and bonding from the grammar of these character sequences. Moreover, \gls{smiles} does not preserve locality: structural motifs that are directly bonded or physically close to each other in a molecule can be very far apart in the \gls{smiles} representation.

A limitation of \gls{smiles} is that not every \gls{smiles} string corresponds to a valid molecule.
A more robust alternative is \gls{selfies}\autocite{krenn2020self,cheng2023group}, where every \gls{selfies} corresponds to a valid molecule, providing a stronger bias towards chemically plausible structures (chemical validity biases). 
The \gls{inchi} is another standardized string representation. 
Unlike \gls{smiles}, \gls{inchi} strings, as identifiers, are canonical---each molecule has exactly one \gls{inchi} representation.  
This eliminates ambiguity, but comes at the cost of human readability and increased string length. 

In the realm of materials, no natural representation has emerged.  
Previous work has indicated that for certain phenomena (e.g., when all structures in a dataset are in the ground state), composition might implicitly encode geometric information \autocite{tian2022information, Jha_2018, Wang_2021} and composition alone can be predictive of various material properties. Thus, it is a widely chosen method to represent materials, depending on the task. 
When structural information is available, \glspl{cif}, initially proposed as a standard way to archive structural data in crystallography \autocite{hall1991crystallographic}, is now a widely used representation. \textcite{gruver2024finetuned, antunes2024crystal} proposed a condensed version of \glspl{cif}, which includes only the parameters necessary for building the crystal structure in a crystal generation application. \textcite{ganose2019robocrystallographer} aimed to create human-readable descriptions by proposing a tool to generate natural-language descriptions of crystal structures automatically. 
For specific material classes, such as \glspl{mof}, specialized representations like \modelname{MOFid} \autocite{Bucior_2019} have been developed.

As an alternative to strings, we can represent molecules and materials as graphs. 
Here, we directly encode atoms (nodes) and bonds (edges). 
This representation introduces strong locality biases that explicitly inform the model about atomic connectivity, so the model does not need to learn this fundamental principle from scratch. 
Symmetry has been incorporated into many of the best-performing graph-based approaches by designing symmetry-constrained representations \autocite{Langer_2022, Musil_2021} and architectures \autocite{satorras2021n, Batzner_2022}. 

Ultimately, weaker inductive biases (like text) offer greater flexibility and can capture unexpected patterns, but may require more data to learn the fundamental rules. 
The successful design of inductive biases requires balancing domain knowledge with learning flexibility. 
Stricter inductive biases (like graphs) incorporate more domain knowledge, leading to greater data efficiency but potentially limiting the model's ability to discover patterns that contradict our initial assumptions. 

Beyond choosing a single optimal representation, \glspl{gpm} allows for the simultaneous use of multiple representations. 
A chemical entity can be described not only by its textual \gls{smiles} string or its connectivity graph, but also by its experimental or simulated spectra (e.g., \gls{nmr}, \gls{ir}), or even a microscopy image. 
Each of these modalities provides a complementary layer of information. 
A more detailed section on using multiple representations is presented in \Cref{sec:multimodal_chem}.

\subsubsection{Tokenization}
Once we have chosen a representation format---whether \gls{smiles} strings, \gls{cif} files, or chemical formulas---we face another fundamental question: How does a model process these variable-length sequences of characters? 
One might imagine creating a unique identifier or encoding for every single molecule or string. 
It is impractical to have a dictionary entry for every sentence in a language due to the similar scaling problems of \gls{ohe}.

Consider the molecule with the \gls{smiles} string \texttt{CN1C=NC=C1C(=O)}. 
We could break down the representation in several ways: as individual characters (\texttt{C}, \texttt{N}, \texttt{1}, \texttt{C}, \texttt{=}, etc.), as atom-bond pairs (\texttt{CN}, \texttt{C=}, \texttt{NC}), or as fragments (\texttt{CN1}, \texttt{C=NC}, etc.). 
Each choice creates a different \enquote{language} for the model to learn, with distinct computational and learning implications (see example \Cref{example:tokenization}).

This is where tokenization becomes essential. 
It is the strategy of breaking down a complex representation (like a \gls{smiles} string) into a sequence of discrete, manageable units called tokens. 
The core idea is to find a set of common, reusable building blocks. Instead of learning about countless individual molecules, the model knows a much smaller, finite vocabulary of tokens. By learning an encoding for each token, the model gains the ability to understand and construct representations for an immense number of molecules---including those it has never seen before---by combining the meanings of their constituent parts. 
This compositional approach enables generalization.

\begin{examplebox}[label={example:tokenization}]{Different tokenization strategies}
Tokenization strategies for caffeine SMILES (\texttt{CN1C=NC2=C1C(=O)N(C(=O)N2C)C}):
\begin{itemize}
    \item Character-level: [\texttt{C}, \texttt{N}, \texttt{1}, \texttt{C}, \texttt{=}, \texttt{N}, \texttt{C}, \texttt{2}, \texttt{=}, \texttt{C}, \texttt{1}, \texttt{C}, \texttt{(}, \texttt{=}, \texttt{O}, \texttt{)}, \texttt{N}, \texttt{(}, ...]
    \item Chemical fragments: [\texttt{CN1C=NC2=C1}, \texttt{C(=O)}, \texttt{N(}, \texttt{C(=O)}, ...]
\end{itemize}

The choice affects what the model learns. Character-level requires learning chemical rules from scratch, while fragment-level embeds chemical knowledge but needs a larger vocabulary.
\end{examplebox}

The concept of tokenization, or defining the fundamental units of input, extends beyond string-based representations. 
In images, it could be patches of images. 
In graph-based models, the analogous decision is how to define the features for each node (atom) and edge (bond). 
Should a node represent an atomic number (a simple \enquote{token}), or should it be a more complex sub-structure like a structural motif\autocite{bouritsas2022improving} (a richer \enquote{token})? This choice determines the level of chemical knowledge initially provided to the model. 
Ultimately, the tokenization strategy defines the elementary units for which the model will learn embeddings, setting the stage for learning the context-aware representations discussed next.

\subsubsection{Embeddings} \label{sec:embeddings}

Through training, models can learn to map discrete inputs into continuous spaces where similar items have meaningful relationships (for example, similar items cluster in this continuous space). 
In the simplest approach, they can be created by training models (so-called \modelname{Word2Vec} models) that take one-hot encoded inputs and predict the probability of words in the context.\autocite{mikolov2013efficient, mikolov2013distributed, Tshitoyan_2019}
Embeddings are powerful because they learn relationships between entities, allowing for the efficient compression of data and the uncovering of hidden patterns that would otherwise be invisible in the raw data.

The advent of \glspl{gpm} has further underscored the usefulness of high-quality embeddings. 
These models, trained on vast amounts of chemical data, learn to create powerful, generalizable embeddings that can be adapted to a wide range of downstream tasks, from property prediction (see \Cref{sec:prediction}) to molecular generation (see \Cref{sec:mol_generation}).  
In the following sections, we describe the process of generating, refining, and using these embeddings through training and different architectures.

\subsection{General Training Workflow}

\begin{figure}[!ht]
    \centering
    \includegraphics[width=1\textwidth]{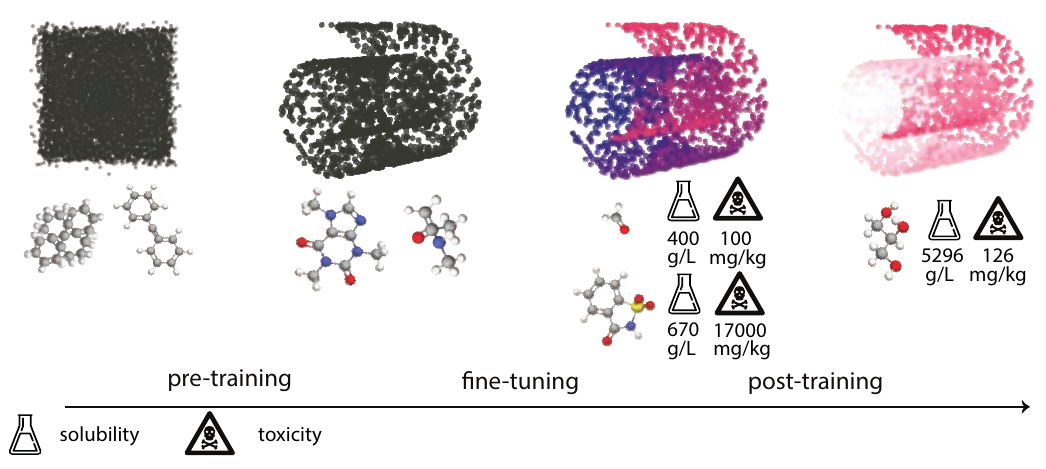}
    \caption{\textbf{General training workflow through the lens of molecular science}. The figure illustrates the progression from pre-training through fine-tuning to post-training stages.  \textbf{(1) Pre-training:} The model learns the underlying data distribution from a vast, unlabeled dataset. This is visualized as transforming an unstructured representation space (left, square cloud) into a structured manifold (the Swiss roll). At this stage, the model has learned the \enquote{shape} of the data: the fundamental rules that make a molecule chemically valid. However, the representations are not yet specialized for any task. 
    \textbf{(2) Fine-tuning:} The model is trained on specific, labeled tasks, such as predicting solubility (flask icon) and toxicity (skull icon). This process \enquote{colors} the manifold, adjusting the learned representations so that their position now also correlates with specific properties (e.g., blue for one property profile, red for another). 
    \textbf{(3) Post-training Alignment:} The model's behavior is biased towards desired outcomes. This is visualized as preferentially sampling from a specific region of the colored manifold, such as generating molecules predicted to have high solubility and low toxicity (right, the brighter red region).}
    \label{fig:training_workflow}
\end{figure}

The entire training process of a \gls{gpm} typically contains multiple steps that can be divided into two broad groups (see \Cref{fig:training_workflow}). \autocite{howard2018universal} 
The first step is pre-training, which is usually done in a self-supervised manner and focuses on learning a data distribution---the underlying set of rules and patterns that make up the data. 
Imagine all possible arrangements of atoms, both real and unfeasible. 
The data distribution describes which molecules are \enquote{likely} (stable, following chemical rules) and which are \enquote{unlikely} or \enquote{impossible} (random assortments of atoms).

 In pre-training the model learns the \enquote{grammar} of chemistry---the principles that make a molecule physically plausible---by observing millions of valid examples. 
 A model that has successfully learned the distribution can distinguish a valid structure from noise and can even generate new, chemically sensible examples, much like someone who has learned the rules of a language can form new, grammatically correct sentences.  

A model does not learn the data distribution by storing an explicit formula. Instead, during pre-training (see \Cref{sec:pretraining} for more details), it learns the high-dimensional transformation (a mapping function) to create an internal representation---an embedding (see \Cref{sec:embeddings}).
The training process guides the model to map inputs to these embeddings in a high-dimensional space, where representations of similar, valid inputs are clustered together.

The second step is post-training, also called fine-tuning, in which the model is adapted to learn task-specific labels and capabilities, essentially \enquote{coloring} the learned structure with domain-specific knowledge. 
Crucially, fine-tuning does not discard the learned distribution but refines it. 
As shown in \Cref{fig:training_workflow}, the fundamental shape of the manifold (the Swiss roll) is preserved. 
The \enquote{coloring} process corresponds to adjusting the internal representations so they now also encode task-specific properties. 
For example, the model learns to map molecules with high solubility to one region of the manifold (e.g., the red area) and those with high toxicity to another. 
The representation of each molecule is thus enriched, now containing information not just about its structural validity but also about its properties.

Finally, techniques such as \gls{rl} are used to align the model's outputs with preferred choices, e.g., human preferences. 
This step further refines the learned distribution by biasing the model's sampling behavior to favor specific modes of the distribution. 
As depicted in the post-training panel of \Cref{fig:training_workflow}, this biases the output towards a specific section of the colored manifold---in this case, perhaps molecules with high solubility (the brighter pink region).

\subsection{Pre-training: Learning the Shape of Data}
\label{sec:pretraining}

Pre-training establishes the foundational knowledge and capabilities of the model. 
During pre-training, the model learns general patterns, relationships, and structures from massive datasets (often trillions of tokens, see \Cref{fig:scale_of_data}). 
The model learns to map input to internal representations or features through so-called \gls{ssl} objectives like reconstructing corrupted inputs (predicting masked tokens, or predicting future sequences, see \Cref{sec:ssl}). 

This large-scale pre-training allows models to capture rich representations of the statistical distributions inherent to the data.  These learned distributions capture the fundamental patterns and structure of the domain (scientific language grammar, physical and chemical principles that govern materials). \Cref{fig:training_workflow} illustrates the distribution captured, from an uninstructed manifold before pre-training (if you randomly pick from this manifold, you get noise or non-physical molecules) to a structured manifold, where if you sample from this distribution (the black Swiss roll) you get a valid molecule.
For example, the model might learn commonly occurring structures, scientific notations, and scientific terms (see example \ref{example:self_supervision}). 
Furthermore, it might construct hierarchical relationships between these concepts, such as those between chemical compounds, elements, and their properties.  
This distributional learning empowers the model to make predictions about new examples by understanding their relation to the learned patterns. Crucially, this ability stems from the development of transferable features, rather than mere data memorization \autocite{brown2020language}.

\begin{examplebox}[label={example:self_supervision}]{Learning chemical grammar through self supervision}
    Imagine training a model on millions of known molecules without any labels about their activity:

\begin{itemize}
    \item The model learns that structures like \texttt{CC(=O)OC1=CC=CC=C1C(=O)O} (aspirin) are chemically valid
    \item It learns that rings like benzene \texttt{C1=CC=CC=C1} appear frequently
    \item It discovers that certain functional groups often co-occur
    \item It learns some statistical patterns (carbon forms 4 bonds, oxygen forms 2)
\end{itemize}
This \enquote{chemical grammar} contains \enquote{soft rules} and is learned just from seeing valid examples, without anyone explicitly teaching them.
\end{examplebox}

As illustrated via a Swiss roll in \Cref{fig:training_workflow}, the pre-training process creates a structured manifold where invalid inputs are mapped far away. Therefore, learning high-quality representations is the concrete computational method for capturing the abstract statistical distribution of the data; the structure of this representation space is the model's learned approximation of the data's true shape.

\subsubsection{Self-Supervision} \label{sec:ssl}

\glspl{ssl} allows models to learn from unlabeled data by generating \enquote{pseudo-labels} from the data's structure. 
The original, unlabeled data serves as its own \enquote{ground truth}. 
This differs significantly from supervised learning, where each piece of data is explicitly tagged with the correct output, which the model then learns to predict. 
Such manual labeling is often an expensive, time-consuming, and domain-specific process. 
\glspl{ssl} has emerged as a particularly effective strategy for pre-training \glspl{llm}, since natural-language corpora are abundant but rarely annotated. 
Proxy strategies have then been applied to other types of model architectures as well. 
The ability to extract structure from data \textit{without labels} is a key enabler for foundation models and underpins the pre-training phase.

\subsubsection{Families of Self-Supervised Learning}

\begin{figure}[H]
    \centering
    \includegraphics[width=1\textwidth]{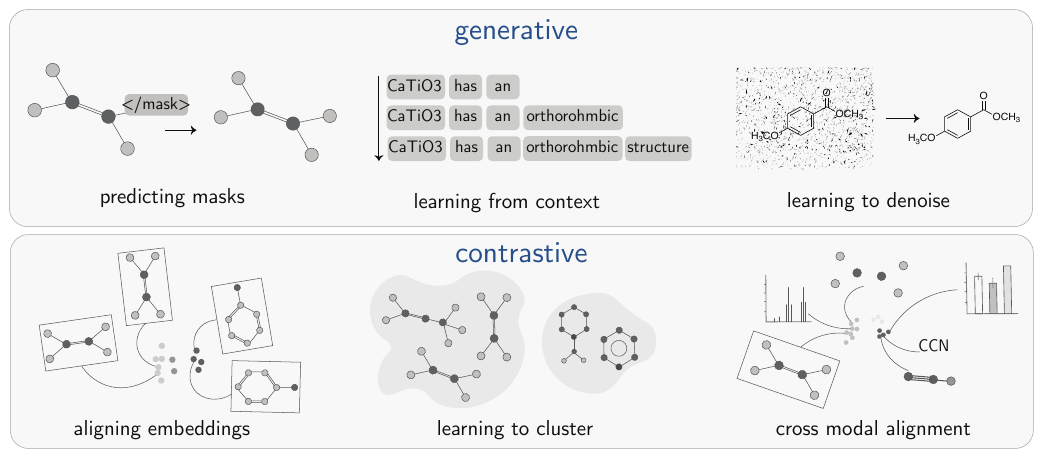}
    \caption{\textbf{Main families in \glspl{ssl}}. The figure illustrates the two primary \gls{ssl} approaches, each using different strategies to generate pseudo-labels from the data itself.
    \textbf{Generative Methods (Top Panel):} This family focuses on reconstruction and prediction. The model learns representations by generating missing information. Examples shown correspond to the pretext tasks discussed in the text: (1) \textit{Predicting masks} in a graph, analogous to masked modeling (more details in \Cref{sec:masked_modeling}); (2) \textit{Learning from context}, which is the basis for next token prediction (more details in \Cref{sec:next_token_prediction}); and (3) \textit{Learning to denoise}, where the model reconstructs a clean input from a corrupted version. (see \Cref{sec:denoising})
    \textbf{Contrastive Learning (Bottom Panel):} This family learns by comparing samples. The model is trained to pull representations of similar samples together while pushing dissimilar ones apart. Examples include: (1) \textit{Aligning embeddings} from different augmentations of the same molecule, a core idea in Instance Discrimination (more details in \Cref{sec:instance_discrimination}); (2) \textit{Learning to cluster} similar molecules together, as in Clustering-based Contrastive Learning (see \Cref{sec:clustering_cl}); and (3) \textit{Cross-modal alignment}, where representations from different data types (e.g., a molecule's graph and its spectral properties) are learned jointly. (see \Cref{sec:contrastive_learning} )}
    \label{fig:types_ssl}
\end{figure}

\gls{ssl} encompasses a variety of approaches. While distinct methods exist, they can be grouped into two main families: generative and contrastive (see \Cref{fig:types_ssl}).

\subsubsection{Generative Methods}

This family of methods focuses on learning representations by reconstructing or predicting parts of the input data from other observed parts. 
The model learns the underlying data distribution by learning to regenerate the missing information. 
Examples shown in \Cref{fig:types_ssl} include predicting masked portions of a graph, learning from surrounding text context, and learning to denoise an image.

\label{sec:masked_modeling}
\paragraph{Masked Modeling}  In this method, portions of the input data are intentionally obscured or \enquote{masked}. 
The model's primary objective is then to reconstruct these hidden segments. \autocite{devlin2018bert0} 
This process can be conceptualized as a \enquote{fill-in-the-blanks} task, compelling the model to infer missing information from its context. This enables the model to develop a deep understanding of contextual dependencies of data's structure and semantics without requiring explicit human-labeled annotations. For chemical data, this could involve masking and predicting tokens in \glspl{smiles} or \glspl{selfies} strings \autocite{chithrananda2020chemberta, zhang2025scientific} (i.e., hiding atoms and training the model to guess what is missing), omitting atom or bond types in molecular graphs \autocite{mahmood2021masked,wang2022molecular, reiser2022graph}, removing atomic coordinates in 3D structures, or masking sites within a crystal lattice (see example \ref{example:masked_modeling}).

\begin{examplebox}[label={example:masked_modeling}]{Crystal structure prediction example}

\textbf{Original version}: \texttt{5.64 5.64 5.64 90 90 90 Na+ 0 0 0 Cl- 0.5 0.5 0.5}
\textbf{~Masked ~~version}: \texttt{~~5.64 5.64 5.64 90 90 90 Na+ 0 0 0 Cl- \textless{}mask\textgreater{}}

The model must predict what goes in \texttt{\textless{}mask\textgreater{}}, a process that may be informed by the following learned rules and contextual clues.

\begin{itemize}
    \item \textbf{Context clues}: The equal cell lengths and  \SI{90}{\degree} angles indicate a cubic symmetry (common for rock salt).
    \item \textbf{Chemical Knowledge}: In ionic crystals like \ce{NaCl}, cations and anions alternate for charge balance; \ce{Cl-} typically occupies octahedral sites offset by half the cell in all directions to minimize repulsion.
    \item \textbf{Correct prediction}: \texttt{0.5 0.5 0.5} (placing \ce{Cl-} at the cube's center for proper packing).
\end{itemize}

This forces the model to understand local ionic coordination (e.g., \ce{Na+} surrounded by 6 \ce{Cl-}) and global crystal architecture.
\end{examplebox}

\paragraph{Next Token Prediction} 
\label{sec:next_token_prediction}
One of the most powerful \gls{ssl} tasks for sequential data, such as text, is next-token prediction.
Here, the core objective is for a model to generate the subsequent token in a given sequence, based on the contextual information provided by preceding tokens. 
Because text unfolds naturally in a sequence, it offers the reference information the model needs to learn. This approach has been applied to chemical and material representations by treating molecular string representations (\glspl{smiles}, \glspl{selfies}, etc.) or material representations as sequences \autocite{adilov2021generative, wang2023cmolgpt, schwaller2019molecular, alampara2024mattext}. 
During training, the model optimization procedure constantly adjusts the model to maximize the likelihood (trying to make good predictions more probable and bad predictions less probable). This is accomplished by making each prediction based on the preceding input, which establishes the conditional context (see \Cref{eq:nexttoken}).

\begin{tcolorbox}[
    title= Cross-Entropy Loss,
    breakable,
    enhanced,
    colback=white,
    colframe=black!30,
    colbacktitle=black!10,
    coltitle=black,
    fonttitle=\bfseries,
    boxrule=0.5mm,
    left=2mm,
    right=2mm,
    top=4mm,
    bottom=4mm,
    middle=10mm
]
\label{eq:nexttoken}
\begin{center}
\begin{minipage}{0.9\linewidth}
\begin{equation}
\mathcal{L} = 
-\mathbb{E} \left[ \sum_{t=1}^{T} 
\log P(\eqnmarkbox[PositiveColor]{pred}{x_t} | \eqnmarkbox[NegativeColor]{context}{x_{context}})
\right]
\label{eq:cross_entropy}
\end{equation}
\annotate[yshift=1.1em]{above}{pred}{Prediction term}
\annotate[yshift=-1.3em]{below}{context}{Conditional context}
\begin{tikzpicture}[overlay, remember picture]
\end{tikzpicture}
\end{minipage}
\end{center}
\tcblower
\begin{itemize}
\item \textbf{Prediction Term} ({\color{PositiveColor}Blue}): The target token $x_t$ that the model is trying to predict at each position.
\item \textbf{Context Tokens} ({\color{NegativeColor}Maroon}): The set of tokens $x_{\text{context}}$ the model uses to make its prediction. The definition of this context depends on the SSL task:
            \begin{itemize}
                \item \textit{For Masked Modeling:} The context is all unmasked tokens in the sequence.
                \item \textit{For Next-Token Prediction:} The context is the preceding tokens ($x_{<t}$).     
            \end{itemize}
\item \textbf{Summation} $\sum_{t=1}^{T}$: The loss is calculated across all token positions in the sequence of length $T$.
\item \textbf{The Logarithm's Role}: The negative logarithm ($\log P$) heavily penalizes highly confident wrong answers (low $P$, high loss) and lightly rewards confident correct answers (high $P$, low loss).
\item \textbf{Overall Loss Structure}: Cross-entropy loss that encourages the model to assign high probability to the correct next token at each position, given all previous tokens.
\end{itemize}
\end{tcolorbox}

\paragraph{Denoising}
\label{sec:denoising}
Denoising \glspl{ssl} works by intentionally adding noise to the inputs and then training models to reconstruct the original data. 
In this context, the original, uncorrupted data implicitly serves as the label or target for the training process.  In this paradigm, we begin with a clean input, which we can call $x$. 
We then apply a random corruption process to create a noisy version, $\tilde{x}$. 
The model is then trained to reverse the corruption process and recover the original $x$. 
This process is formally expressed as sampling a corrupted input $\tilde{x}$ and optimizing the network to predict $x$. \autocite{vincent2010stacked}
By learning to recover the input, the model is compelled to develop robust representations that are inherently invariant to the types of noise it encounters during training.
This directly forces the model to learn the underlying data distribution.  
To distinguish the original signal from the artificial noise, the model must learn the features of high-probability samples within that distribution. 
For example, to successfully \enquote{denoise} a molecule, it must implicitly understand the rules of chemical plausibility that separate valid structures from random noise.  
Denoising objectives are popular in images \autocite{vincent2008extracting, bengio2013generalized} and have consequently been applied to graph representations of molecules \autocite{wang2023denoise, ni2024pre}. 
For instance, one can randomly perturb atoms or edges in a molecular graph and train a graph neural network to predict the original attributes. 

\subsubsection{Contrastive Learning}
\label{sec:contrastive_learning}
The other main family of \gls{ssl} techniques is contrastive learning. The objective is to train models to understand data by distinguishing between similar and dissimilar samples. 
This is achieved by learning an embedding space where representations of samples that are alike in their core chemical properties or identity are pulled closer together. 
In contrast, representations of samples that are fundamentally different are pushed further apart. \autocite{hadsell2006dimensionality}

This process creates meaningful clusters for related concepts while enforcing separation between unrelated ones. 
In effect, the model learns the data's underlying distribution by defining the distance between its points. 
The resulting internal representations become highly robust because they are trained for invariance; the model learns to focus on essential, identity-defining features while disregarding irrelevant variations. 
This process, often referred to as embedding alignment, ensures that the representations capture the core characteristics shared among similar samples (see example \ref{example:contrastive}).

There are many contrastive learning approaches with variations in loss functions. A key design choice in contrastive learning is whether to compute the contrastive loss on an instance basis or a cluster basis.

\paragraph{Instance Discrimination}
\label{sec:instance_discrimination}
Instance Discrimination is the most dominant paradigm in recent contrastive learning.
Each instance (sample) in the dataset is treated as its own distinct class.  
This is typically achieved using contrastive loss functions like \modelname{InfoNCE} see \Cref{eq:infonce}. \autocite{oord2018representation} As detailed in \Cref{eq:infonce}, the loss function is formulated as a categorical cross-entropy loss where the task is to classify the positive sample correctly among a set of negatives plus the positive itself.

In materials and chemistry, this can involve aligning the textual representation of a structure with a graphical representation, image, or other visual method to represent a molecule.\autocite{seidl2023enhancing}
The model could also learn from augmentations of a structure, such as being given several valid \gls{smiles} strings that all describe the identical molecule. 

\begin{tcolorbox}[
    title=InfoNCE Loss Function,
    breakable,
    enhanced,
    colback=white,
    colframe=black!30,
    colbacktitle=black!10,
    coltitle=black,
    fonttitle=\bfseries,
    boxrule=0.5mm,
    left=2mm,
    right=2mm,
    top=4mm,
    bottom=4mm,
    middle=10mm
]

\begin{center}
\begin{minipage}{0.9\linewidth}

\begin{equation}
\label{eq:infonce}
\hspace*{-2em}
\mathcal{L} = 
-\mathbb{E} \left[ \log \frac{
\eqnmarkbox[PositiveColor]{pos}{\exp({\text{sim}(f(\mathbf{x}_i), f(\mathbf{x}_i^+))/\tikzmarknode{tau1}{\tau}})}
}{
\eqnmarkbox[PositiveColor]{pos2}{\exp({\text{sim}(f(\mathbf{x}_i), f(\mathbf{x}_i^+))/\tau})} + 
\eqnmarkbox[NegativeColor]{neg}{\sum_{j=1}^{N} \exp({\text{sim}(f(\mathbf{x}_i), f(\mathbf{x}_j^-))/\tikzmarknode{tau2}{\tau}})}
} \right]
\end{equation}

\annotate[yshift=1.5em]{above}{pos}{Positive pair similarity}
\annotate[yshift=-1.5em]{below}{neg}{Negative pairs similarity}
\annotate[yshift=1em, xshift=1em]{right}{tau1}{Temperature parameter}

\begin{tikzpicture}[overlay, remember picture]
\end{tikzpicture}
\end{minipage}
\end{center}

\tcblower

\begin{itemize}
\item \textbf{Positive Pair Term} ({\color{PositiveColor}Blue}): Measures similarity between an anchor sample $\mathbf{x}_i$ and its positive pair $\mathbf{x}_i^+$ (e.g., different view of the same molecule).

\item \textbf{Negative Pairs Term} ({\color{NegativeColor}Maroon}): Sum of similarities between anchor sample $\mathbf{x}_i$ and all negative pairs $\mathbf{x}_j^-$ (e.g., different molecules).

\item \textbf{Temperature Parameter} $\boldsymbol{\tau}$ 
: Controls the sharpness of the distribution. Lower values make the model more sensitive to hard negatives.

\item \textbf{Overall Loss Structure} 
: A negative log probability that encourages the model to maximize similarity for positive pairs while minimizing it for negative pairs.
\end{itemize}

\end{tcolorbox}

\begin{examplebox}[label={example:contrastive}]{Learning chemcial similarity through contrast}
Consider these three molecules:

Molecule A: Aspirin \texttt{(CC(=O)OC1=CC=CC=C1C(=O)O)}

Molecule B: Salicylic acid \texttt{(OC1=CC=CC=C1C(=O)O)} 

Molecule C: Glucose \texttt{(OC[C@H]1OC(O)[C@H](O)[C@@H](O)[C@@H]1O)} \\

Contrastive learning might:
\begin{itemize}
    \item Pull A and B together (both contain benzene ring + carboxylic acid)
    \item Push A and C apart (completely different structures and properties)
    \item Learn that aromatic compounds cluster separately from sugars
\end{itemize}
The model learns that structural similarity often correlates with chemical properties.
\end{examplebox}

\paragraph{Clustering-based Contrastive Learning} 
\label{sec:clustering_cl}
Clustering approaches leverage the idea that similarity often translates to closeness in the feature space. Methods like \modelname{DeepCluster} \autocite{caron2018deep}  iteratively train a model. 
First, they group the generated features (internal representation) of a dataset into distinct sets using a common grouping algorithm, such as $k$-means clustering. Imagine you have a pile of diverse objects; $k$-means would help you sort them into a predefined number of piles based on their similarities, like color or shape. 
These assigned groups then act as temporary \enquote{pseudo-labels} to train the network. 
The supervised training step implicitly contrasts samples from different clusters. The clustering and training steps alternate. 
Take a dataset of molecular fingerprints as an example. A model can be trained to predict the clustering pattern of this fingerprint data, distinguishing between functional group types or structures. Thus, the model learns representations that group chemically or structurally similar fingerprints.

\subsection{Building Good Internal Representation} 

The design of effective pretext tasks---such as specific versions of instance discrimination (identifying unique examples) or denoising (recovering original data from corrupted versions) is where deep domain expertise becomes invaluable.

The pretext tasks must be meaningful, preserving the core identity of the molecule or material while introducing sufficient diversity to challenge the model and allow it to learn robust invariances.

For instance, a suboptimal technique would be to shuffle all the atoms in the text representation of a molecule. 
This would destroy the molecule's chemical meaning, which would hinder the model's ability to learn chemically meaningful features. 
Good augmentations enable richer features by providing additional layers of information to learn from, such as generating different low-energy conformers or using non-canonical string representations.

\paragraph{Parallels between Generative and Contrastive Objectives}

While it might seem that generative and contrastive \glspl{ssl} methods optimize different things, their underlying goals can be equivalent. 
A generative masked language model learns the conditional probability
(see \Cref{eq:infonce})
, aiming to assign a high probability to the correct masked token by effectively discriminating it from other vocabulary tokens. 
The \modelname{InfoNCE} loss in contrastive learning can be viewed as a log-loss for a $(K+1)$-way classification task (see \Cref{eq:infonce}). 
Here, the model learns to identify the positive pair $f(x_i^+)$ as matching $f(x_i)$ from a set including $f(x_i^+)$ and $K$ negative features $f(x_j^-)$. 
Both approaches effectively learn to select the \enquote{correct} item (a token or a positive feature) from a set of candidates based on the provided context or an anchor. 
To do so, they must effectively build strong internal representations.

\paragraph{Pre-training beyond \gls{ssl}}
Pre-training be performed using \gls{ssl} on multiple modalities.  For example, in models that consider multiple input formats (multimodality, as explained in detail in \Cref{sec:multimodal_chem}), alignments between different modalities (e.g., text-image, text-graph) serve as a pre-training step.\autocite{weng2022vlm,girdhar2023imagebind0}  
General-purpose force fields are commonly trained in a supervised manner on relaxation and simulation trajectories.\autocite{batatia2022mace, wood2025uma0}
Thus, the model learns a representation of connectivity patterns to energies. 
However, these representations also implicitly encode structural patterns (commonly observed coordination environments) and their correlations with each other and with abstract properties. 
A distinct and powerful pre-training paradigm moves away from real-world data entirely, instead training models like \modelname{TabPFN} on millions of synthetically generated datasets to become general-purpose learning algorithms (see \Cref{sec:dataset_creation} about dataset creation). 
This allows them to perform in-context learning on new, small datasets during a single inference call, often outperforming traditional methods. \autocite{hollmann2025accurate} \\

\noindent The core principle remains: \emph{learning on large datasets to build generalizable internal representations before task-specific fine-tuning.}

\subsection{Fine-Tuning: Learning the Coloring of Data}
\label{sec:fine_tuning_coloring}

While pre-training enables models to learn general structural representations of chemical data, fine-tuning refines these representations for specific downstream tasks.
If pre-training can be conceptualized as learning the \enquote{structure} of chemical knowledge, fine-tuning can be viewed as learning to \enquote{color} this structure with task-specific knowledge and capabilities (see \Cref{fig:training_workflow}). 
This specialization process transforms general-purpose internal representations into powerful task-specific predictors while retaining the foundational knowledge acquired during pre-training.

Fine-tuning adapts pre-trained model parameters through training on domain-specific datasets. 
This typically requires substantially less data than pre-training. To make this process even more efficient, a common strategy is to \enquote{freeze} the majority of the model’s layers and only train a small subset of the final layers (see \Cref{sec:peft}).
Fine-tuning is particularly valuable in chemistry, where datasets are often limited in size.
Traditionally, addressing chemistry-specific problems required heavily engineered and specialized algorithms that directly incorporated chemical knowledge into model architectures.  
However, fine-tuned \glspl{llm}, for example, have shown comparable or superior performance to these specialized techniques, particularly when data is limited \autocite{jablonka2024leveraging}. 
The efficiency of fine-tuning stems from the transferability of chemical knowledge embedded during pre-training, where the model has already learned to spot patterns in molecular structure, reactivity, and chemical terminology sequences. 

\subsection{Post-Supervised Adaptation: Learning to Align and Shape Behavior} \label{sec:rl}

Pre-training and fine-tuning equip the model with a learned distribution, which represents its knowledge about what outputs are plausible or likely. 
Post-training biases this distribution towards preferred outcomes---such as task-specific goals. 
The new, desired behavior of the model (called the policy, $\pi$, in \gls{rl}) comes from this refined distribution.
This shift has a subtle but crucial effect on the internal representations. 

Post-training alignment workflows commonly use \gls{rl}, as the classic loss-minimization approaches---simply fine-tuning on more \enquote{correct} examples---can struggle to capture more nuanced, hard-to-label objectives.\autocite{Huan2025mathLLM} When the goal is to steer the model toward more intangible qualities, formulating loss functions and collecting a pre-labeled dataset become very challenging. 
In \gls{rl}-based alignment, the model is treated as an agent that takes actions (generates text in the case of an \gls{llm}) in a trial-and-error environment and receives a reward signal based on the actions it chooses. The \gls{rl} objective is to maximize this reward by changing the model's behavior. 
In the case of \gls{llm}, this means compelling it to generate text with the preferred properties. This process transforms the model into a goal-oriented one, where the goal can be to generate stable molecules, solve tasks step-by-step, or utilize tools, depending on the reward function.

During alignment, the foundational embeddings for basic concepts (e.g., a carbon atom) learned during pre-training remain largely intact. 
This initial state is critical; without a robust, pre-trained \gls{llm}, the \gls{rl} process would be forced to blindly explore an intractably vast space, making it highly unlikely to discover preferred sequences (that it could then reinforce).   
 
The mapping from an input to its final representation is adjusted to become \enquote{reward-aware}. For example, the representation of a molecule might now encode not just its chemical structure, but also its potential to become a high-reward final molecule (stable and soluble molecule) \autocite{narayanan2025training}. 

\begin{tcolorbox}[
    title=Reinforcement Learning Framework for \glspl{gpm},
    breakable,
    enhanced,
    colback=white,
    colframe=black!30,
    colbacktitle=black!10,
    coltitle=black,
    fonttitle=\bfseries,
    boxrule=0.5mm,
    left=2mm,
    right=2mm,
    top=6mm,
    bottom=4mm,
    middle=12mm
]
\label{eq:rl}

\begin{center}
\begin{minipage}{0.9\linewidth}
\vspace{1.5em}

\begin{equation}
\label{eq:rl_objective}
\hspace*{-1em}
\pi(a|s) = \eqnmarkbox[PolicyColor]{policy}{P_{\text{LLM}}} \left( \eqnmarkbox[ActionColor]{action}{\text{next tokens}} \mid \eqnmarkbox[StateColor]{context}{\text{context}} \right) 
\rightarrow \text{Maximize } \eqnmarkbox[RewardColor]{reward}{\mathbb{E}_{\tau \sim \pi_{\theta}}[R(\tau)]}
\end{equation}

\vspace{1.5em}

\annotate[yshift=1.5em, xshift=-1em]{above,right}{context}{State}
\annotate[yshift=-1.5em, xshift=1em]{below,right}{action}{Action}
\annotate[yshift=1.5em]{above}{policy}{Policy (\gls{llm})}
\annotate[yshift=-1.5em]{below}{reward}{Expected Reward}

\vspace{0.5em}

\end{minipage}
\end{center}

\tcblower

\begin{itemize}
\item \textbf{State ($s$)} ({\color{StateColor}Blue}): The sequence of tokens generated so far, including the original prompt and any partial response. Represents the current context that the model uses to make decisions.

\item \textbf{Action ($a$)} ({\color{ActionColor}Maroon}): The next token that the model chooses to generate from its vocabulary. This is the discrete decision the agent makes at each step.

\item \textbf{Policy ($\pi$ or $P_{\text{LLM}}$)} ({\color{PolicyColor}Red}): The \gls{llm} itself, whose parameters define the probability distribution over possible following tokens given the current state. This is what gets optimized during training.

\item \textbf{Reward ($R$)} ({\color{RewardColor}Peach}): A numerical score assigned to complete generated sequences, measuring how well the output achieves the desired goal. Used only during training to guide parameter updates.


\item \textbf{Expectation ($\mathbb{E}_{\tau \sim \pi_{\theta}}[R(\tau)]$)} ({\color{RewardColor}Peach}): The average reward, calculated over many possible sequences $\tau$ generated by policy $\pi_{\theta}$. The subscript $\tau \sim \pi_{\theta}$ indicates that the expectation is taken over the distribution of sequences produced by the current policy. Training adjusts $\theta$ to maximize this expectation, enabling the generation of high-quality outputs.
\end{itemize}

\end{tcolorbox}

\paragraph{The Challenge of Reward Design}
A critical factor for the success of this framework is the design of the reward function. The training process is most stable and effective when rewards are \textit{verifiable} and based on objective, computable metrics. 
In contrast, training with \textit{sparse} rewards (where feedback is infrequent) or \textit{fuzzy} signals (where the goal is subjective or ill-defined) makes the credit assignment problem significantly more difficult. 
This is a central challenge in aligning models with complex human preferences, as crafting precise reward functions that capture the full nuance of a desired behavior remains an active area of research \autocite{ouyang2022training}.

\paragraph{The \gls{llm} as a Policy}
When using a \gls{llm} as the agent in \gls{rl}, the policy (see \Cref{eq:rl_objective}) is the \gls{llm} itself. 
Consider teaching a model to design multi-step synthetic routes for pharmaceutical compounds, using a retrosynthetic strategy. 
The \textbf{state} ($s$) represents the synthetic plan generated so far. Initially, the state consists of just the target molecule but evolves to include each proposed step in the route. Each \textbf{action} ($a$) is the next retrosynthetic decision---for example, which bonds to break or what reagents to use. The \gls{llm} serves as the policy ($\pi$), using its parameters to determine the probability of choosing different possible actions given the current context. 
To put it mathematically, this would be $\pi(a|s) = P_{\text{LLM}}(\text{next synthetic step}|\text{current plan})$ (see \Cref{eq:rl_objective}). 
The model leverages its chemical knowledge to identify the most promising decisions. 
The \emph{reward} ($R$) scores the completed retrosynthetic route based on practical criteria that could be the number of steps, predicted yield, reagent cost, etc. 
This score can directly come from the feedback of real chemists (\gls{rlhf}), or from a small model trained to predict human preference scores or pre-defined criteria (see another example \ref{example:rl} of optimizing solar cells with \gls{rl} ). 

Theoretical work in reinforcement learning has shown that the complexity of such problems scales quadratically with the size of the action space \autocite{dann2015sample}. 
At each step, the model must choose from tens of thousands of possible tokens, and the number of possible sequences (and therefore actions) grows exponentially. 
Without pre-training, this would make the learning process computationally prohibitive. 
Pre-training provides a strong initialization that effectively constrains the action space to reasonable chemical language and valid synthetic steps, dramatically reducing the exploration requirements (see how pre-training creates a structured manifold in \Cref{fig:training_workflow}).

Recent developments have revealed that \gls{rl} training can elicit reasoning capabilities that were previously thought to require explicit programming or extensive domain-specific architectures. 
Models trained with \gls{rl} demonstrate the ability to decompose complex problems, perform backtracking when approaches fail, and engage in multi-step planning without being explicitly taught these strategies. \autocite{xu2025towards}

\begin{examplebox}[label={example:rl}]{Optimizing solar cells}

\textbf{Goal:} Design perovskites that are both stable (high thermal/phase stability) and high-performing (optimal bandgap for light absorption).

\textbf{State:} Current partial perovskite formula (e.g., \ce{ABX3} lattice with partial cation/anion substitutions). \\

\textbf{Action:} Substitute the next ion or additive (e.g., replace \ce{A}-site with \ce{Cs+} or add \ce{Cl-} dopant). \\

\textbf{Reward:}
\begin{itemize}
    \item \texttt{+1} for stability (e.g., based on Goldschmidt tolerance factor).
    \item \texttt{+1} for bandgap tuning (\texttt{1.5 to 1.8 eV} ideal for single-junction cells).
    \item \texttt{-1} for defect proneness (e.g., halide migration).
\end{itemize}

\textbf{Episode Example:}

\begin{itemize}
    \item Start with \ce{PbI2} scaffold (\ce{B} and \ce{X} sites) $\rightarrow$ add \ce{MA+} (methylammonium) at \ce{A}-site $\rightarrow$ Reward: \texttt{+0.5} (good initial bandgap but low stability due to organic volatility).
    \item Co-substitute with \ce{Cs+} and \ce{FA+} (formamidinium) $\rightarrow$ Reward: \texttt{+1.7} (improved tolerance factor \textgreater{}\texttt{0.9}, stable mixed-cation perovskite with approx. ~ \texttt{1.6 eV} bandgap).
    \item Over-dope with excess \ce{Br-} $\rightarrow$ Reward: \texttt{-0.7} (widens bandgap too much \textgreater{}\texttt{1.8 eV}, reducing PCE; also increases defects).
\end{itemize}

The model learns to generate perovskite compositions that maximize cumulative reward, naturally discovering stable, high-PCE materials in the vast compositional space for next-gen solar cells. Note, however, that attempting this or a problem would require designing reward functions that can score actions taken by the agent.

\end{examplebox}

\paragraph{Updating the LLM Policy} 
After the model takes actions (generates a sequence of tokens), the reward it receives for the chosen actions is used to update the \glspl{llm} parameters using an \gls{rl} algorithm, such as \gls{ppo} \autocite{schulman2017proximal}.
\gls{ppo} works by encouraging the model to favor actions (outputs) that lead to higher rewards, but it also includes a mechanism to constrain how much the model's behavior can change in a single update by introducing s a penalty term that discourages the \glspl{llm} policy from deviating too far from its original, pre-trained distribution. 
This ensures the model does not \enquote{forget} its foundational knowledge about language or chemistry while it is learning to pursue the reward, thus biasing the distribution rather than completely overwriting it. 
The result is a controlled shift: the model becomes more aligned without losing what it already knows.

\paragraph{Inference and Sampling from the Adapted Model}

The \gls{rl} training process permanently updates the weights of the \gls{llm}. 
When we sample from this model, we are drawing from this new, biased distribution. 
For a given context (state), the probabilities for tokens (actions) that were historically part of high-reward sequences are now intrinsically higher. 
At the same time, pathways that led to low rewards are suppressed. 
The model is now inherently more likely to generate outputs that align with the preferences and goals encoded in the reward function.

\subsection{Example Architectures} \label{sec:example_architectures}
While much effort is currently invested in building foundation models based on transformer-based \glspl{llm}, the foundation model paradigm is not limited to this model class.

\begin{figure}[H]
    \centering
    \includegraphics[width=1\textwidth]{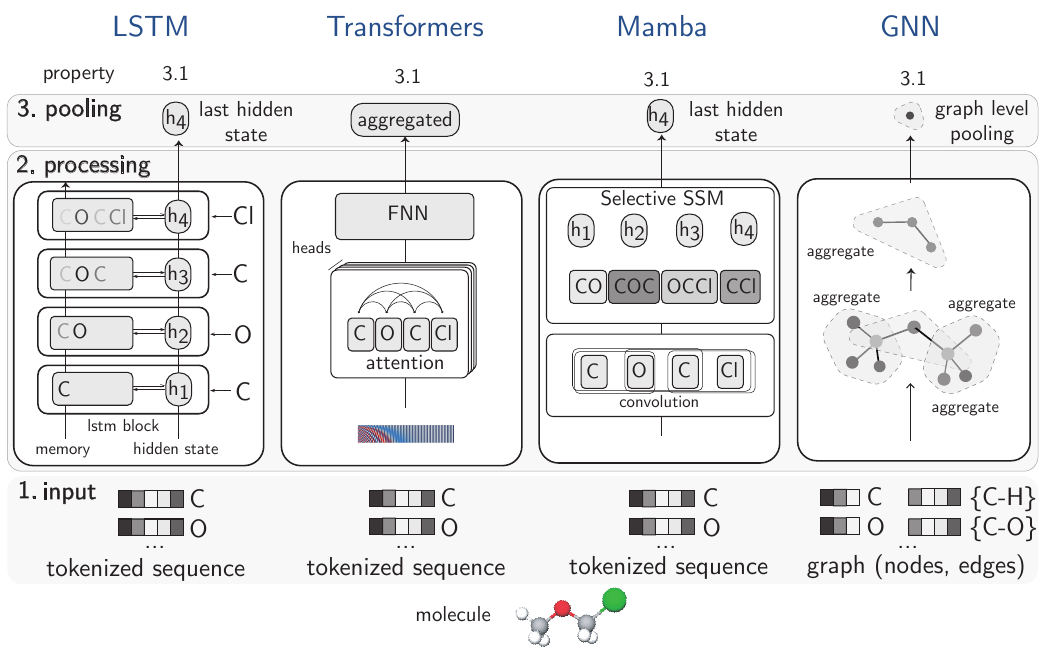}
    \caption{\textbf{Blueprint for \glspl{gpm} architectures}. This diagram illustrates four distinct neural network architectures (\gls{lstm}, Transformer, Mamba, and \gls{gnn}), highlighting their unique approaches to input representation, information processing, and output pooling. \glspl{lstm} sequentially process tokens, accumulating information in a final hidden state with an inductive bias towards sequential arrangement. Transformers, conversely, use multi-head attention and positional encodings to capture global interactions simultaneously, offering minimal inductive bias but enabling rich contextual understanding. Mamba combines local convolutional processing with selective state space modeling to efficiently focus on chemically relevant parts, also typically using a final hidden state. \glspl{gnn} leverage the inherent graph structure of molecules, where atoms are nodes and bonds are edges, to model local chemical environments through message passing, followed by graph-level pooling to create a unified representation. Each approach offers unique strengths in how it captures molecular features, ranging from sequential to global, and graph-based relationships. Notably, despite their differences, all architectures follow a common three-stage pipeline. Tokenization of molecular inputs, iterative representation updates via their specific mechanisms, and pooling to produce a unified molecular representation.}
    \label{fig:architectures}
\end{figure}

In the chemical domain, where heterogeneous data such as \gls{smiles} and graphs for molecular structures prevail, the use of a diverse array of architectures is expected.
The architectures shown in \Cref{fig:architectures} are examples of foundational backbones that we discuss in the following sections.

\paragraph{\gls{lstm}}

\gls{lstm} networks \autocite{hochreiter1997long} are well-suited for processing sequential data, such as text or time series. 
\Cref{fig:architectures} illustrates how chemical information is processed to predictions in \glspl{lstm}.
\begin{itemize}
    \item \textbf{Input}: Molecules are represented as tokenized sequences (e.g., \gls{smiles} strings like \enquote{COCCl}), processed one token at a time. Each token corresponds to an atom.
    \item \textbf{Processing}: Information flows sequentially through \gls{lstm} blocks where each hidden state ($h_{1}$, $h_{2}$, $h_{3}$, $h_{4}$) accumulates information about the molecule. The memory cell maintains chemical context through gating mechanisms. The inductive bias is sequential processing---assuming chemical properties emerge from analyzing tokens in order.
    \item \textbf{Pooling}: The final hidden state ($h_{4}$) captures the entire molecular information after processing the complete sequence. This last state serves as the molecular representation for the downstream task.
\end{itemize}

\glspl{lstm} process information in a strict sequence.  
For the model to connect the first word to the last, that information must pass through every single step in between. 
The computational cost of \enquote{talking} across the sequence grows with the sequence length. 
Furthermore, the entire history of the sequence must be compressed into a single, fixed-size hidden state.

An \gls{xlstm} overcomes this with two key changes. \gls{xlstm} uses enhanced gates (act like filters to control what information flows) to precisely revise its memory. Second, instead of a single memory bottleneck, it uses a parallel \enquote{matrix memory}. This provides multiple \enquote{slots} to store different pieces of information at the same time. This structure allows it to process information in parallel, making it much more efficient.
\modelname{Bio-xLSTM} adapts this architecture for biological and chemical sequences, demonstrating proficiency in generative tasks and in-context learning for DNA, proteins, and small molecules.\autocite{SchmidingerSSSH25}

\paragraph{Transformer}

Transformers \autocite{vaswani2017attention} are also designed for sequential data, but are particularly powerful in capturing long-range dependencies and rich contextual relationships within sequences. 
Their core \enquote{attention mechanism} allows them to weigh the importance of different parts of the input simultaneously (quadratic computational scaling---if you double the length of the sequence, the amount of work the model needs to do quadruples). 
Effectively, they can be thought of as a fully connected graph model,\autocite{velivckovic2023everything, joshi2025transformers} where each representation of a token is connected to every other token and can impact its representation.

\begin{itemize}
    \item \textbf{Input}: Similar to \glspl{lstm}, data is tokenized and often enhanced with positional encodings (see \Cref{fig:architectures}, the tokenized sequence is added with positional information, e.g., using a sinusoidal signal---the red-blue spectrum) to maintain information about where in a sequence a token is placed (the attention mechanism itself does not preserve sequence order information).
    \item \textbf{Processing}:  Uses attention mechanisms, where every atom/token attends to every other token simultaneously. This enables the capture of long-range interactions between distant elements of the sequence, regardless of their sequential distance. The \gls{fnn} transforms these attention-weighted representations.  To get a more robust and comprehensive understanding of the relationships within a sequence, models don't just rely on a single way of \enquote{paying attention}.  Instead, they employ multiple independent \enquote{attention heads} known as multi-head attention.
    \item \textbf{Pooling}:  Uses an aggregated representation or special token that combines information from all tokens, enabling global molecular property prediction.
\end{itemize}

\paragraph{Mamba}
Mamba\autocite{gu2023mamba0} is designed to be highly efficient (linear computational scaling with respect to sequence length) and effective at modeling very long sequences, offering a potentially more scalable alternative to Transformers for certain sequential tasks (for example, modelling very long protein sequences or polymer chains, while retaining strong performance in capturing dependencies.)

\begin{itemize}
    \item \textbf{Input}:  Sequences similar to \gls{lstm}.
    \item \textbf{Processing}:  First applies convolution to capture local contexts, creating representations that incorporate neighboring information. These contextualized tokens are then processed through a \gls{ssm}. An \gls{ssm} is a type of sequence model that efficiently captures and summarizes long-range dependencies by tracking an evolving internal \enquote{state} (evolving representation of all the relevant information) based on inputs. This \gls{ssm} dynamically focuses on relevant parts. The inductive bias combines local patterns (through convolution) with efficient selective attention for handling long-range dependencies.
    \item \textbf{Pooling}:  Uses the final hidden state ($h_{4}$) similar to \gls{lstm}, but this state contains selectively processed information that more efficiently captures important features.
\end{itemize}
This architectural approach has been successfully applied to chemical foundation models, demonstrating \gls{sota} results in tasks like molecular property prediction and generation while maintaining fast inference on a large dataset of \gls{smiles} samples.\autocite{soares2025mamba-based}

\paragraph{\gls{gnn}}
\gls{gnn} is an architecture that complements graph representations (see the section discussing graph-based representation \Cref{sec:common_representations}). Molecules are represented as graphs. \glspl{gnn} then operate on them by processing node and edge representations. Based on how the nodes are connected through edges, the information in these representations is updated multiple times. This procedure is called message passing (see \Cref{fig:architectures}). Information from neighbors is aggregated, and this aggregation occurs for all nodes and sometimes also for edges.

\begin{itemize}
    \item \textbf{Input}: Graphs, which are collections of nodes (e.g., atoms) and edges (e.g., bonds).
    \item \textbf{Processing}: Uses message passing through multiple aggregation steps (message would be the information in node or edge at the current stage, and aggregation can be different types of operations like adding information, taking mean, etc, depending on the architecture choice). Each node updates its representation based on messages from its bonded neighbors. The inductive bias is the graph structure itself, which naturally aligns with chemical bonding patterns.
    \item \textbf{Pooling}: Graph-level pooling (e.g., taking the mean of all node representations) aggregates information from all atoms and bonds to create a unified molecular representation, respecting the molecular graph structure.
\end{itemize}

\noindent These architectures cannot solve all problems equally well because they are tailored to different data structures. 
\gls{lstm} and Mamba inherently excel at processing sequential data; Transformers are powerful at capturing global relationships across the entire input,
whereas \glspl{gnn} are designed for graph-structured information. 
Forcing one type to handle data optimally it was not intended for, often leads to suboptimal performance, inefficiency, or requires extensive, task-specific adaptations that dilute its \enquote{general-purpose} nature.\autocite{alampara2024mattext}

\subsection{Multimodality}
Multimodal capabilities enable systems to process and understand multiple types of data simultaneously. 
Unlike traditional unimodal models, which work with a single data type (e.g., text-only or image-only), multimodal models can integrate and reason across different modalities, such as text, images, molecular structures, and spectroscopic data.

The core principle behind multimodal models lies in learning shared representations across different data types.  The challenge of creating this shared representation can be addressed through several architectural strategies, each with a different approach to learning the joint distribution of multimodal data. One dominant strategy is joint embedding alignment, where separate, specialized encoders are used for each modality (e.g., a \gls{gnn} for molecular structures and a Transformer for text). These encoders independently map their respective inputs into their own high-dimensional vector spaces. 
The key learning objective, often driven by contrastive learning (see \Cref{sec:contrastive_learning}), is to align these separate spaces.

Another common approach is input-level fusion, where different data types are tokenized into a common format and fed into a single, unified architecture. 
For instance, a molecular structure might be converted into a \gls{smiles} string, an image into a sequence of patches, and text into its standard tokens. These disparate token sequences are then concatenated and processed by a single large model, typically a Transformer.\autocite{xu2025qwen3} 
The model's attention mechanism can learn correlations between modalities, e.g., an image patch can \enquote{attend} to a word in the description.
A more recent and highly efficient variant is adapter-based integration, where a powerful, pre-trained unimodal model (models that take a single type of representation) (like an \gls{llm}) is frozen, and a small \enquote{adapter network} (see discussion about adapter in \Cref{sec:model_adaptation}) is trained to project the embeddings from a secondary modality (e.g., a molecule) into the \gls{llm}'s existing latent space.\autocite{liu2023multi} 
This adapter effectively learns to translate the new data type into the \gls{llm}'s native \enquote{language}, leveraging the \gls{llm}'s vast pre-existing knowledge without the need for complete re-training. 
For instance, a model might learn that the textual description \enquote{benzene ring} corresponds to a specific visual pattern in molecular diagrams and produces characteristic peaks in \gls{nmr} spectroscopy. 
This cross-modal understanding enables more comprehensive and contextually rich analysis than any single modality alone could provide.

\subsubsection{Multimodal Integration in Chemistry}\label{sec:multimodal_chem}

A molecule’s \gls{smiles} string alone might not reveal its 3D conformational preferences. 
A spectrum alone could suggest many molecular structures. 
However, coupling these modalities with textual knowledge (e.g., \enquote{the sample was prepared by X method}) could narrow down possibilities. 
Multimodal models have the potential to emulate a human expert who simultaneously considers spectral patterns, chemical rules, and prior knowledge to deduce a structure (see an example \ref{example:multimodal} of leveraging multimodal data for identifying the chemical). 
Another motivation is to create generalist \gls{ai} models. 
Instead of having multiple independent models---one for spectral analysis, another for molecule property prediction, and another for text mining---a single model could handle diverse tasks by understanding multiple data types.
In this way, a researcher can ask a question in natural language, provide a molecule (in the form of a structure file or image) as context, and receive a helpful answer that leverages both structural and textual knowledge.

\begin{examplebox}[label={example:multimodal}]{Multimodal reasoning for chemical identification}
A chemist finds an unknown white powder in a food sample that is highly soluble in water and tastes sweet. Chemist wants to identify it: \\
\textbf{Input Modalities:}
\begin{itemize}
    \item \textbf{$^{13}$C-NMR spectrum:} Shows peaks in the 60--100 ppm region, characteristic of carbons bonded to oxygen.
    
    \item \textbf{$^{1}$H-NMR spectrum:} Multiple peaks in the 3--5 ppm region with complex splitting patterns, indicating protons on oxygen-bearing carbons.
    
    \item \textbf{IR spectrum:} Broad, intense peak at 3000--3500 cm$^{-1}$ (O-H stretching), peaks around 2900 cm$^{-1}$ (C-H stretching), strong peaks at 1000--1150 cm$^{-1}$ (C-O stretching, ether and alcohol).
    
    \item \textbf{Mass spectrometry:} Molecular ion peak at \texttt{m/z = 180}.
    
    \item \textbf{Text description:} \enquote{White crystalline solid, sweet taste, highly soluble in water, found in food sample}.
    
\end{itemize}
\textbf{Multimodal Model Reasoning:}
\begin{itemize}
    \item \textbf{MS + IR $\rightarrow$} Molecular formula \ce{C6H12O6} with multiple hydroxyl groups (polyol).
    \item \textbf{NMR $\rightarrow$} Chemical shifts and splitting patterns consistent with a cyclic structure containing multiple oxygen-bearing carbons; presence of anomeric signals suggests equilibrium between ring forms.
    \item \textbf{Text + Chemical Knowledge $\rightarrow$} Sweet taste and high water solubility point to a simple sugar (monosaccharide).
\end{itemize}
Based on the evidence, the model hypothesizes the molecule to be D-glucose. Single modality would be insufficient---mass spectrometry alone could match any hexose isomer (fructose, galactose, etc.), but combining spectroscopic fingerprints with physical properties enables confident identification.
\end{examplebox}

\modelname{MolT5} \autocite{edwards2022translation} adapted the \modelname{T5} transformer for chemical language by training on scientific text and \gls{smiles} strings, using a masking objective to reconstruct masked segments. 
This approach treats \gls{smiles} as a \enquote{language}, enabling \modelname{MolT5}  to generate both valid molecules and fluent text. 
Similarly, \modelname{Galactica} \autocite{taylor2022galactica}, an \gls{llm}, also incorporated \gls{smiles} into its training. Later, the \modelname{MolXPT}\autocite{liu2023molxpt0} model used \enquote{paired} examples (\gls{smiles} and textual description) by replacing chemical names in scientific texts with their corresponding \gls{smiles} strings and description. 
This pre-training approach enables \modelname{MolXPT} to learn the context of molecules within text and achieve zero-shot text-to-molecule generation (see \Cref{sec:mol_generation} for more details on this application).

Contrastive learning emerged as an alternative, aligning separate text and molecule encoders in a shared embedding space (see \Cref{sec:contrastive_learning} for the principle of learning). \modelname{MoleculeSTM} \autocite{liu2023multi} aligns separate text and molecule encoders in a shared space using paired data. 
This dual-encoder approach enables tasks such as retrieving molecules from text queries and shows strong zero-shot generalization for chemical concepts.  
\modelname{CLOOME} \autocite{sanchez2023cloome}  used contrastive learning to embed bioimaging data (microscopy images of cell assays) and chemical structures of small molecules into a shared space. 
Multimodal learning also enables the determination of molecular structure from spectroscopic data. 
Models trained on large datasets of simulated spectra \autocite{alberts2024unraveling}, which combine multiple spectral inputs, could accurately translate spectra into molecular structures. \autocite{chacko2024spectro,mirza2024elucidating}

Beyond prediction, some multimodal models aim for cross-modal generation, creating one type of data from another (e.g., generating an \gls{ir} spectrum from a molecular structure). \textcite{takeda2023multi} developed a multimodal foundation model for materials design, integrating \gls{selfies} strings, \gls{dft} properties, and optical absorption spectra. 
Their approach involves encoding each type of data separately into a shared, compressed representation space. 
Then, a network learns to combine these compressed representations to understand the connections between them. 
This pre-training on a big dataset of samples enables both combined representations (joint embeddings summarizing all modalities) and cross-modal generation, allowing tasks like predicting a spectrum from a molecule or generating a molecule from desired properties. However, inverse generation remains challenging with lower accuracy than forward prediction, generated outputs often require experimental validation, and performance degrades for out-of-distribution molecules.

A more recent approach is the integration of molecular encoders with pre-trained \glspl{llm}. 
Models like \modelname{InstructMol} \autocite{cao2023instructmol0} and \modelname{ChemVLM} \autocite{li2024seeing} use an \enquote{adapter} (see discussion about \modelname{LoRa} in \Cref{sec:model_adaptation}) to project molecular information into the \gls{llm}'s existing knowledge space. 
This two-stage process first projects molecule representations into the \gls{llm}'s token space through pre-training on molecule-description pairs. 
Subsequently, instruction tuning on diverse chemistry tasks (e.g., \gls{qa}, reaction reasoning) enables the \gls{llm} to leverage molecular inputs, significantly enhancing its performance on chemistry-specific problems.

The latest generation of \glspl{gpm} is often natively multimodal, designed from the ground up to process text, images, and other data types seamlessly. 
Natively multimodal systems are characterized by a single, unified neural network trained end-to-end on a diverse range of data modalities. 
In the scientific domain, natively multimodal systems are still being explored. However, evaluations suggest that these models are not yet robust for solving complex scientific research tasks.\autocite{alampara2024probing}

\subsection{Optimizations}

As \glspl{gpm} continue to grow in size and complexity, optimization techniques (performance or resource consumptions optimization) become critical for making these models practically deployable while maintaining their accuracy. 
This section discusses three key optimization approaches that have particular promise for chemistry foundation models: \gls{moe} architectures for efficient scaling, quantization, and mixed precision for memory and computational efficiency, and knowledge distillation for creating specialized, lightweight models.

\subsubsection{Mixture-of-Experts} \label{sec:arch-moes}
\gls{moe} is a neural network architecture that uses multiple specialized \enquote{expert} networks instead of one single, monolithic model. The core idea is to divide the vast problem space---the embedding space of all possible inputs---into more manageable, homogeneous regions. A region is considered \enquote{homogeneous} not because all inputs within it are identical, but because they share similar characteristics and can be processed using a consistent set of rules. For instance, in a chemistry model, one expert might specialize in organic molecules, while another focuses on inorganic crystals; each expert sees a more consistent, or homogeneous, set of problems. This division of labor is managed by a gating network, which acts like a smart dispatcher. This gating network is itself a small neural network, often referred to as a trainable router, because it learns during training how best to route the data to the most appropriate expert, thereby improving its decisions over time.
\gls{moe} models achieve efficiency through selectively activating only the specific experts needed for a given task, rather than activating the entire neural network for every task. Modern transformer models using \gls{moe} layers can scale to billions of parameters while maintaining manageable computational costs, as demonstrated by models like \modelname{Mixtral-8x7B}\autocite{jiang2024mixtral}, which uses eight experts with sparsity.

\textcite{shazeer2017outrageously} demonstrated that using a sparsely-gated \gls{moe} layer can expand a network’s capacity (by over 1000 times) with only minor increases in computation. In this architecture, each expert is typically a \gls{fnn}, and a trainable router determines which tokens are sent to which experts, allowing only a subset of the total parameters to be active for any given input. 
 
An \gls{llm} for science with \gls{moe} architecture (\modelname{SciDFM} \autocite{sun2024scidfm}) shows that the results of expert selection vary with data from different disciplines, i.e., activating distinct experts for chemistry vs.\ other disciplines. 
They consist of multiple \enquote{expert} subnetworks, each potentially specializing in different facets of chemical knowledge or types of chemical tasks. A routing mechanism directs inputs to the most relevant expert(s). This allows the foundation model to be more adaptable and perform across the broad chemical landscape.  

Extending this concept, a recent multi-view \gls{moe} model (\modelname{Mol-MVMoE} \autocite{shirasuna2024multi}) treats entire, distinct chemical models as individual \enquote{experts}. Rather than routing tokens within one large model, a gating network learns to create a combined molecular representation by dynamically weighting the embeddings from each expert model. This method showed strong performance on \modelname{MoleculeNet}, a widely used benchmark suite for molecular property prediction, outperforming competitors on 9 of 11 tasks.

Training \gls{moe} models can be a complex process. The gating mechanism must be carefully learned to balance expert usage and instability, or some experts may end up underutilized (most of the data would be processed by a subset of networks). \autocite {fedus2022switch} For chemistry tasks, an additional challenge is to ensure that each expert has access to sufficient relevant chemical data to specialize. If the data is sparse, some experts may not learn meaningful functions. Despite these hurdles, \gls{moe} remains a promising optimization strategy to handle the breadth of chemical space.

\subsubsection{Quantization and Mixed Precision}
\label{sec:quantization}
Quantization is a technique for making models more computationally efficient by reducing their numerical precision. 
In experimental science, precision often relates to the number of significant figures in a measurement; a highly precise value, such as $3.14159$, carries more information than a rounded one, like $3.14$. 
Similarly, a model's knowledge is stored in its weights, which are organized into large matrices of numbers. 
Standard models typically use high-precision formats, such as 32-bit floating-point, which can represent a wide range of numbers with many decimal places to store weights. During inference, these weight matrices are multiplied by the input data to produce a prediction. Quantization involves converting these numbers into a lower-precision format, such as 8-bit integers, which are whole numbers with a much smaller range. This process is similar to rounding experimental data---it simplifies the numbers, uses less memory, and allows calculations to run much faster.

\textcite{dettmers2022gpt3} introduced an 8-bit inference approach (\texttt{LLM.int8}) enabling models as large as \modelname{GPT-3} (175B parameters) to run with no loss in predictive performance (less than $50\%$ GPU-memory usage).  
A key insight in this paper is that while most numbers in a model can be safely rounded, a few \enquote{outlier} values with large magnitudes are critical for performance. 

A different, yet related, strategy is mixed-precision quantization.\autocite{micikevicius2017mixed} Instead of applying a single precision format (like 8-bit) across the entire model, this approach uses a mix of different precisions for different parts of the network. The guiding principle is that some layers of the model might be more sensitive to rounding errors than others.

Many chemistry applications, particularly in automated laboratory setups, require deployment on edge devices---local computing hardware, such as the controllers for robotic arms or the onboard computers in analytical instruments---or cloud platforms with limited computational resources. Quantization can be a valuable optimization tool for reducing computational burden while increasing inference speed, which is crucial for real-time applications.

\subsubsection{Parameter-Efficient Tuning}
\label{sec:peft}

While full fine-tuning is computationally expensive, memory-intensive, and results in a complete, multi-gigabyte copy of the model for every new task. 
\gls{peft} methods offer a solution to this problem by freezing the vast majority of the trained model's weights and only training a few new parameters.

A prominent and widely used \gls{peft} technique is \gls{lora}.\autocite{hu2022lora}  
The key insight of \gls{lora} is that the change needed to adapt a pre-trained weight matrix for a new task can be approximated effectively using much smaller matrices. \gls{lora} freezes the original model weights and introduces small trainable rank-decomposition matrices into each transformer layer, significantly reducing the number of trainable parameters. 
Because these new matrices contain far fewer parameters---often less than 0.1\% of the original model---the computational and memory requirements for training are drastically reduced.
 
These optimization strategies can be combined with quantization (see \Cref{sec:quantization}) for even greater efficiency. \textcite{dettmers2023qlora} introduced \gls{qlora}. 
In this approach, the large pre-trained model is first quantized down to a very low precision (typically 4-bit), dramatically shrinking its memory footprint. 
Then, the lightweight \gls{lora} adapters are added and fine-tuned. \gls{qlora} enables the fine-tuning of massive models---such as a 70-billion-parameter model---on a single, consumer-grade GPU.

\subsubsection{Distillation} \label{sec:distillation}
Knowledge distillation is a technique that aims to transfer the learning of a large pre-trained model (the \enquote{teacher model}) to a smaller \enquote{student model}. \autocite{hinton2015distilling}
The computationally more efficient \enquote{student model} is trained to mimic the behavior (e.g., output probabilities or internal representations) of the larger teacher model. 
This allows the rich, nuanced understanding learned by the large foundation model to be compressed into a more compact and faster student model.\autocite{sanh2019distilbert}

Effective distillation requires that the teacher model is both competent at the task and that its knowledge is representable by the student. 
If the teacher is too large or complex compared to the student, the student may struggle to emulate it, leading to degraded performance. \autocite{liu2024wisdom}

\subsection{Model Level Adaptation}
\label{sec:model_adaptation}
Although promising, as shown in \Cref{tab:model_adaptation}, \glspl{gpm} such as \glspl{llm} rarely work straight out of the box for specialized tasks and often need customization. This is especially true for scientific problems where data is a limiting factor. 
By prompting an \gls{llm}---for example, by asking a question or giving instructions---one can observe that these models perform much better on general tasks than on those related to chemistry. 
This difference arises because \glspl{llm} are not typically trained on domain-specific chemical tasks and therefore they lack the necessary knowledge and reasoning skills. 

To bridge this gap, two complementary families of approaches exist. The first approach involves adapting the model's knowledge or behavior directly. The simplest method is to embed information directly in the prompt, for instance by providing examples (\gls{icl}) \autocite{brown2020language} or by introducing intermediate reasoning steps (\gls{cot}) \autocite{wei2022chain}. However, not all problems can be solved in this way, and sometimes it is necessary to tune the model to new data, which updates its parameters. 
The second approach involves coupling the model into a larger system that can interact with external sources of information and tools.

\begin{table}[!ht]
    \centering
    \caption{\textbf{Model Adaptation Approaches Overview:} This table provides a rough overview of the estimated time, data, required \gls{ml} knowledge and estimated energy cost for each approach. Each method (listed in the first column) is paired with an approximate implementation time, the estimated dataset size, the level of \gls{ml} expertise needed and estimated energy cost.  These estimates assume that you have at least a bachelor's level of understanding of chemistry and at least some computational background. The energy costs are estimated by simulations \cite{ozcan2025quantifying, luccioni2024light} accounting for the number of model calls, architecture, model size and input length.}
\begin{tabular}{lllll}
\toprule
    \textbf{Model adaptation}  & \textbf{Time}    & \textbf{Data}   & \textbf{ML knowledge} & 
    \textbf{Energy Cost} \\ 
\midrule
Pre-training       & Weeks   & 1M--1B+      & Very High & 50MWh - 1GWh  \\
Zero-shot prompting & Minutes & None        & None & \textless 10Wh \\
Few-shot Prompting   & Hours   & \textless 10 & None  & \textless 1kWh  \\
Fine-tuning         & Days    & \textless 10k         & High     &   100kWh - 50MWh  \\
\toprule
    \textbf{Coupling into systems} \\ 
\midrule
RAG                & Days    & 100k--1M+    & Low     &   \textless 1kWh   \\
Tool-Augmentation   & Days    & None / 10k+ & Low     &  \textless 1kWh  \\
\bottomrule
\end{tabular}
    \label{tab:model_adaptation}
\end{table}

\paragraph{Prompting}\label{sec:prompting} \glspl{llm} have demonstrated the ability to perform a wide range of tasks based solely on prompt instructions---without the need for fine-tuning \autocite{radford2019language}. 
This ability, for \glspl{llm} to complete tasks without any additional information, is often referred to as zero-shot prompting.
By providing task-specific examples directly within the input prompt, \glspl{llm} can draw analogies and generalize to new tasks, a capability known as \gls{icl} \autocite{brown2020language, chowdhery2023palm, openai2023gpt04}. 
In \gls{icl}, the model is presented with a few demonstration examples alongside a query, all within the same input—--a technique known as few-shot prompting. 
The model's parameters remain unchanged; instead, it is expected that the model can recognize patterns within the prompt and generate an appropriate response \autocite{von2023transformers}. \gls{icl} enables models to learn on the fly, reducing the barrier to entry for users without deep \gls{ml} expertise. 
However, because the model does not retain memory between queries, the learned knowledge is temporary and is subsequently lost in subsequent queries. 
Additionally, \gls{icl} tends to struggle with tasks that require multi-step reasoning \autocite{brown2020language}. 
To address this limitation, task decomposition techniques have been introduced, with the earliest being \gls{cot} \autocite{wei2022chain}. 
Rather than relying solely on examples, this approach enriches the prompt with a series of reasoning steps that guide the model toward the correct answer \autocite{wei2022chain}. 
Considering that prompting approaches do not require an in-depth understanding of machine learning, they have proven very useful for a range of chemical tasks, including chemical data extraction, \gls{qa}, and property prediction \autocite{liu2025integrating, zheng2023chatgpt, mirza2024large, ramos2023bayesian}.

\paragraph{Fine-tuning} \label{sec:fine-tuning} 
Fine-tuning directly changes the weights of the model (see \Cref{sec:fine_tuning_coloring}). 
The fine-tuning strategy depends on the size and complexity of the target dataset as well as the pre-trained model. For many tasks, especially when using a powerful pre-trained model, it is often sufficient to freeze the entire model except for the final layer and only train that layer's parameters. 
However, as the target task diverges more significantly from the pre-trained model’s original objectives, more adaptation may be necessary. This can include replacing specific layers in the model to better suit the new task. 
For instance, in autoencoder architectures, it's common to freeze the encoder and replace the decoder. In \glspl{gnn}, the graph convolutional layers are typically frozen, while the final fully connected layers are replaced and re-trained. In some cases, it may be necessary to fine-tune the entire model, an especially resource-intensive process for \glspl{llm}, whose parameters can be in billions.
Despite these innovations, one key limitation of fine-tuning remains: adapting to a new modality, which often requires architectural changes or switching to a different model. However, \glspl{llm} offer a unique workaround. Many regression or classification tasks can be reformulated into a text-based format, allowing a single language model to be fine-tuned across a wide range of tasks. This is known as \gls{lift} \autocite{dinh2022lift}, which enables us to utilize a single \gls{gpm} for a diverse set of tasks.

Beyond adapting a model's internal knowledge through prompting or fine-tuning, its capabilities can be expanded by coupling it with external resources. This approach transforms a static model into a dynamic problem-solver that can access up-to-date information and perform actions in the world. This practice of designing and delivering task-relevant information is often referred to as context engineering. The necessary context can be provided through several complementary approaches that operate during inference time. 

\subsection{System-level Integration: Agents} \label{sec:agents}

While powerful, \glspl{gpm} are fundamentally static entities. Their knowledge is frozen at the time of training, and they cannot interact with the world beyond the information they process.  They cannot browse the web for the latest research, execute code to perform a calculation, or control a robot to run an experiment. 
To overcome these limitations and apply the reasoning capabilities of \glspl{gpm} to complex, multistep scientific problems, so-called \gls{llm}-based agents have emerged.

An \gls{llm}-based agent is a system that leverages an \glspl{llm} as its core \enquote{brain} but couples it with a set of tools to perceive and act upon its environment. To use a tool, the agent generates text containing the tool's name and its required inputs (arguments). The framework managing the agent recognizes this specific text, executes the corresponding tool, and then feeds the result back to the \gls{llm}. This transforms the model from a passive generator of text into an active problem-solver that can formulate plans, execute actions, observe the results, and adapt its strategy accordingly.  For chemists and material scientists, this paradigm shift is profound. It moves from asking a model a question to giving it a research goal, which it can then pursue autonomously.

\begin{figure}[htb]
    \centering
    \includegraphics[width=1\textwidth]{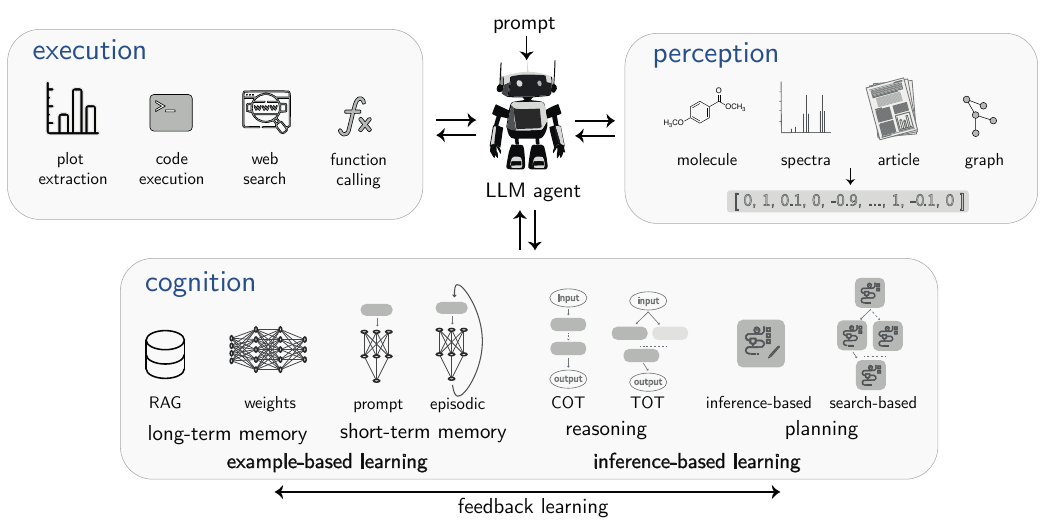}
    \caption{\textbf{The execution-cognition-perception capabilities of an \gls{llm}-agent:} This figure illustrates how agents orchestrate complex problems. At the core, an \gls{llm}-agent coordinates multiple capabilities. Upon prompting, the agent can execute tools. Tools include, but are not limited to, running code, searching for information, or calling functions. These actions feed into the agent's perception system, which transforms raw data into structured representations (in combination, for example, agents can obtain figures' information by executing \gls{ocr} tools on paper). The cognitive architecture underneath serves as the \enquote{agent's brain}, utilizing both memory systems (long-term knowledge storage and short-term contextual awareness) alongside reasoning mechanisms and planning strategies. This creates a dynamic setup, where execution produces observations, cognition interprets those observations and formulates plans, and new actions are taken based on improved understanding.}
    \label{fig:agent-loop}
\end{figure}

\Cref{fig:agent-loop} illustrates the fundamental components of an agentic framework, often conceptualized through the interacting modules of perception, cognition, and execution. It is important to note that this is one possible way to formalize an agent's architecture; other organizational structures exist. Rather than a strict, sequential loop, these components represent a set of capabilities that the agent's core \gls{llm} can dynamically draw upon to achieve complex objectives.

\subsubsection{Core Components of an Agentic System}\label{sec:arch_agents}

An agent is a system composed of several key components that work in concert.

\paragraph{Cognitive Engine} This is typically a powerful \gls{llm}, though other approaches exist, including classical planning systems or hybrid systems. Modern implementations may interface with \glspl{llm} via internal representations (hidden states) rather than solely through text output. 

It is responsible for all high-level reasoning, including understanding the user's objective, breaking it down into smaller, manageable steps (see planning \Cref{sec:planning}), and deciding which tools to use to accomplish each step.

\paragraph{Tool Augmentations (Execution)}
\label{sec:tool_augmentation}
Tools are external programs or functions that the agent can call upon to perform actions. They allow agents to interact with the world beyond their internal knowledge. \autocite{schick2023toolformer, parisi2022talm}. Tool augmentation can range from simple tools, such as calculators, to more complex systems that involve web searches, code execution, and integration with robots \autocite{darvish2025organa, chan2024mle, wei2025browsecomp}. In a chemical context, tools can be as simple as a stoichiometry calculator or as complex as a Python script that runs a \gls{dft} simulation using specialized software, a search \gls{api} for querying chemical databases like \modelname{PubChem}, or a controller for a robotic synthesis platform \autocite{boiko2023autonomous, darvish2025organa, bran2024augmenting}.

\paragraph{Memory \& \gls{rag} }
\label{sec:rag}
Agents need to maintain context over long and complex tasks. The memory module provides this capability. Short-term memory is often handled within the finite context window (i.e., number of tokens an \gls{llm} can process), keeping track of the immediate chain of thought and recent actions.
While short-term memory (e.g., context) is transient, \glspl{llm} model weights serve as long-term memory. However, these weights often lead to reduced performance on knowledge-intensive scientific tasks and increased susceptibility to hallucinations---generating incorrect or fabricated information \autocite{marcus2020next}.
One effective way to address this limitation is to pair the model with an external knowledge base.\autocite{lewis2020retrieval} Such Long-term memory can be implemented using external databases (e.g., vector stores) where the agent can store and retrieve key findings, successful strategies, or experimental results from past interactions, enabling it to learn and improve over time \autocite{chen2023chemist}. The widely adopted execution of Long-term memory is \gls{rag}. \gls{rag} works by retrieving a set of relevant documents from a designated knowledge database based on the input query. These retrieved documents are then concatenated with the original prompt and passed to the \gls{llm}, which generates the final output. In scientific applications, this is particularly valuable, as the system can be continuously updated with the latest research and discoveries. In the field of chemistry, \gls{rag} has primarily been used to answer domain-specific questions based on scientific literature and assist in experimental design \autocite{chen2023chemist, skarlinski2024language}.

\subsubsection{Approaches for Building Agentic Systems}

A well-known approach for building \gls{llm}-based agents is called \gls{react}\autocite{yao2023react}. In \gls{react}, the agent repeatedly goes through a cycle of thinking, performing an action, and then reasoning about the tool output. 
This structured problem-solving is achieved by prompting the model to generate its response following a specific \enquote{Think}, \enquote{Act}, \enquote{Observe} format.
First, the agent considers the problem it needs to solve, focusing on its primary objective. It devises a plan, identifies any missing information, and determines which tool can help it move forward. Next, the agent acts by selecting and utilizing the appropriate tool with the necessary information. For instance, if it needs to find a compound's boiling point, it might use a tool that searches a chemical database using the compound's name or its \gls{smiles} string. After that, the agent observes the outcome by examining the tool's output. This output then becomes new information for the agent. The agent then repeats the cycle, taking this new observation into account as it plans its following action. This loop continues until the agent reaches its main goal (see example~\ref{example:structure_eluc}). This repeating process helps the agent deal with mistakes, adjust to unexpected results, and break down a big task, like \enquote{finding a better catalyst for this reaction}, into smaller, manageable steps that involve using tools (see example below).

\begin{examplebox}[label={example:structure_eluc}]{Agent-based reasoning for chemical structure elucidation} 
\textbf{User Query:} \enquote{Identify this unknown white solid organic compound from analytical data.}

\textbf{Agent Reasoning Cycle:}
\begin{enumerate}
\item \textbf{THINK:} \enquote{Unknown white solid. Need to analyze all spectra systematically.}
\item \textbf{ACT:} Analyze \texttt{IR} spectrum tool $\rightarrow$ Output peak data (\SIrange{3000}{3500}{\per\cm} broad, \SI{1680}{\per\cm} sharp, \SI{1600}{\per\cm}, \SI{1500}{\per\cm}).
\item \textbf{ACT:} Analyze \texttt{MS} tool $\rightarrow$ Output peak data (\texttt{Top 3 peaks (135.0, 26.48)}, \texttt{(106.9, 0.43)}, \texttt{(120.4, 0.31)}).
\item \textbf{OBSERVE:} \enquote{Broad \texttt{O-H} stretch suggests carboxylic acid. Sharp \texttt{C=O} at \texttt{1680} indicates aromatic carboxylic acid. Aromatic \texttt{C=C} stretches confirm benzene ring, and \texttt{MS} tool suggests molecular formula \ce{C8H8O2}.}
\item \textbf{THINK:} \enquote{Aromatic carboxylic acid with \ce{C8H8O2}... probably substituted benzoic acid. Need to determine substitution pattern.}
\item \textbf{ACT:} Run \ce{^1H} \texttt{NMR} integration tool $\rightarrow$ Input chemical shifts (\texttt{7.8 ppm} doublet, \texttt{7.3 ppm} doublet, \texttt{2.4 ppm} singlet, \texttt{12.1 ppm} broad).
\item \textbf{OBSERVE:} \enquote{Two aromatic doublets suggest para-disubstitution. Singlet at \texttt{2.4 ppm} integrates for \texttt{3H} --- likely \ce{CH3}. Broad peak at \texttt{12.1 ppm} confirms \ce{COOH}.}
\item \textbf{THINK:} \enquote{Para-disubstituted benzene with \ce{CH3} and \ce{COOH} groups. Could be \textit{p}-toluic acid.}
\item \textbf{ACT:} Search chemical database for \enquote{\texttt{para-methylbenzoic acid}}.
\item \textbf{OBSERVE:} \enquote{\texttt{4-methylbenzoic acid} (\textit{p}-toluic acid) matches: \ce{C8H8O2}, \texttt{mp} \SI{180}{\degree C}.}
\item \textbf{ACT:} Run structure validation tool $\rightarrow$ Compare predicted vs. observed spectra.
\item \textbf{OBSERVE:} \enquote{Predicted \ce{^1H} \texttt{NMR}: $\delta$  \texttt{7.9 (d, 2H)}, \texttt{7.2 (d, 2H)}, \texttt{2.4 (s, 3H)}, \texttt{12.0 (br, 1H)}. Perfect match!}
\item \textbf{ACT:} Cross-reference \texttt{MS} fragmentation pattern tool.
\item \textbf{OBSERVE:} \enquote{Base peak at \texttt{m/z 91} (tropylium ion, loss of \ce{COOH}), molecular ion at \texttt{136}. Consistent with \textit{p}-toluic acid fragmentation.}
\end{enumerate}

\textbf{RESULT:} Structure confirmed as \texttt{4-methylbenzoic acid}. All spectroscopic evidence supports this assignment.

The agent systematically combines \texttt{IR}, \texttt{NMR}, and \texttt{MS} data with database searching to solve a realistic structure determination challenge.
\end{examplebox}

Scientific discovery often requires diverse expertise and collaborative problem-solving. This has led to the development of multi-agent systems, which move beyond a single cognitive engine to orchestrate a team of agents that work together \autocite{wu2023autogen}. These systems can solve tasks that are too complex or multifaceted for any single agent to handle alone by enabling agents to communicate, delegate, and debate. \autocite{lazaridou2020emergent}
Several collaborative paradigms have emerged, each offering unique advantages:

\paragraph{Specialization and Division of Labor} \label{sec:multi-agent}
Just as a human research group has members with different roles, multi-agent systems can be composed of specialized agents. For example, in a chemistry context, a \enquote{Planner} agent might design a high-level research plan, a \enquote{Literature Searcher} agent could retrieve relevant papers, a \enquote{Computational Chemist} agent could run \gls{dft} simulations, and a \enquote{Safety Expert} agent could check proposed reaction steps for hazards.\autocite{Zou2025ElAgente} This division of labor shows this role-playing approach to be highly effective for complex tasks like software development, where agents take on roles such as \enquote{programmer}, \enquote{tester}, and \enquote{documenter} \autocite{qian2024chatdevcommunicativeagentssoftware}.

\paragraph{Refinement of Answers} A key weakness of single \glspl{llm} is their tendency to hallucinate or pursue a flawed line of reasoning. Multi-agent systems can mitigate this by introducing criticism and debate. In this paradigm, one agent might propose a solution (e.g., a synthetic pathway), while a \enquote{Critic} agent is tasked with finding flaws in the proposal. This adversarial or collaborative process forces the system to refine its ideas, correct errors, and explore alternatives, leading to more robust and reliable outcomes \autocite{liang2024encouragingdivergentthinkinglarge, du2023improving}. 

\paragraph{Context Compression through Parallelism}
A significant operational challenge for any \gls{llm}-based system is the finite context window. 
As a task becomes more complex, the conversational history can grow cluttered with irrelevant details, degrading the model's performance.\autocite{chirkova2025provence0, lee2024long1context} 
Multi-agent systems offer a  solution to this problem through a strategy that can be described as context compression. 
By assigning sub-tasks to specialized agents, the system allows each agent to operate in parallel with its own clean, dedicated context window. 
For example, a \enquote{Literature Searcher} agent's context is filled only with search queries and retrieved text, while a \enquote{Computational Chemist} agent's context contains only simulation inputs and results. 
These sub-agents essentially act as filters; they process large amounts of information and then \enquote{compress} their findings into concise summaries or structured data. 
These distilled insights are then passed back to a lead agent or aggregated. 
This not only speeds up information gathering but also ensures that the primary reasoning process is not diluted by excessive or irrelevant information \autocite{Breunig2025HowToFixYourContext}.

\paragraph{Swarm Intelligence and Parallel Exploration}
Inspired by natural systems like ant colonies, some multi-agent approaches use a \enquote{swarm} of less-specialized agents to explore a vast problem space in parallel. Instead of assigning fixed roles, a multitude of agents can independently investigate different hypotheses or search different regions of a chemical space. Their collective findings can then be aggregated to identify the most promising solutions. This is particularly powerful for optimization and discovery tasks, such as high-throughput virtual screening or materials design, where the goal is to efficiently search an enormous number of possibilities \autocite{chen2023agentversefacilitatingmultiagentcollaboration}.

\subsection{Models vs.\ Systems}
It is also crucial to distinguish between the foundational model itself (e.g., the GPT-4 \gls{llm}) and the \enquote{system} with which the user interacts (e.g., ChatGPT). 
Such systems are not merely the raw model; they incorporate additional layers for safety, prompt management, and some adaptation techniques discussed in this section. 
\section{Evaluations} \label{sec:evals}
\subsection{The Evolution of Model Evaluation}

Assessing modern \glspl{gpm} is challenging due to their broad applicability across diverse domains. 
Unlike specialized \gls{ml} models, which are designed for specific tasks and can be directly tested on well-defined objectives \autocite{raschka2018model}, it is impractical to evaluate \glspl{gpm} on every possible capability. As a result, many evaluations rely on structured benchmarks that measure proficiency in key areas such as mathematics, chemistry, and language understanding \autocite{tikhonov2023post}. 
However, such benchmarks often fall short in capturing open-ended problem-solving or emergent abilities that arise without explicit training for them and are sensitive to factors such as prompt phrasing and task framing \autocite{Siska2024, Andriushchenko2024agent}.

\begin{figure}
    \centering
    \includegraphics[width=1\textwidth]{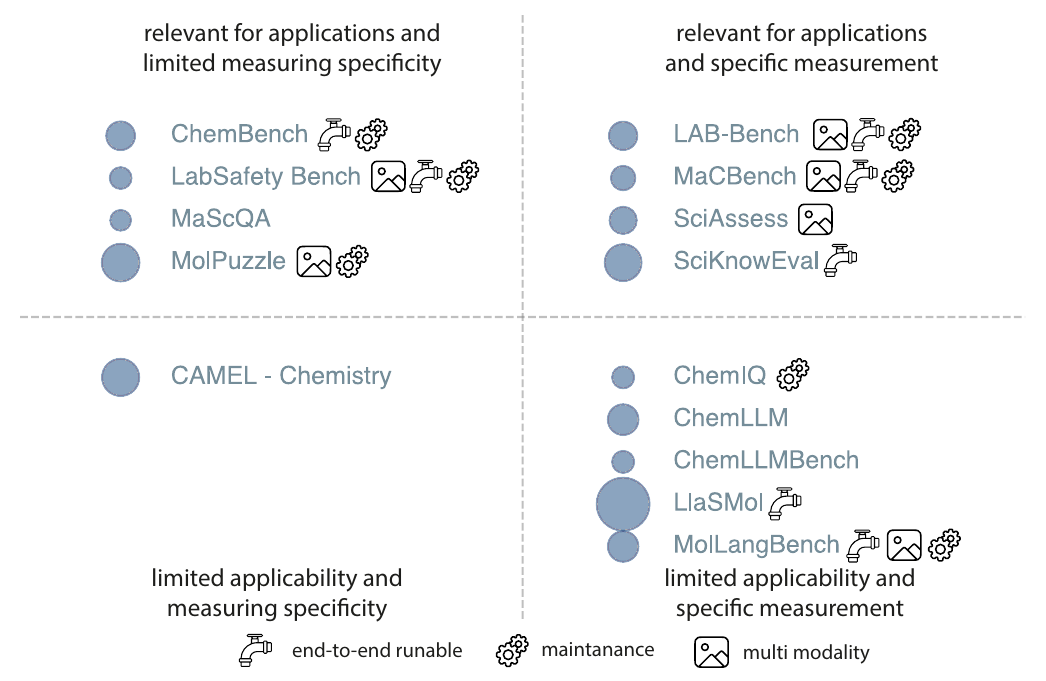}
    \caption{\textbf{Comparison of measurement specificity and application relevance of chemistry benchmarks.} This figure presents a subjective, qualitative positioning of selected chemistry-related benchmarks along the axes of application relevance and measurement specificity (the extent to which a benchmark evaluates well-defined and objectively measurable outputs---such as physical or chemical properties---rather than more ambiguous tasks like answering general chemistry trivia or \gls{mcq}). The icons indicate whether a benchmark incorporates multimodal inputs (e.g., images or spectra), if it is actively maintained (based on GitHub activity within the past 6 months), and if it supports end-to-end evaluation via a clearly described pipeline with code provided by the authors.}
    \label{fig:benchmarks}
\end{figure}

Early benchmarks primarily focused on evaluating specialized models based on their ability to predict molecular properties from molecular structures. \autocite{wu2018moleculenet} 
While useful, these evaluations largely emphasized numerical accuracy on the isolated tasks the models were fine-tuned on, without probing the more complex reasoning or generative capabilities that \glspl{gpm} aim to capture. 
Over time, this evolution expanded to exam-like problem-solving, assessing structured tasks similar to those found in academic chemistry courses. \autocite{zaki2023mascqa0, li2023camel} 
More recent efforts aim to evaluate a broader range of skills, including knowledge retrieval, logical reasoning, and even the ability to mimic human intuition when solving complex chemical problems.\autocite{feng2024sciknoweval0, mirza2024large} 
This shift highlights the need for more flexible evaluation methods that consider the specific context and nature of each task. Rather than relying solely on static benchmarks, there is a growing demand for assessments that dynamically account for the diversity of chemical tasks and the specific capabilities required to solve them---mirroring the multifaceted potential of \glspl{gpm} in chemistry. At the same time, the broadened scope of evaluation introduces additional choices about what and how to measure, which increases complexity. In the following section, we therefore discuss some of the most important of these choices in more detail. \Cref{tab:chemistry_benchmarks,fig:benchmarks} gives an overview of some benchmarks that have been used in the chemical sciences.

\begin{table}
\centering
\footnotesize
\caption{\textbf{Non-comprehensive overview of chemistry benchmarks}. Overview of chemistry benchmarks, including the topics covered, the curation method (automated, using \glspl{llm}, manual), and the approximate number of questions. We limit our scope here to benchmarks (and exclude other evaluation methods), since they constitute the most actively used and publicly available resources in the field at present.}
\label{tab:chemistry_benchmarks}
\begin{tabular}{p{3cm} p{6.9cm} p{1.2cm} p{1cm}}
\toprule
\textbf{Benchmark name} & \textbf{Overall topic} & \textbf{Curation method} & \textbf{Approx. count} \\
\midrule
CAMEL - Chemistry \autocite{li2023camel} & General Chemistry \gls{mcq} & A, L & 20 K \\
ChemBench \autocite{mirza2024large} & General Chemistry \gls{mcq}, Reasoning & M & 2.7 K \\
ChemIQ \autocite{runcie2025assessing} & Molecule Naming, Reasoning, Reaction Generation, Spectrum Interpretation & A & 796 \\
ChemLLM \autocite{zhang2024chemllm} & Molecule Naming, Property Prediction, Reaction Prediction, Reaction Conditions Prediction, Molecule \& Reaction Generation, Molecule Description & A, L & 4.1 K \\
ChemLLMBench \autocite{guo2023large} & Molecule Naming, Property Prediction, Reaction Prediction, Reaction Conditions Prediction, Molecule \& Reaction Generation & A, L & 800 \\
LAB-Bench \autocite{laurent2024lab0bench0} & Information Extraction, Reasoning, Molecule \& Reaction Generation & A, M & 2.5 K \\
LabSafety Bench \autocite{zhou2024labsafety} & Lab Safety, Experimental Chemistry & M, L & 765 \\
LlaSMol \autocite{yu2024llasmol} & Molecule Naming, Property Prediction, Reaction Prediction, Reaction Conditions Prediction, Molecule \& Reaction Generation, Molecule Description & A, L & 3.3 M \\
MaCBench \autocite{alampara2024probing} & Multimodal Chemistry, Information Extraction, Experimental Chemistry, Material Properties \& Characterization & M & 1.2 K \\
MaScQA \autocite{zaki2023mascqa0} & Material Properties \& Characterization, Reasoning, Experimental Chemistry & A & 650 \\
MolLangBench \autocite{cai2025mollangbench} & Molecule Structure Understanding, Molecule Generation & A, M & 4 K\\
MolPuzzle \autocite{guocan} & Molecule Understanding, Spectrum Interpretation, Molecule construction & A, L, M & 23 K \\
SciAssess \autocite{cai2024sciassess0} & General Chemistry MCQ, Information Extraction, Reasoning & A, M & 2 K \\
SciKnowEval \autocite{feng2024sciknoweval0} & General Chemistry MCQ, Information Extraction, Reasoning, Lab Safety, Experimental Chemistry & A, L & 18.3 K \\
\bottomrule
\end{tabular}
\begin{tablenotes}
\footnotesize
\item \textbf{Abbreviations:} 
A: Automated methods, L: Usage of \glspl{llm}, M: Manual curation.
\end{tablenotes}
\end{table}

\subsection{Design of Evaluations}
\label{sec:eval_design}

\paragraph{Desired Properties for Evaluations}
To meaningfully evaluate \glspl{gpm}, we must first consider what makes a good evaluation. 
Ideally, an evaluation should provide insights that translate into real-world impact, allowing comparisons between models. 
Thus, evaluation results must be stable over time and reproducible across different environments---assuming access to the same model weights or version, which is not always guaranteed when using proprietary \glspl{api}. \autocite{Ollion2024dangers}
A key challenge is so-called construct validity---ensuring that evaluations measure what truly matters rather than what is easiest to quantify. For example, asking a model to generate valid \gls{smiles} strings may test surface-level structure learning but fails to assess whether the model understands chemical reactivity or synthesis planning.
Many methods fall into the trap of assessing proxy tasks instead of meaningful competencies, which leads to misleading reports of progress. 
However, it is important to note that proxy tasks are often chosen because measurements at higher fidelity are more expensive or time-consuming to construct.

\paragraph{Data and Biases} The choice of what and how to measure is highly impacted by the data. 
Datasets in chemistry often differ in subtle but impactful ways: biases in chemical space coverage (e.g., overrepresented reaction types) \autocite{Jia_2019anthropogenic,Fujinuma_2022why}, variations in data fidelity (e.g., \gls{dft} vs. experimental measurements),  inconsistent underlying assumptions (e.g., simulation level or experimental conditions), and differences in task difficulty. 
These differences can distort what evaluations actually measure, making comparisons across models or tasks unreliable. \autocite{peng2024survey}
Moreover, the process of collecting or curating data itself introduces further variability, introduced by incomplete or biased coverage of the chemical space, computational constraints, or design decisions in the construction of tasks. 
As such, evaluations must be built using transparent, well-documented construction protocols, with clearly stated scope and limitations.

\paragraph{Scoring Mechanism} 

\begin{figure}
    \centering
    \includegraphics[width=1\textwidth]{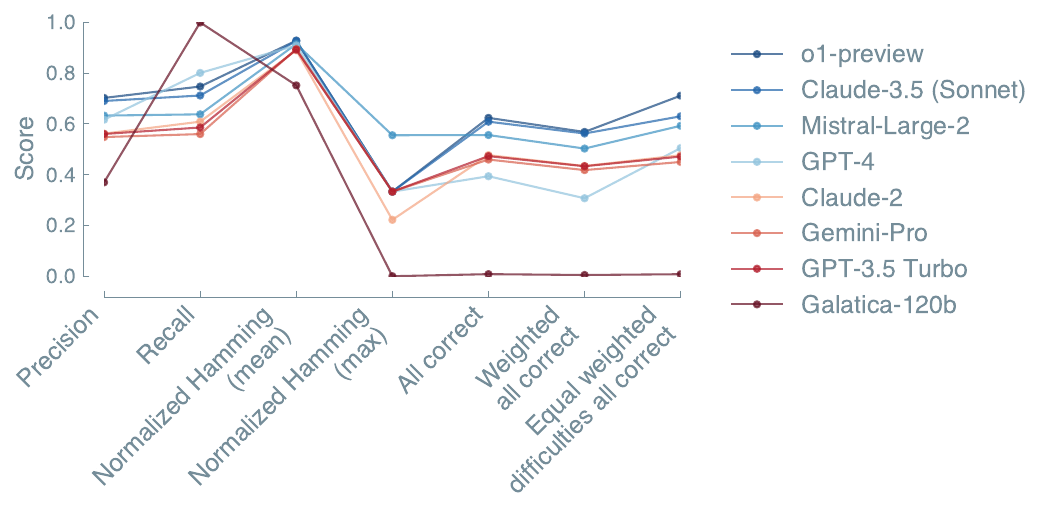}
    \caption{\textbf{\modelname{ChemBench} \autocite{mirza2024large} rankings based on different scoring metrics}: All metrics are a sum, weighted sum, or maximum values over all \glspl{mcq}. The weighted sums are calculated by taking the manually rated difficulty (basic, immediate, advanced) of the question into account. For equal weighting, all categories are weighted equally, regardless of the number of questions. The metric \enquote{all correct} is a binary metric indicating if a given answer is completely correct. For normalized Hamming (max), the normalized maximum value of the Hamming loss of each model was taken. We find that the ranking of models changes if we change the metric, or even just the aggregation---showcasing the importance of proper and transparent evaluation design.}
    \label{fig:scoring}
\end{figure}

The way model performance is scored has a direct impact on how results are interpreted and compared. 
Leaderboards and summary statistics often shape which models are considered state-of-the-art, making even small design choices in scoring, such as metric selection, aggregation, or treatment of uncertainty, highly consequential (see \Cref{fig:scoring}). 
Inconsistent or poorly designed scoring can lead to misleading conclusions or unfair comparisons.
For example, in tasks where multiple correct answers are possible, different evaluation strategies yield different insights: a permissive metric may assign partial credit for each correct option selected, while a stricter \enquote{all-or-nothing} metric only gives credit if the full set of correct choices is identified without any mistakes. The former captures varying levels of performance, while the latter enforces a binary pass or fail threshold. Aggregation strategies, such as task averaging or difficulty-based weighting, further influence the overall score and can substantially shift model rankings. 

\paragraph{Statistical significance and uncertainty estimation} Statistical significance and uncertainty estimation are essential for drawing robust conclusions. Evaluations must include a sufficient number of questions per task type to ensure statistical power, and repeated runs (e.g., different seeds or sampling variations) are needed to report confidence intervals or error bars. \autocite{tikhonov2023post} 
While this is feasible for automated, large-scale benchmarks, it becomes significantly more challenging in resource-intensive settings such as real-world deployment studies (e.g., testing a model in a wet-lab setting), where replicability and scale are limited.

\paragraph{Reproducibility and Reporting} Without clear and consistent documentation, even well-designed evaluations risk being misunderstood or unreproducible. 
To ensure that results are interpretable, verifiable, and extensible, every step of the evaluation process should be clearly specified, from prompt formulation and data pre-processing to metric selection and aggregation. 
Standardized evaluation protocols, careful tracking of environmental variables (e.g., hardware, model version, and sampling settings such as the inference temperature for \glspl{llm}), and consistent version control (documenting the exact version of the used model) are essential to avoid unintentional variation across runs. 
Additionally, communicating limitations and design decisions openly can help the broader community understand the scope and reliability of the reported results. 
\textcite{alampara2025lessons} proposed evaluation cards as a structured format to transparently report all relevant details of an evaluation, including design choices, assumptions, and known limitations, making it easier for others to interpret, reproduce, and build upon the results. See example \ref{example:best_practices} for some of the best practices while building benchmarks.

\begin{examplebox}[label={example:best_practices}]{Best practices for robust model evaluation}
\textbf{Setup and Reporting}
\begin{itemize}
    \item Ensure full reproducibility: document all steps from data preparation to scoring.
    \item Fix and report: model checkpoint or \gls{api} version, date and time, inference settings (e.g., temperature, context length).
    \item Store all evaluation assets (prompts, datasets, scripts) under version control for traceability.
    \item Clearly define: evaluation goal, task scope, data origin, and underlying assumptions.
\end{itemize}

\textbf{Validation and Sanity Checks}
\begin{itemize}
    \item Check for potential training data leakage by testing partial completions or measuring accuracy drop under paraphrasing \autocite{duarte2024detecting}.
    \item Use clear-cut factual questions phrased in multiple ways to test consistency.
    \item Include prompts containing obvious errors (e.g., “benzene is not aromatic”) to probe factual robustness.
    \item Perform repeated runs with different random seeds or sampling temperatures to estimate uncertainty.
\end{itemize}

\textbf{Bias and Construct Validity}
\begin{itemize}
    \item Verify that tasks measure intended skills (reasoning vs. recall).
    \item Balance question difficulty; include control and counterfactual items.
    \item Check dataset composition for chemical space or label bias.
\end{itemize}

\textbf{Publication and Communication}
\begin{itemize}
    \item Double-check factual claims against trusted references.
    \item Report both metrics and qualitative error patterns.
    \item Use structured evaluation cards \autocite{alampara2025lessons} for transparency.
\end{itemize}
\end{examplebox}

\subsection{Evaluation Methodologies}
\label{sec:eval_methods}
\paragraph{Representational vs.\ Pragmatic Evaluations}
It is useful to think of evaluations in \gls{ml} as living on a spectrum from representational to pragmatic. 
Representational evaluations focus on measuring concepts that exist in the real world. For instance, one might ask how well a model predicts a physically significant quantity like the band gap of a material or the yield of a chemical reaction. In contrast, in pragmatic evaluations, the concept we measure is defined by the evaluation procedure itself. A well-known example of this is IQ tests. The IQ is not a physical property that exists in the world independent of the test. It is rather defined by the measurement procedure. 
In the practice of evaluating \gls{ml} models, this includes tasks like answering \gls{mcq} or completing structured benchmarks, where meaning emerges primarily through performance comparison. Such evaluations are essential for standardization and progress tracking; however, they risk creating feedback loops, as models may end up optimized for benchmark success rather than real-world usefulness, thereby reinforcing narrow objectives or biases.\autocite{alampara2025lessons, skalse2022defining}

\paragraph{Estimator Types} To systematically evaluate \glspl{gpm}, various methods (so-called estimators) have been developed, yet their application in chemistry remains limited.
Broadly, these approaches can be categorized into traditional benchmarks, challenges and competitions, red teaming and capability discovery, real-world deployment studies, ablation studies, and systematic testing. 
Each evaluation method has its own strengths and limitations, and no single approach can comprehensively capture a model’s performance across all potential application scenarios.

\begin{figure}
    \centering    \includegraphics[width=1\textwidth]{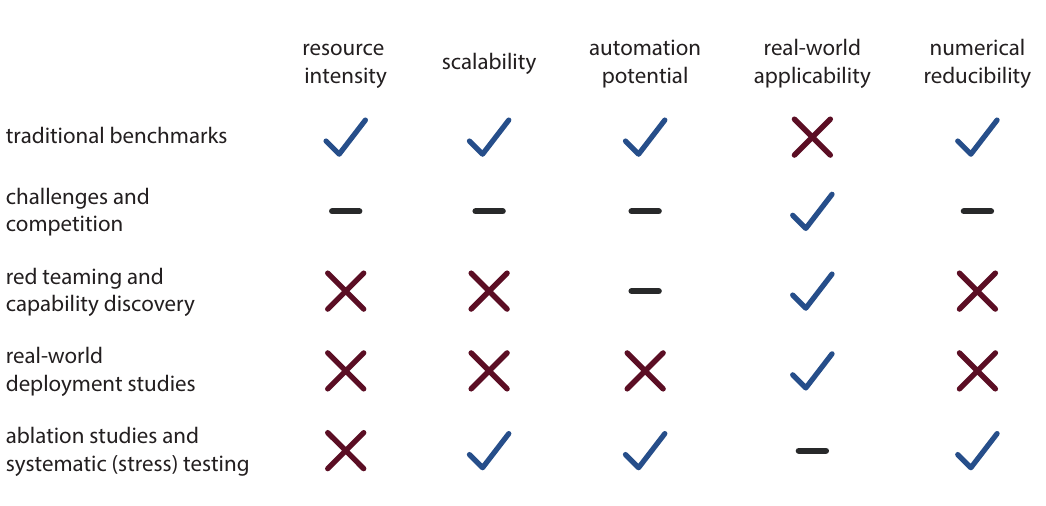}
    \caption{\textbf{Comparison of Evaluation Methodologies}: This figure compares five common evaluation methodologies for \glspl{gpm} across the dimensions of resource intensity (required human and computational effort), scalability (ease of applying the method across tasks and models), automation potential (need for manual intervention), real-world applicability (alignment with practical use cases), and numerical reducibility (ability to express results quantitatively). Checkmarks indicate a strength in the respective dimension, crosses denote a limitation, and dashes represent a neutral position.}
       \label{fig:estimators}
\end{figure}

\paragraph{Traditional Benchmarks}\label{para:trad_benchmarks}
Traditional benchmarks can provide a fast and scalable evaluation of the models. In the context of \gls{ml}, a benchmark typically refers to a curated collection of tasks or questions alongside a defined evaluation protocol, which allows different models to be compared under the same conditions. 
Despite their advantages, summarized in \Cref{fig:estimators}, benchmarks come with  limitations. They often struggle to capture real-world impact, as they evaluate models in controlled environments that may not reflect the complexity and open-endedness of real-world applications. Current chemistry benchmarks like ChemBench\autocite{mirza2024large} and CAMEL - Chemistry\autocite{li2023camel} mostly fall on the pragmatic side of the measurement spectrum, as they are designed for comparison and decision-making rather than assessing inherent model properties. 
Classical benchmarks such as \modelname{MoleculeNet} \autocite{wu2018moleculenet} typically target molecular properties---such as solubility or binding affinity---that are experimentally measurable and representationally grounded. However, their evaluation setup, which relies on static and narrowly defined datasets, often fails to capture real-world applicability.

An additional disadvantage of traditional benchmarks is ease of overfitting, where models are optimized for high scores rather than genuine improvements in real-world applications. 
This problem is closely linked to data leakage---also known as test set pollution---which refers to the unintentional incorporation of test set information into the training process. Such leakage can distort performance estimates, especially when models are trained on publicly available benchmarks.\autocite{thompson2025chatbots}
This exposure increases the risk that models learn to perform well on specific benchmark questions rather than developing a deeper, more generalizable understanding of the underlying concepts, leading to an overestimation of their true capabilities. 

Several strategies have been proposed to mitigate these issues. One approach is to keep a portion of the benchmark private, preventing models from being exposed to evaluation data during training---as implemented in \modelname{LAB-Bench}, where $20\%$ of the questions are held out to safeguard against data leakage and overfitting.\autocite{laurent2024lab0bench0} 
Alternatively, some initiatives explore privacy-preserving methods or the use of trusted third parties to evaluate models on held-out data without releasing it publicly. \autocite{eleutherai2024thirdparty}
Another strategy is to regularly update benchmarks by introducing more difficult or diverse tasks over time. \autocite{jimenez2023swe} 
While this helps reduce overfitting by continually challenging models, it complicates long-term comparisons across versions and can undermine stability in performance tracking. \autocite{alampara2025lessons}

Another limitation is that most benchmarks force models to respond to every question, even when uncertain, preventing evaluation of their ability to recognize what they do not know---an essential skill in real-world settings. \modelname{LAB-Bench} addresses this by allowing models to abstain via an \enquote{insufficient information} option, enabling a more nuanced assessment that distinguishes between confident knowledge and uncertainty.

\paragraph{Challenges and Competitions}
Competitions and challenges offer a structured way to evaluate models under realistic conditions, emphasizing prospective prediction and reducing overfitting risks.\autocite{Moult2005} 
Participants typically must submit predictions for a hidden test set, enabling blind, fair evaluation that more closely mirrors real-world deployment. Unlike standard benchmarking setups, these challenges often involve tasks with temporal, external, or domain-specific novelty, making them more resistant to subtle data leakage or overfitting through test set familiarity. 
While such challenges allow for systematic and rigorous assessment of model capabilities, they also require substantial coordination and oversight by a trusted third party.

Such challenges have been rare in chemistry to date, though recent examples in areas like polymer property prediction \autocite{gang2025neurips} suggest this is beginning to change. 
More successful examples exist in related domains---most notably the \gls{caspr} competition\autocite{Moult2005} for protein structure prediction in biology, and the crystal structure prediction blind test challenge\autocite{Lommerse2000} in materials science.

\paragraph{Red Teaming and Capability Discovery} \label{para:red_teaming}
Red teaming focuses on testing models in ways that they were not explicitly designed for, often probing their weaknesses, as well as unintended and unknown behaviors. \autocite{perez2022red, ganguli2022red} 
This group of evaluations includes attempts to bypass alignment mechanisms through adversarial prompting---deliberate attempts to elicit harmful, unsafe, or hidden model outputs by manipulating the input in subtle ways---or to reveal unintended capabilities. \autocite{zhu2023prompt, kumar2023computation} 
Unlike standardized benchmarks, red teaming can reveal model abilities that remain undetected in evaluations with predefined tasks. However, a major challenge is the lack of systematic comparability. 
Results often depend on specific test strategies and are harder to quantify across models. Currently, most red teaming is conducted by human experts, making it a time-consuming process. Automated approaches are emerging to scale these evaluations \autocite{ge2023mart0}, but in domains like chemistry, effective automation requires models to possess deep scientific knowledge, which remains a significant challenge. 

A concrete example of red teaming in chemistry was presented in the \modelname{GPT-4} Technical Report \autocite{openai2023gpt04}. 
By augmenting \modelname{GPT-4} with tools like molecule search, synthesis planning, literature retrieval, and purchasability checks, red-teamers were able to identify purchasable chemical analogs of a given compound, and even managed to have one delivered to a home address. 
While the demonstration used a benign leukemia drug, the same approach could, in principle, be applied to identify alternatives to harmful substances.\autocite{urbina2022dual}

\paragraph{Real-World Deployment Studies} 
Real-world deployment studies evaluate models in practical use settings, such as testing a \gls{gpm} in a laboratory environment.
Unlike controlled benchmarks, these studies provide insights into how models perform in dynamic, real-world conditions, capturing challenges that predefined evaluations may overlook. 
For example, a generative model might be used to suggest synthesis routes that are then tested experimentally, revealing failures due to overlooked side reactions or missing reaction feasibility. However, they come with significant drawbacks: they are highly time-consuming, and systematic comparisons between models are difficult, as real-world environments introduce variability that is hard to control. 

To date, such evaluations remain rare in the chemical sciences.

\paragraph{Ablation Studies and Systematic Testing} 
Ablation studies analyze models by systematically isolating and testing individual components or capabilities. 
By removing or modifying specific parts of the models, the impact on performance can be evaluated, providing information on the model’s functionality and potential weaknesses. This approach can be relatively scalable and structured, allowing for thorough and reproducible assessments. Ablation studies reveal limitations and improve overall reliability by ensuring a deeper understanding of how different elements contribute to the model’s behavior. 

\textcite{alampara2024probing} conducted ablation studies to isolate the effects of scientific terminology, task complexity, and prompt guidance on model performance in multimodal chemistry tasks. 
In \modelname{MaCBench}, they showed that removing scientific terms or adding explicit guidance substantially improved model accuracy, suggesting that current models often rely on shallow heuristics rather than deep understanding. These structured ablations highlight specific failure modes and inform targeted improvements in prompt design and training strategies.

\subsection{Future Directions} \label{sec:evals-future}

\paragraph{Emerging Evaluations Needs} To evaluate \glspl{gpm} in real-world conditions, more open-ended, multimodal, and robust approaches are needed. This is particularly evident in chemistry, where meaningful tasks often extend beyond text and require interpreting molecular structures, reaction schemes, or lab settings.
Here, vision plays a fundamental role, enabling perception and reasoning in complex environments---such as reading labels, observing color changes, or manipulating an apparatus. \autocite{Eppel2020computer}
In addition to visual cues, auditory signals---such as timer alerts or mechanical noise---can play a critical role in ensuring safety and coordination in lab environments. 
Sensorimotor input may also be relevant for simulating or guiding physical manipulation tasks, such as pipetting, adjusting equipment, or following multistep experimental procedures.

Beyond multimodality, another crucial challenge lies in evaluating open-ended scientific capabilities. 
Unlike well-defined benchmarks with fixed answers, real-world scientific inquiry is inherently open-ended.\autocite{mitchener2025bixbench0} This not only demands flexible and adaptive evaluation schemes but also raises deeper questions about what constitutes scientific understanding in generative models. 
This becomes even more important as agent-based systems (\Cref{sec:agents} gain traction---models that do not simply respond to prompts but autonomously plan, reason, and execute multistep tasks in interaction with tools, databases, and lab environments. \autocite{cao2024agents, mandal2024autonomous}
Simple input-output benchmarks are insufficient; instead, we need frameworks that can track progress in dynamic, goal-driven settings, where multiple valid solutions may exist. \autocite{nie2025assessing} 

In parallel to capability assessments, the evaluation of safety (see \Cref{sec:safety}) is becoming increasingly important---especially as \glspl{gpm} gain access to sensitive scientific knowledge and tools.\autocite{bran2024augmenting, boiko2023autonomous} 
Current safety evaluations often rely on manual red teaming \Cref{sec:eval_methods}, which is neither scalable nor systematic. 
Future evaluation frameworks must therefore include robust, automated, and scalable safety testing pipelines, capable of detecting misuse potential and risky behaviors across modalities and contexts.\autocite{goldstein2023generative}

Moreover, evaluations should not be limited to static benchmarks.
One promising avenue could be the organization of recurring community-wide challenges, similar to established competitions in other fields (e.g., \gls{caspr}\autocite{Moult2005}). These challenges---ideally coordinated by major research consortia or national labs---can serve as shared reference points, drive innovation in evaluation design.

\paragraph{Standardization Efforts} One persistent challenge in evaluating \glspl{gpm} is the lack of common standards---whether in benchmark design, metric selection, or reporting protocols. This fragmentation makes it challenging to compare results or ensure reproducibility. 
While some degree of standardization can support transparency and cumulative progress, rigid frameworks risk constraining innovation and may conflict with the need in scientific discovery for more open-ended, adaptive evaluations. 

A more feasible path may lie in promoting transparent documentation of evaluation choices and developing meta-evaluation tools that assess the validity, coverage, and robustness of different approaches. Emerging frameworks such as \gls{irt} offer promising directions. \autocite{schilling2025lifting}

\section{Applications}

\subsection{Automating the Scientific Workflow} \label{sec:ai-scientists}

Recent advances in \glspl{gpm}, particularly \glspl{llm}, have enabled initial demonstrations of fully autonomous \gls{ai} scientists \autocite{schmidgall2025agent}. 
We define these as \gls{llm}-powered architectures capable of executing end-to-end research workflows based solely on the final objectives, e.g., \enquote{\textit{Unexplained rise of antimicrobial resistance in Pseudomonas. Formulate hypotheses, design confirmatory in vitro assays, and suggest repurposing candidates for liver-fibrosis drugs}}. 
Such systems navigate partially or entirely through all components of the scientific process outlined in \Cref{fig:applications}, and detailed in the subsequent sections. 

While there are plenty of demonstrations of such systems in \gls{ml} and programming, scientific implementations remain less explored.

\subsubsection{Coding and ML Applications of AI Scientists}

Frameworks such as \modelname{Co-Scientist} \autocite{gottweis2025towards}, and \modelname{\gls{ai}-Scientist} \autocite{yamada2025ai} aim to automate the entire \gls{ml} research pipeline. They typically use multi-agent architectures (described in detail in \Cref{sec:multi-agent}) where specialized agents manage distinct phases of the scientific method \autocite{schmidgall2025agentrxiv}. 
Critical to these systems is self-reflection \autocite{renze2024self0reflection}---iterative evaluation and criticism of results within validation loops. 
However, comparative analyses reveal that \gls{llm}-based evaluations frequently overscore outputs relative to human assessments \autocite{huang2023mlagentbench0, chan2024mle, starace2025paperbench0}. 
An alternative is to couple these systems with evolutionary algorithms.  \modelname{AlphaEvolve} \autocite{novikov2025alphaevolve} is an \gls{llm} operating within a \gls{ga} environment and discovered novel algorithms for matrix multiplication (which had remained unchanged for fifty years) and sorting.

\subsubsection{Chemistry and Related Fields}

In chemistry, proposed systems showed some initial sparks. \modelname{Robin} identified ripasudil as a treatment for \gls{damd} \autocite{ghareeb2025robin0}---despite pending clinical trials and the general debate for these systems about the novelty of their findings\autocite{Listgarten2024perpetual}.
However, automation of experiment execution poses a major constraint for the chemistry-focused \gls{ai}-scientists due to hardware requirements, making computational chemistry the most feasible subfield in which agents have successfully run simple quantum simulations \autocite{Zou2025ElAgente}. 
Further, the \glspl{llm} powering these systems exhibit limited chemical knowledge \autocite{mirza2024large}. 
Despite this, \modelname{ether0} \autocite{narayanan2025training}---the first chemistry-specialized reasoning \gls{llm} (see \Cref{sec:rl} for a deeper discussion on reasoning models)---demonstrated strong capabilities in molecular design and accurate reaction prediction, positioning it as a possible foundation for chemistry-focused \gls{ai} scientists.

\subsubsection{Are these Systems Capable of Real Autonomous Research?}

Although agents like \modelname{Zochi} \autocite{intologyai2025zochi} achieved peer-reviewed publication in top-tier venues (\gls{acl} 2025), their capacity for truly autonomous end-to-end research remains debatable \autocite{son2025ai}. 
Even when generating hypotheses that appear novel and impactful, their execution and reporting of these ideas, as demonstrated by \textcite{si2025ideation1execution}, yield results deemed less attractive than those produced by humans.  
Additionally, while \gls{ai} autonomous systems can generate hypotheses, conduct experiments, and produce publication-ready manuscripts, their integration requires careful consideration (refer to \Cref{sec:ethics} for further discussion on moral concerns surrounding these systems).

\subsubsection{Limitations}

Despite impressive demos, current \enquote{\gls{ai} scientists} systems are still hindered by fundamental limitations. 
Their literature synthesis is often shallow, and they have weak novelty checks, with audits of tools showing that they can misclassify familiar ideas as \enquote{new}, and generally struggle to ground claims in prior work \autocite{beel2025evaluating}. 
Execution is equally fraught; automated pipelines exhibit fragile performance, and silent errors that self-review cannot catch \autocite{luo2025automate3}, necessitating expert oversight and validation \autocite{gottweis2025towards}. 
In chemistry, beyond idea generation, the hardest problems are infrastructural: closing the loop in real labs still runs into orchestration and integration headaches, heterogeneous instrument \glspl{api}, messy data plumbing-issues \autocite{Canty2025science, fehlis2025accelerating}, and safety and governance concerns \autocite{Leong2024steering}.
Finally, evaluation is immature (refer to \Cref{para:trad_benchmarks} for a deeper discussion on these evaluation methods): benchmarks for autonomous agents remain narrow and inconsistent, making it hard to trust self-reported improvements or generalize beyond tightly curated demos \autocite{yehudai2025survey}.

\subsubsection{Open Challenges}

\begin{itemize}
    \item \textbf{Capability}: Developing reliable long-horizon planning with verifiable execution traces \autocite{starace2025paperbench0}, evaluated on standardized, domain-grounded benchmarks \autocite{yehudai2025survey}.
    \item \textbf{Evaluation Beyond \gls{llm}-as-Judge}: Persistent biases and instability call for human-grounded protocols, adversarial test sets, and cross-lab replication \autocite{thakur2024judging, starace2025paperbench0}.
    \item \textbf{Infrastructure}: Evolving literature operations from generic \gls{rag} to auditable claim–evidence graphs, contradiction mining, and robust novelty checks.
    \item \textbf{Trust \& Governance}: Ensuring trustworthy autonomy through calibrated uncertainty quantification and clear authorship policies aligned with scholarly norms \autocite{ACS_AI_Policy_2024}.
\end{itemize}

\begin{figure}[!ht]
    \centering
    \includegraphics[width=0.5\textwidth]{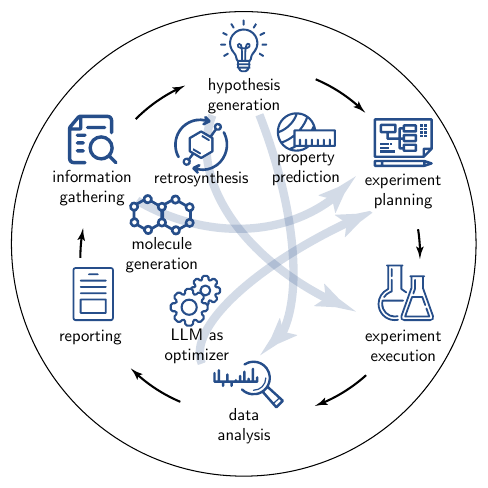}
    \caption{\textbf{Overview of the scientific process}. The outer elements represent the typical scientific research process: from gathering information and generating hypotheses based on the observations, to executing experiments and analyzing the results. The terms that are in the center represent data-driven \enquote{shortcuts} that \enquote{accelerate} the typical scientific method. All stages represented in the figure are discussed in detail in the following sections.}
    \label{fig:applications}
\end{figure}

\textit{True acceleration of chemical research and the ultimate goal of fully autonomous science require \gls{ai} systems that can operate across the entire scientific workflow. The stages outlined in \Cref{fig:applications}---from hypothesis generation to experimental execution and final reporting---represent the core of this process. While \glspl{gpm}, especially \glspl{llm}, show promise in individual components, for \gls{ai} to evolve from a showcase into a driver of fully autonomous research, it must master the entire workflow, seamlessly navigating between each of these stages.}

\subsection{Existing GPMs for Chemical Science}

The development of \glspl{gpm} for chemical science represents a departure from traditional single-task approaches. Rather than fine-tuning pre-trained models for specific tasks such as property prediction or molecular generation, these chemistry-aware models are intentionally designed to be capable of performing different chemical tasks. This multitask learning unifies related chemical tasks, boosting data efficiency and enabling emergent capabilities through joint training across domains. \Cref{tab:existing_gpms} shows an overview of the existing \glspl{gpm} in the chemical sciences.

\begin{table}[htbp]
\centering
\setlength{\tabcolsep}{5pt}
\renewcommand{\arraystretch}{1.3}
\setlength{\extrarowheight}{0.3ex}
\caption{\textbf{Non-comprehensive list of existing \glspl{gpm} in chemical sciences.} Most current models are trained on a mixture of general and domain-specific corpora. Models are sorted by the order they appear in the main text.}
\label{tab:existing_gpms}
\footnotesize
\begin{tabularx}{\linewidth}{
  l
  >{\raggedright\arraybackslash}X
  >{\raggedright\arraybackslash}X
  >{\raggedright\arraybackslash}X
  >{\raggedright\arraybackslash}X
}
\toprule
\textbf{Model} & \textbf{Architecture} & \textbf{Training} & \textbf{Representative Tasks} & \textbf{Representative Systems} \\
\midrule
\modelname{DARWIN 1.5} \autocite{xie2025darwin} & decoder-only transformer (LLaMA) with 7B parameters & fine-tuning on \glspl{qa}; multi-task fine-tuning on language-interfaced classification and regression tasks &  materials property prediction from natural language prompts & inorganic materials \\
\addlinespace[2.5pt]
\modelname{ChemLLM} \autocite{zhang2024chemllm} & decoder-only transformer (InternLM2) with 7B parameters & instruction-tuning on template-based \glsps{qa} using \gls{sft} or \gls{dpo} & multi-topic chemistry \gls{qa} & organic molecules \& reactions \\
\addlinespace[2.5pt]
\modelname{nach0} \autocite{livne2024nach0} &
encoder–decoder transformer (T5) with 250M or 780M parameters &
pre-trained using \gls{ssl} on scientific literature and \gls{smiles} strings; instruction-tuned &
natural language to \gls{smiles}; small-molecule property prediction &
organic molecules \& reactions \\
\addlinespace[2.5pt]
\modelname{LLaMat} \autocite{mishra2024foundational} &
decoder-only transformer (LLaMA) with 7B parameters &
continued pre-training on materials literature and crystallographic data &
materials information extraction; understanding and generating \glspl{cif} &
inorganic materials \\
\addlinespace[2.5pt]
\modelname{ChemDFM} \autocite{zhao2024chemdfm} &
decoder-only transformer (LLaMA) with 13B parameters &
pre-trained on 34B tokens from chemistry papers and textbooks; instruction-tuned with \gls{qa} and \gls{smiles} &
\gls{smiles} to text; molecule/reaction property prediction &
organic molecules \& reactions \\
\addlinespace[2.5pt]
\modelname{ether0} \autocite{narayanan2025training} &
decoder-only transformer (Mistral-Small) with 24B parameters; \gls{moe} &
\gls{sft} on reasoning traces, then \gls{rl} &
molecular editing; one-step retrosynthesis; reaction prediction &
organic molecules \& reactions \\
\bottomrule
\end{tabularx}
\end{table}

\paragraph{Domain Pre-Training and Multitask Learning} \modelname{DARWIN 1.5} used a multitask approach to fine-tune \modelname{Llama-7B} through a two-stage process \autocite{xie2025darwin}. In the first step, the base model was fine-tuned on $332k$ scientific \gls{qa} pairs to establish foundational scientific reasoning. Subsequently, the model underwent multitask learning on $22$ different regression and classification tasks based on experimental datasets. \modelname{DARWIN 1.5}'s core idea is the use of \gls{lift} (see \Cref{sec:model_adaptation}) on diverse materials tasks to induce cross-task transfer during training. However, in some cases, the results revealed negative task-interaction, i.e., some tasks had diminished performance under multi-tasking.

\modelname{ChemLLM} followed a similar approach to \modelname{DARWIN 1.5}: template-based instruction tuning (ChemData) on $~7M$ \gls{qa} pairs. \autocite{zhang2024chemllm}.

While the above examples focus on combining different in-domain tasks, \modelname{nach0} coupled natural language with chemical data \autocite{livne2024nach0}, based on a unified encoder-decoder transformer architecture. The model was pre-trained using \gls{ssl} on a combination of \gls{smiles} strings and natural language from scientific literature, and then instruction-tuned on chemistry and \gls{nlp} tasks. This allows \modelname{nach0} to translate between natural language and \gls{smiles}, in addition to tasks like \gls{qa}, information extraction, and molecule/reaction generation.

\modelname{LLaMat} employed \gls{peft} to continue the pre-training on crystal structure data in \gls{cif} format, enabling the generation of thermodynamically stable structures \autocite{mishra2024foundational}.

\modelname{ChemDFM} scaled this concept significantly, implementing domain pre-training on over 34 billion tokens from chemical textbooks and research articles \autocite{zhao2024chemdfm}. After that, through comprehensive instruction tuning, \modelname{ChemDFM} familiarizes itself with chemical notation and patterns, distinguishing it from more materials-focused approaches like \modelname{LLaMat} through its broader chemical knowledge base.

\paragraph{Reasoning-Based Approaches} A recent development in chemical \glspl{gpm} incorporates explicit reasoning capabilities. The \modelname{ether0} model demonstrated this approach as a 24 billion-parameter reasoning model trained on over 640k chemically-verifiable problems (e.g., through code) across 375 tasks, including single-step retrosynthesis, molecular editing, and reaction prediction \autocite{narayanan2025training}. 
Unlike previous models, \modelname{ether0}'s training used a novel \gls{rl} approach (see \Cref{sec:rl}). First, \modelname{DeepSeek-R1} was prompted to create long \gls{cot} traces ending in a \gls{smiles}. These traces were used to fine-tune a pre-trained \modelname{Mistral-Small-24B} using \gls{sft}, resulting in a \enquote{base reasoner} model. This model underwent \gls{rl} using \gls{grpo} on several verifiable chemical tasks to create \enquote{specialist} models---one for each task. High-quality outputs from these specialist models were selected and used to fine-tune the \enquote{base reasoner} to create a \enquote{generalist} model, capable of reasoning about all the tasks. In the last step, the generalist model undergoes another round of \gls{rl} on the chemical tasks to improve the performance. This method shows that structured problem-solving approaches can significantly improve performance on complex chemical tasks without the need for massive domain-specific corpora. 

These diverse approaches illustrate the evolving landscape of chemical \glspl{gpm}.
Still, most applications of \glspl{gpm} focus on using such models for one specific application, and we will review those in the following.

\subsubsection{Limitations}
Chemical \glspl{gpm} are emerging, but we lack systematic understanding of how to build effective ones.

The tokenization question remains unresolved. Domain-specific tokenizers might improve data efficiency, but they prevent reuse of models trained on general text datasets. The tradeoff remains unclear.

A deeper challenge exists: we might lack sufficient high-quality data. Existing chemical datasets are much smaller than those used in domains like mathematics (\Cref{sec:data-section}). The literature reports only successes. Failed experiments, discarded results, and negative outcomes never appear in published datasets. More critically, chemical practitioners rely on tacit knowledge\autocite{Polanyi_2009}---expertise acquired through experience that experts cannot fully articulate. This implicit understanding remains absent from existing datasets.

\subsection{Knowledge Gathering}\label{sec:information_gathering}
\begin{figure}[H]
    \centering
\includegraphics[width=1\textwidth]{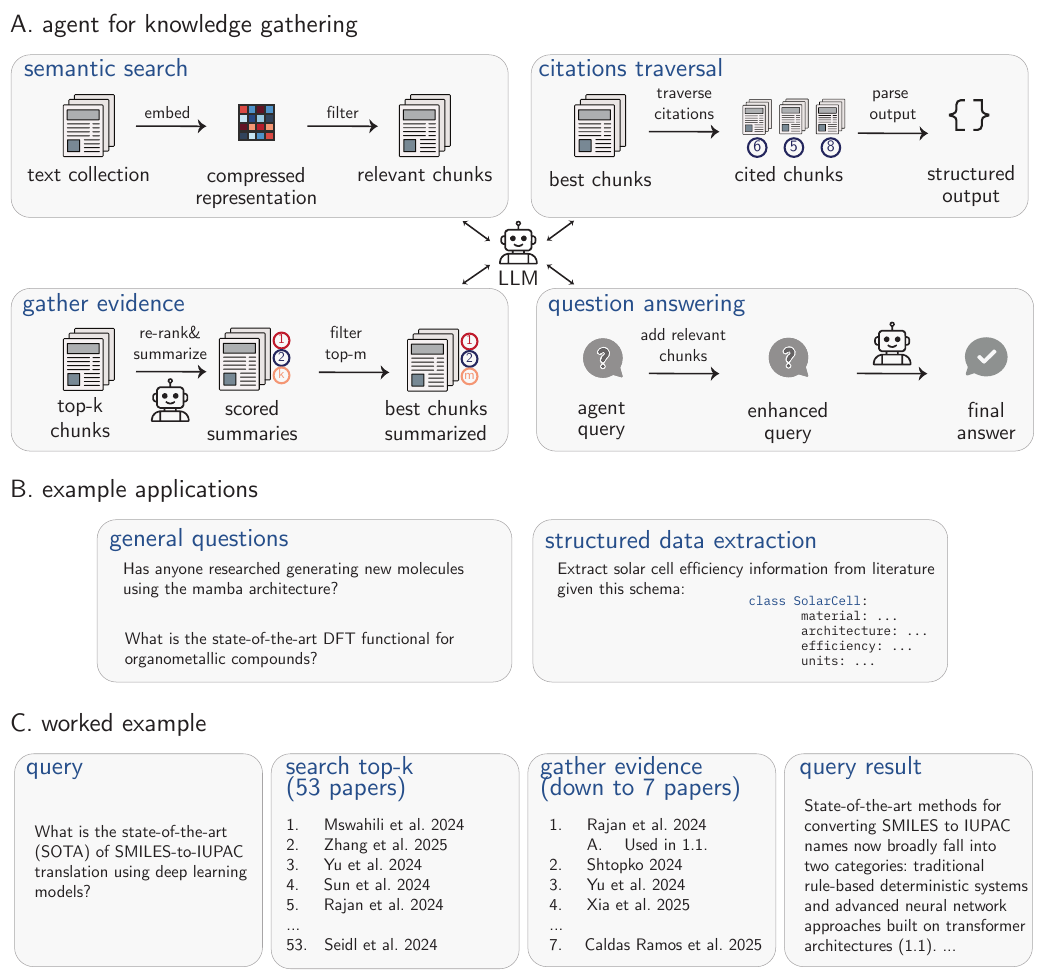}
    \caption{\textbf{A. A representation of a typical agent for scientific queries.} The \gls{llm} is the central piece of the system, surrounded by typical tools that improve its question-answering capabilities, together forming an agentic system. The tools represented in this figure are semantic search, citation traversal, evidence gathering, and question answering. Semantic search finds relevant documents. Evidence gathering ranks and filters chunks of text using \glspl{llm}. The citation traversal tool provides model access to citation graphs, enabling accurate referencing of each chunk and facilitating the discovery of additional sources. Finally, the question-answering tool (an \gls{llm}) collects all the information found by other tools and generates a final response to a user's query. This part of the figure is inspired by the \modelname{PaperQA2} agent.\autocite{skarlinski2024language} \textbf{B.} Two examples of applications are discussed in this section. \textbf{C.} worked example using the \texttt{FutureHouse}'s platform (the Crow agent). \enquote{used in 1.1} indicates that this reference has been used in the final response of the model\autocite{skarlinski2025futurehouse}}
\label{fig:knowledge_gathering}
\end{figure}

The rate of publishing keeps growing, and as a result, it is increasingly challenging to manually collect all relevant knowledge, potentially stymying scientific progress.\autocite{schilling2025text, Chu_2021}
Even though knowledge collection might seem like a simple task, it often involves multiple steps, visually described in \Cref{fig:knowledge_gathering}A. 
Here, we focus on structured data extraction and question answering. Example queries for both sections are in \Cref{fig:knowledge_gathering}B.

\paragraph{Semantic Search} A step that is key to most, if not all, knowledge-gathering tasks is \gls{rag}, discussed in more detail in \Cref{sec:rag}. 
Most commonly, this involves semantic search, intended to identify chunks of text with similar meaning. 
The difference between semantic search and conventional search lies in how each approach interprets queries. The latter operates through lexical matching---whether exact or fuzzy---focusing on the literal words and their variations. Semantic search, however, focuses on the underlying meaning and contextual relationships within the content. 

To enable semantic search, documents are converted into embeddings (see \Cref{sec:embeddings}) \autocite{bojanowski2017enriching}. 
They allow for similarity search by vector comparison (e.g., using cosine similarity for small databases or more sophisticated algorithms like \gls{hnsw} for large databases\autocite{malkov2018efficient}). 

In chemistry, semantic search has been used extensively to classify and identify chemical text.\autocite{Guo2021,beltagy2019scibert0,trewartha2022quantifying}

\subsubsection{Structured Data Extraction}

For many applications, it can be useful to collect data in a structured form, e.g., tables with fixed columns.
Obtaining such a dataset by extracting data from the literature using \glspl{llm} is currently one of the most practical avenues for \glspl{llm} in the chemical sciences \autocite{schilling2025text}.

\paragraph{Data Extraction Using Prompting} 

For most applications, zero-shot prompting should be the starting point. Zero-shot prompting has been used to extract data about organic reactions\autocite{rios2025llm,vangala2024suitability, Patiny2023automatic}, synthesis procedures for metal-organic frameworks\autocite{zheng2023chatgpt}, polymer data\autocite{schilling2024using,gupta2024data}, or other materials data\autocite{polak2024extracting,hira2024reconstructing,kumar2025mechbert,wu2025large,huang2022batterybert}. 
 
\paragraph{Fine-tuning Based Data Extraction} 

If a (commercial) model needs to be run very often, it can be more cost-efficient to fine-tune a smaller, open-source model compared to prompting a large model (see \Cref{sec:distillation}). 
In addition, models might lack specialized knowledge and might not follow certain style guides, which can be introduced with fine-tuning. 
\textcite{ai2024extracting} fine-tuned the \modelname{LLaMa-2-7B} model to extract chemical reaction data from \gls{uspto} into a \gls{json} format compatible with the schema of \gls{ord}\autocite{Kearnes_2021}, achieving an overall accuracy of more than $90\%$. 
In a different approach, \textcite{zhang2024fine} fine-tuned a closed-source model --- \modelname{GPT-3.5-Turbo} to recognize and extract chemical entities from \gls{uspto}. Fine-tuning improved the performance of the base model on the same task by more than $15\%$. 
\textcite{dagdelen2024structured} went beyond the afore-mentioned methods by using a human-in-the-loop data annotation process. Here, humans corrected the outputs from an \gls{llm} extraction instead of annotating data from scratch.

\paragraph{Agents for Data Extraction} 

Agents (\Cref{sec:agents}) have shown their potential in data extraction, though to a limited extent.\autocite{chen2024autonomous,kang2024chatmof}
For example, \textcite{ansari2024agent} introduced \modelname{Eunomia}, an agent that autonomously extracts structured materials data from scientific literature without requiring fine-tuning. Their agent is an \gls{llm} with access to tools such as chemical databases (e.g., the \modelname{Materials Project} database) and research papers from various sources. The document search tool leverages an \gls{api}-based embedding model to find the top-k most relevant passages based on semantic similarity. Other tools used include the chain-of-verification tool, which queries the agent independently (to avoid bias) in order to generate verification queries and finally produces a verified response.

While the authors claim this approach simplifies dataset creation for materials discovery, the evaluation is limited to a narrow set of materials science tasks (mostly focusing on \glspl{mof}), indicating the need for the creation of agent evaluation tools.

\subsubsection{Question Answering}
Besides extracting information from documents in a structured format, \glspl{llm} can also be used to answer questions---such as \enquote{Has X been tried before} by synthesizing knowledge from a corpus of documents (and potentially automatically retrieving additional documents). 

An example of a system that can do that is \modelname{PaperQA}. This agentic system contains tools for search, evidence-gathering, and question answering as well as for traversing citation graphs, which are shown in \Cref{fig:knowledge_gathering}. The evidence-gathering tool collects the most relevant chunks of information via the semantic search and performs \gls{llm}-based re-ranking of these chunks (i.e., the \gls{llm} changes the order of the chunks depending on what is needed to answer the query).
Subsequently, only the top-$n$ most relevant chunks are kept. To further ground the responses, citation traversal tools (e.g., Semantic Scholar\autocite{kinney2023semantic}) are used. 
These leverage the citation graph as a means of discovering supplementary literature references. Ultimately, to address the user's query, a question-answering tool (a specially prompted \gls{llm}) is employed. It initially augments the query with all the collected information before providing a definitive answer.
The knowledge aggregated by these systems could be used to generate new hypotheses or challenge existing ones. 

\subsubsection{Limitations} 

\Glspl{llm} and \gls{llm}-based systems are valuable information-gathering tools, but they have critical limitations. Their knowledge becomes outdated immediately after training. Without additional tools, they cannot identify when retrieved sources have been retracted or corrected. 

Effective knowledge gathering requires human-\gls{ai} collaboration. Current systems struggle to ask relevant clarifying questions.\autocite{choudhury2025bed0llm0} Domain-specific benchmarks for evaluating this capability remain absent, hindering progress in developing truly interactive knowledge-gathering agents.

\subsubsection{Open Challenges}

Critical challenges remain unresolved:

\begin{itemize}
    \item \textbf{Multimodal Understanding.} Current models cannot reliably extract data from plots, tables, and figures.\autocite{alampara2024probing,gupta2022discomat0} This severely limits data extraction because substantial chemical knowledge exists only in visual formats. Improved multimodal capabilities are essential for comprehensive literature mining.
    
    \item \textbf{Source Quality Assessment.} Selecting high-quality sources, that are also relevant to the scientific query, remains an open challenge, with scholarly metrics alone being insufficient.
\end{itemize}

\subsection{Hypothesis Generation} \label{sec:hypothesis-gen}

Coming up with new hypotheses represents a cornerstone of the scientific process \autocite{rock2018hypothesis}. Historically, hypotheses have emerged from systematic observation of natural phenomena, exemplified by Isaac Newton’s formulation of the law of universal gravitation \autocite{newton1999principia}, which was inspired by the seemingly mundane observation of a falling apple \autocite{kosso2017whatgoesup}.

In modern research, hypothesis generation increasingly relies on data-driven tools. 
For example, clinical research employs frameworks such as \gls{viads} to derive testable hypotheses from well-curated datasets \autocite{Jing2022roles}. Similarly, advances in \glspl{llm} are now being explored for their potential to automate and enhance idea generation across scientific domains \autocite{oneill2025sparks}. However, such approaches face significant challenges due to the inherently open-ended nature of scientific discovery \autocite{stanley2017openendedness}. 
Open-ended domains, as discussed in \Cref{sec:data-section}, risk intractability, as an unbounded combinatorial space of potential variables, interactions, and experimental parameters complicates systematic exploration \autocite{clune2019ai0gas0}.
Moreover, the quantitative evaluation of the novelty and impact of generated hypotheses remains non-trivial. 
As Karl Popper argued, scientific discovery defies rigid logical frameworks \autocite{popper1959logic}, and objective metrics for \enquote{greatness} of ideas are elusive \autocite{stanley2015greatness}. These challenges underscore the complexity of automating or systematizing the creative core of scientific inquiry.

\subsubsection{Initial Sparks}

Recent efforts in the \gls{ml} community have sought to simulate the hypothesis formulation process \autocite{Gu2025forecasting, arlt2024meta0designing}, primarily leveraging multi-agent systems \autocite{jansen2025codescientist0, kumbhar2025hypothesis}. 
In such frameworks, agents typically retrieve prior knowledge to contextualize previous related work, grounding hypothesis generation in existing literature \autocite{naumov2025dora, ghareeb2025robin0, gu2024interesting}. 
A key challenge, however, lies in evaluating the generated hypotheses. 
While some studies leverage \glspl{llm} to evaluate novelty or interestingness \autocite{zhang2024omni0}, recent work has introduced critic agents---specialized components designed to monitor and iteratively correct outputs from other agents---into multi-agent frameworks (see \Cref{sec:multi-agent}). 
For instance, \textcite{Ghafarollahi2024} demonstrated how integrating such critics enables systematic hypothesis refinement through continuous feedback mechanisms.

\begin{figure}[H]
    \centering
        \includegraphics[width=1\textwidth]{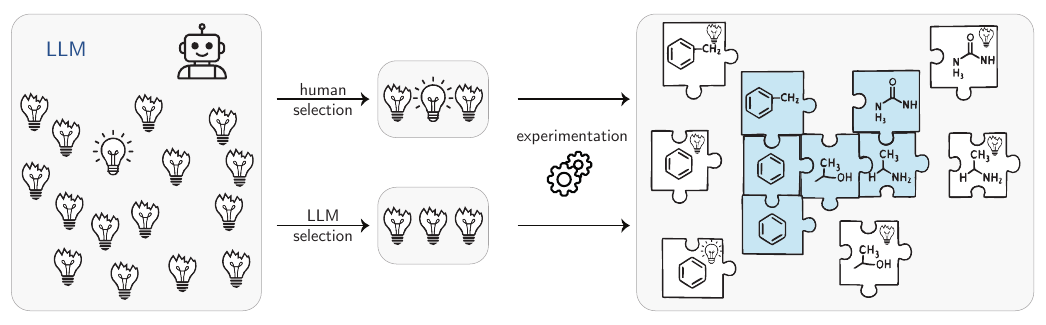}
    \caption{\textbf{Overview of \gls{llm}-based hypothesis generation}. 
    Current methods are based on \gls{llm}-sampling methods in which an \gls{llm} proposes new hypotheses. 
    The generated hypotheses are evaluated in terms of novelty and impact either by another \gls{llm} or by a human. Then, through experimentation, the hypotheses are transformed into results which showcase that current \glspl{llm} cannot produce groundbreaking ideas, limited to their training corpus, resulting in the best cases, in incremental work. This is shown metaphorically with the puzzle. The \enquote{pieces of chemical knowledge} based on the hypothesis produced by \glspl{llm} are already present in the \enquote{chemistry puzzle}, not unveiling new parts of it.}
    \label{fig:hypothesis-generation}
\end{figure}

However, the reliability of purely model-based evaluation remains contentious. 
\textcite{si2025llms} argued that relying on a \gls{llm} to evaluate hypotheses lacks robustness, advocating instead for human assessment. 
This approach was adopted in their work, where human evaluators validated hypotheses produced by their system, finding more novel \gls{llm}-produced hypotheses compared to the ones proposed by humans.
Notably, \textcite{yamada2025ai} advanced the scope of such systems by automating the entire research \gls{ml} process, from hypothesis generation to article writing. 
Their system’s outputs were submitted to workshops at the \gls{iclr} 2025, with one contribution ultimately accepted. However, the advancements made by such works are currently incremental instead of unveiling new, paradigm-shifting research (see \Cref{fig:hypothesis-generation}).

\subsubsection{Chemistry-Focused Hypotheses}

In chemistry and materials science, hypothesis generation requires domain intuition, mastery of specialized terminology, and the ability to reason through foundational concepts \autocite{miret2024llms}. 
To address potential knowledge gaps in \glspl{llm}, \textcite{wang2023scimon0} proposed a few-shot learning approach (see \Cref{sec:prompting}) for hypothesis generation and compared it with model fine-tuning for the same task. 
Their method strategically incorporates in-context examples to supplement domain knowledge while discouraging over-reliance on existing literature. 
For fine-tuning, they designed a loss function that penalizes possible biases---e.g., given the context \enquote{hierarchical tables challenge numerical reasoning}, the model would be penalized if it generated an overly generic prediction like \enquote{table analysis} instead of a task-specific one---when trained on such examples. 
Human evaluations of ablation studies revealed that \modelname{GPT-4}, augmented with a knowledge graph of prior research, outperformed fine-tuned models in generating hypotheses with greater technical specificity and iterative refinement of such hypotheses.

Complementing this work, \textcite{yang2025moose} introduced the \modelname{Moose-Chem} framework to evaluate the novelty of \gls{llm}-generated hypotheses. 
To avoid data contamination, their benchmark exclusively uses papers published after the knowledge cutoff date of the evaluated model, \modelname{GPT-4o}. 
Ground-truth hypotheses were derived from articles in high-impact journals (e.g., Nature, Science) and validated by domain-specialized PhD researchers.
By iteratively providing the model with context from prior studies, \modelname{GPT-4o} achieved coverage of over $80\%$ of the evaluation set’s hypotheses while accessing only $4\%$ of the retrieval corpus, demonstrating efficient synthesis of ideas presumably not present in its training corpus.

\subsubsection{Are LLMs Actually Capable of Novel Hypothesis Generation?}

Automatic hypothesis generation is often regarded as the Holy Grail of automating the scientific process \autocite{coley2020autonomous}. However, achieving this milestone remains challenging, as generating novel and impactful ideas requires questioning current scientific paradigms \autocite{Kuhn1962Structure}---a skill typically refined through years of experience---which is currently impossible for most \gls{ml} systems.

Current progress in \gls{ml} illustrates these limitations \autocite{kon2025exp0bench0, gu2024interesting}. Although some studies claim success, as \gls{ai}-generated ideas being accepted at workshops in \gls{ml} conferences via double-blind review \autocite{zhou2025tempest0}, these achievements are limited. 
First, accepted submissions often focus on coding tasks, one of the strongest domains for \glspl{llm}. Second, workshop acceptances are less competitive than main conferences, as they prioritize early-stage ideas over rigorously validated contributions. 
In chemistry, despite some works showing promise on these systems \autocite{yang2025moose0chem20}, \glspl{llm} struggle to propose innovative hypotheses \autocite{si2025ideation1execution}.
Their apparent success often hinges on extensive sampling and iterative refinement, rather than genuine conceptual innovation.

As \textcite{Kuhn1962Structure} argued, generating groundbreaking ideas demands challenging prevailing paradigms---a capability missing in current \gls{ml} models (they are trained to make the existing paradigm more likely in training rather than questioning their training data), as shown in \Cref{fig:hypothesis-generation}. 
Thus, while accidental discoveries can arise from non-programmed events (e.g., Fleming’s identification of penicillin \autocite{Fleming1929antibacterial, Fleming1945penicillin}), transformative scientific advances typically originate from deliberate critique of existing knowledge \autocite{popper1959logic, Lakatos1970falsification}. In addition, very often breakthroughs can not be achieved by optimizing for a simple metric---as we often do not fully understand the problem and, hence, cannot design a metric.\autocite{stanley2015greatness}
Despite some publications suggesting that \gls{ai} scientists already exist, such claims are supported only by narrow evaluations that yield incremental progress \autocite{novikov2025alphaevolve}, not paradigm-shifting insights. For \gls{ai} to evolve from research assistants into autonomous scientists, it must demonstrate efficacy in addressing societally consequential challenges, such as solving complex, open-ended problems at scale (e.g., \enquote{millennium} math problems
\autocite{Carlson2006millennium}).

Finally, ethical considerations become critical as hypothesis generation grows more data-driven and automated. Adherence to legal and ethical standards must guide these efforts (see \Cref{sec:safety}) \autocite{danish_gov2024hypothesis}.

\subsubsection{Limitations}

Current \gls{llm}-driven systems lack the kind of creativity needed for paradigm-shifting hypotheses, tending to rearrange training data and retrieved content rather than propose genuinely new mechanisms. As a result, the evaluation of such outputs is also fragile, because using proxy metrics for \enquote{novelty} or \enquote{impact} that poorly track real scientific value can be deceiving. Finally, the problem’s open-ended nature (\Cref{sec:data-section}) makes systematic benchmarking ill-posed.

\subsubsection{Open Challenges}

\begin{itemize}
    \item \textbf{Scalable Evaluation}: The open-ended assessment of a hypothesis's potential impact remains a core challenge, as current methods are difficult to scale, costly, and inefficient due to a heavy reliance on human input \autocite{nie2025assessing}.
    \item \textbf{The Integration Gap}: A critical disconnect persists between hypothesis generation and automated experimental validation, especially in fields like chemistry.
    \item \textbf{The Paradigm Limitation}: The underlying operational constraints of current modeling approaches inherently favor incremental progress over transformative breakthroughs.
\end{itemize}


\subsection{Experiment Planning}
\label{sec:planning}
Before a human or robot can execute any experiments, a plan must be created. 
Planning can be formalized as the process of decomposing a high-level task into a structured sequence of actionable steps aimed at achieving a specific goal. 
The term planning is often confused with scheduling and \gls{rl}, which are closely related but distinct concepts. Scheduling is a more specific process focused on the timing and sequence of tasks. 
It ensures that resources are efficiently allocated, experiments are conducted in an optimal order, and constraints (such as lab availability, time, and equipment) are respected.\autocite{kambhampati2023llmplanning} 
\gls{rl} is about adapting and improving plans over time based on ongoing results.\autocite{chen2022deep}

\subsubsection{Conventional Planning} 

Early chemical planning systems, such as \gls{lhasa} \autocite{corey1972computer} and \modelname{Chematica}\autocite{grzybowski2018chematica}, relied on simple rules and templates. In particular, \modelname{Chematica} used heuristic-guided graph search with rule-based transforms and scoring functions to prune and prioritize routes. Modern systems, like ASKCOS\autocite{tu2025askcos}, explicitly use search algorithms such as \gls{bfs}, \gls{mcts}\autocite{segler2017towards} to explore the combinatorially large space. But these planning search algorithms remain inefficient for long-horizon or complex planning tasks. \autocite{liu2024entropy, zhao2024efficient}

\subsubsection{LLMs to Decompose Problems into Plans}
\glspl{gpm}, in particular \glspl{llm}, can potentially assist in planning with two modes of thinking. 
Deliberate thinking can be used to score potential options or to decompose problems into plans. 
Intuitive thinking can be used to efficiently prune search spaces. 
These two modes align with psychological frameworks known as system-1 (intuitive) and system-2 (deliberate) thinking. \autocite{kahneman2011thinking}
In the system-1 thinking, \glspl{llm} support rapid decision-making by leveraging heuristics and pattern recognition to quickly narrow down options. 
In contrast, system-2 thinking represents a slower, more analytical process, in which \glspl{llm} solve complex tasks by explicitly generating step-by-step reasoning. \autocite{ji2025test}

\begin{figure}[!htbp]
    \centering
        \includegraphics[width=1\textwidth]{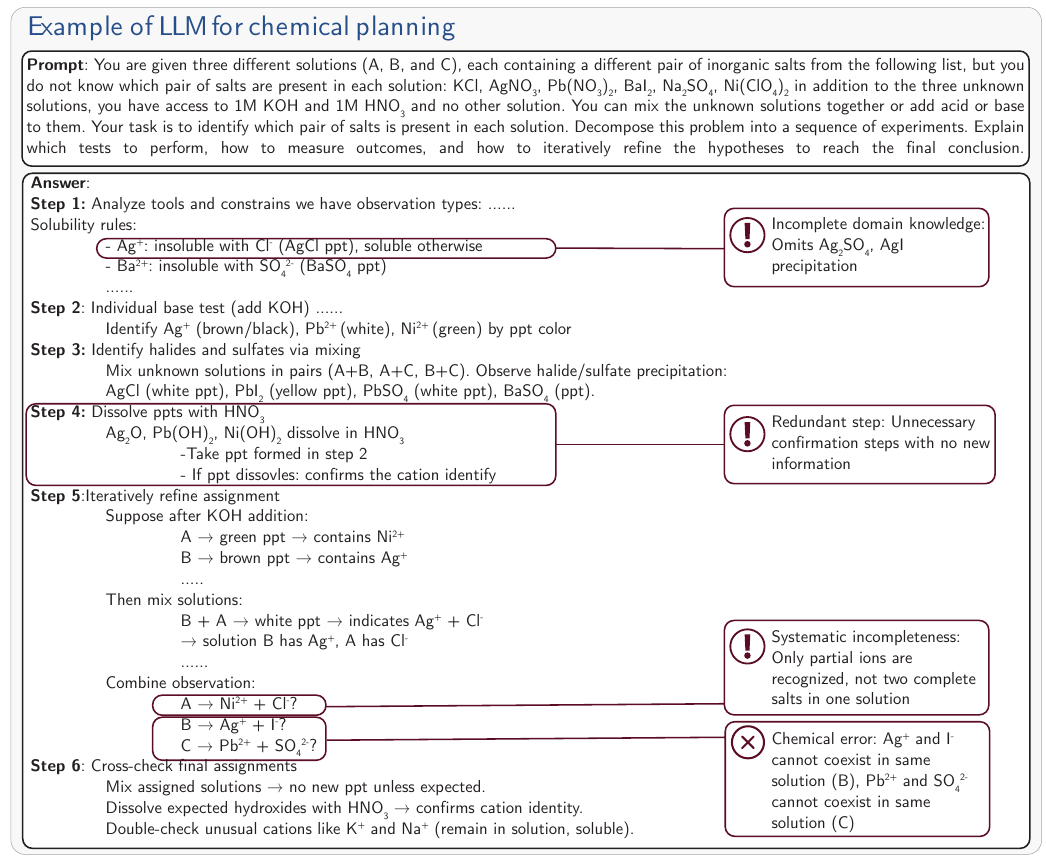}
    \caption{\textbf{An example of using \gls{llm} for chemical planning to decompose problem}. \gls{llm} decomposes a chemical problem into sequential steps with detailed procedures. We manually evaluate the plan step by step, and highlight the problematic steps in text boxes, each linked to the corresponding reason. Plan was generated by ChatGPT-5.
}
    \label{fig:planning_example}
\end{figure}

\Cref{fig:planning_example} shows how a \gls{llm} applies this deliberate, system-2-style reasoning to decompose a chemical problem into a sequence of planned steps.
A variety of strategies have been proposed to improve the reasoning capabilities of \glspl{llm} during inference. 
Methods such as \gls{cot} and least-to-most prompting guide models to decompose. 
However, their effectiveness in planning is limited by error accumulation and linear thinking patterns.\autocite{stechly2024chain}
To address these limitations, recent test-time strategies such as repeat sampling and tree search have been proposed. 
Repeated sampling allows the model to generate multiple candidate reasoning paths, encouraging diversity in thought and increasing the chances of discovering effective subgoal decompositions. \autocite{wang2024planning}
Meanwhile, tree search methods like \gls{tot} and \gls{rap} treat reasoning as a structured search, using algorithms like \gls{mcts} to explore and evaluate multiple solution paths, facilitating more global and strategic decision-making. \autocite{hao2023reasoning}

\glspl{llm} have also been applied to generate structured procedures from limited observations. For example, in quantum physics, a model was trained to infer reusable experimental templates from measurement data, producing Python code that generalized across system sizes. \autocite{arlt2024meta0designing} 

\subsubsection{Pruning of Search Spaces}

Pruning refers to the process of eliminating unlikely or suboptimal options during the search to reduce the computational burden.
Classical planners employ heuristics, value functions, or logical filters to perform pruning\autocite{bonet2012action}. 

\begin{figure}[!htbp]
    \centering
        \includegraphics[width=1\textwidth]{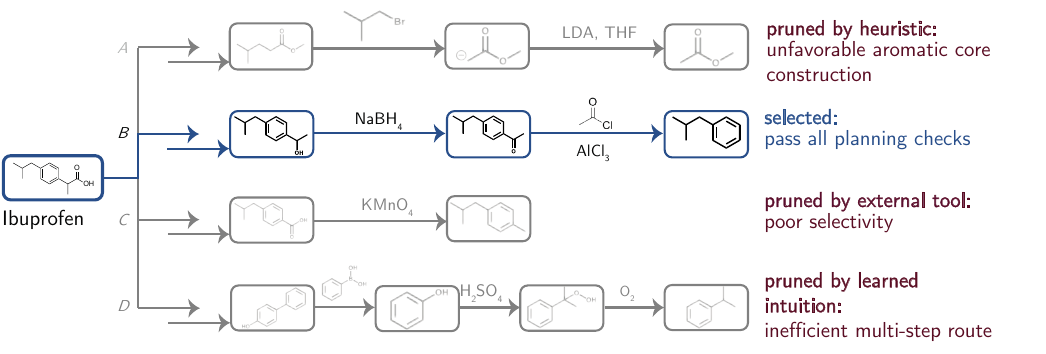}
    \caption{\textbf{\gls{gpm}-guided retrosynthesis route planning and pruning}. \glspl{gpm} can systematically evaluate and prune retrosynthetic routes using multiple reasoning capabilities to discriminate between viable and problematic approaches. The partially overlapping arrows at the start of each route indicate multiple steps. \textbf{Route A}: This route was pruned by heuristic reasoning due to the unfavorable aromatic core construction.
    \textbf{Route B}: This route was selected as it successfully passes all \gls{gpm} planning checks, demonstrating optimal synthetic feasibility.
    \textbf{Route C}: This pathway was pruned by external tools due to the poor regio-selectivity of the oxidation step.
    \textbf{Route D}: This route was pruned based on learned intuition, as it represents an inefficient multistep pathway; the route could just start with phenol instead of synthesizing it.
}
    \label{fig:planning}
\end{figure}

\Cref{fig:planning} illustrates how \glspl{llm} can support experimental planning by pruning options by emulating an expert chemist's intuition by discarding synthetic routes that appear unnecessarily long, inefficient, or mechanistically implausible. 

To further enhance planning efficacy, \glspl{llm} can be augmented with external tools that estimate the feasibility or performance of candidate plans, enabling targeted pruning of the search space before costly execution. 

Beyond external tools, \glspl{llm} can self-correct by pruning flawed reasoning steps to produce more coherent plans. At a higher level of oversight, human-in-the-loop frameworks such as \modelname{ORGANA}\autocite{darvish2025organa} incorporate expert chemist feedback to refine goals, resolve ambiguities, and eliminate invalid directions. Example~\ref{example:chemcrow_organa} presents examples illustrating how planning extends to real laboratory practice.

\subsubsection{Limitations} 

\Cref{fig:planning_example} demonstrates how \glspl{llm} can generate sophisticated-looking plans that conceal critical flaws, making complex, long-horizon chemical planning tasks particularly difficult to execute reliably \autocite{Cao2025LargeLM}. 
A limitation is that they often produce outputs that appear chemically plausible but are invalid or unsafe, due to their optimization for linguistic plausibility rather than chemical correctness and their lack of mechanistic understanding\autocite{Bran2023TransformersAL, Evans2021TruthfulAD}
Second, errors can also propagate from external tools like retrosynthesis planners, whose data and algorithmic shortcomings constrain reliability \autocite{Li2025ChemHASHA}. 
Finally, a broader limitation is the lack of grounding in experimental feedback, which creates persistent gaps between theoretical planning and practical feasibility.

\subsubsection{Open Challenges}

\begin{itemize}
    \item \textbf{Reasoning Complexity Beyond Knowledge Retrieval}: Complex chemistry problems require long, tightly interconnected chains of reasoning, where minor errors can cascade and require understanding of dynamic interactions such as temperature effects on molecular behavior. Current \glspl{llm} lack effective reasoning structures to guide domain-specific reasoning. \autocite{Ouyang2023StructuredCR, Tang2025ChemAgentSL} 
    \item \textbf{Evaluation and Feedback Bottleneck}: Current evaluation methods are often performed manually or indirectly, either relying on expert review as in \modelname{ChemCrow} \autocite{bran2024augmenting} or on pseudocode-based comparisons as in \modelname{BioPlanner}.\autocite{o2023bioplanner} Integrating feedback remains an open direction for improving the practical feasibility of generated plans.
\end{itemize}

\subsection{Experiment Execution}
Once an experimental plan is available, whether from a human scientist's idea or an AI model, the next step is to execute it. 
Regardless of its source, the plan must be translated into concrete, low-level actions for execution. 

It is worth noting that, despite their methodological differences, executing experiments \textit{in silico} (running simulations or code) and \textit{in vitro} are not fundamentally different---both follow an essentially identical workflow: Plan $\rightarrow$ Instructions $\rightarrow$ Execution $\rightarrow$ Analysis. 

The execution of \textit{in silico} experiments can be reduced to two essential steps: preparing input files and running the computational code; \glspl{gpm} can be used in both steps.\autocite{Liu2025ASA, Mendible‑Barreto2025DynaMate, Zou2025ElAgente, Campbell2025MDCrow} \textcite{Jacobs2025orca} found that using a combination of fine-tuning, \gls{cot} and \gls{rag} (see \Cref{sec:model_adaptation}) can improve the performance of \glspl{llm} in generating executable input files for the quantum chemistry software \emph{ORCA}\autocite{ORCA5}, while \textcite{Gadde2025chatbot} created \modelname{AutosolvateWeb}, an \gls{llm}-based platform that assists users in preparing input files for \gls{qmmm} simulations of explicitly solvated molecules and running them on a remote computer. 
Examples of \gls{llm}-based autonomous agents (see 
\Cref{sec:agents}) capable of performing the entire computational workflow (i.e., preparing inputs, executing the code, and analyzing the results) are \modelname{MDCrow} \autocite{Campbell2025MDCrow} (for molecular dynamics) and \modelname{El Agente Q} \autocite{Zou2025ElAgente} (for quantum chemistry).

\begin{figure}[H]
    \centering
\includegraphics[width=\textwidth]{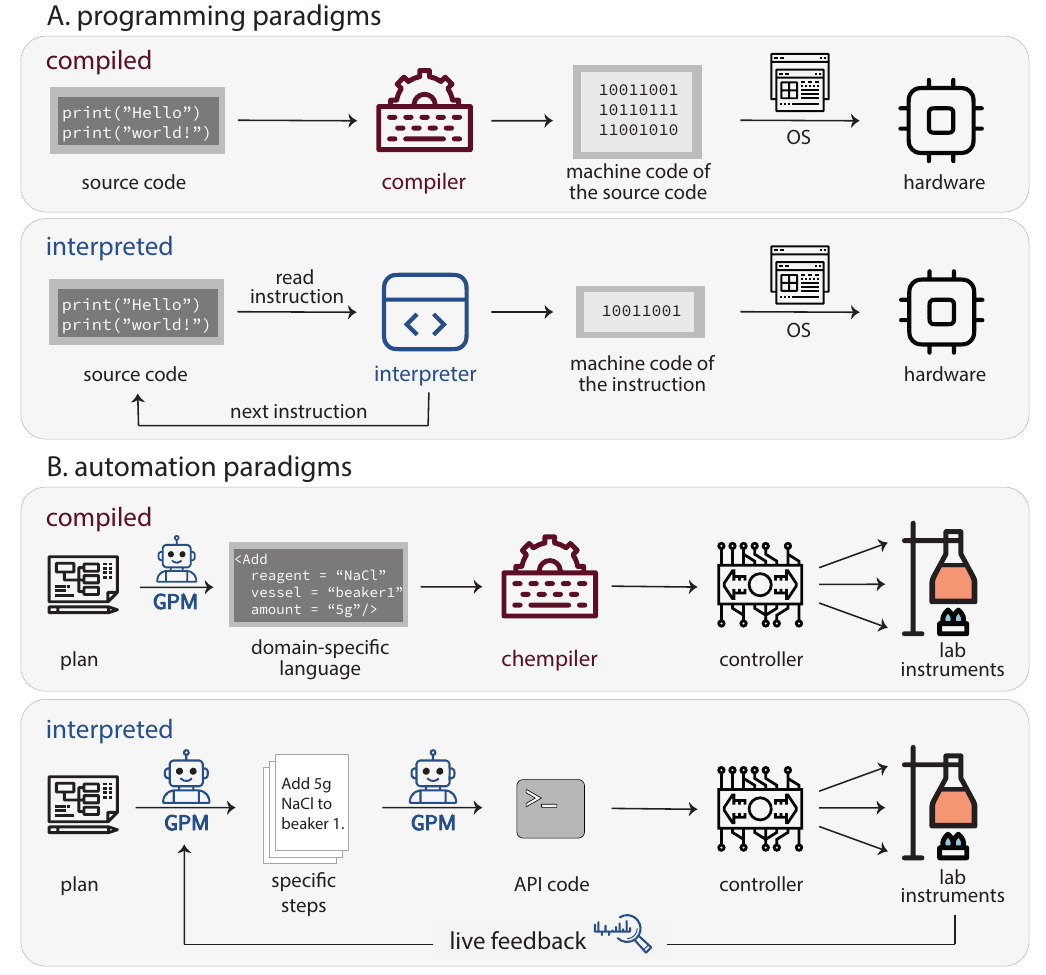}
    \caption{\textbf{Programming languages vs.\ lab automation. A) programming paradigms}: In compiled languages, the entire source code is translated ahead of time to machine code by the compiler. This stand-alone code is then given to the \gls{os}, which is responsible for scheduling and distributing tasks to the hardware. In interpreted languages, the interpreter reads and translates each line of the source code to machine code and hands it to the \gls{os} for execution. \textbf{B) automation paradigms}: In the compiled approach, a \gls{gpm} formalizes the protocol, a compiler, such as the Chempiler\autocite{steiner2019organic}, translates the formalized protocol to hardware-specific low-level steps, which the controller then executes---a central hub tasked with scheduling and distributing commands to chemical hardware. In the interpreted approach, a \gls{gpm}, acting as the interpreter, first breaks down the protocol into specific steps, then sends them (via an \gls{api}) for execution one by one. The strength of interpreted systems is dynamic feedback: after the execution of each step, the \gls{gpm} receives a signal (e.g., data, errors), which can influence its behavior for the next steps.}
    \label{fig:exec}
\end{figure}

Emerging examples show \glspl{gpm} assisting in \textit{in vitro} experiment automation. Programming language paradigms---compiled vs.\ interpreted (\Cref{fig:exec}A)---provide a useful analogy for understanding different automation approaches.

Compiled languages (\modelname{C++}, \modelname{Fortran}) convert entire programs to machine code before execution. Interpreted languages (\modelname{Python}, \modelname{JavaScript}) translate and execute instructions line-by-line at runtime. The tradeoff is that compiled languages offer higher performance and early error detection but require separate compilation steps. Interpreted languages enable rapid development and on-the-fly modification, but run slower and catch errors only during execution.

Similarly, experiment automation follows two paradigms (\Cref{fig:exec}B): \enquote{compiled automation} translates entire protocols---by human or \gls{gpm}---into low-level instructions before execution. \enquote{Interpreted automation} uses the \gls{gpm} as runtime interpreter, executing protocols step-by-step.

\subsubsection{Compiled Automation}
In \enquote{compiled automation}, protocols are formalized in high-level languages or \glspl{dsl}. A chemical compiler (\enquote{chempiler}\autocite{mehr2020universal}) converts these into low-level hardware instructions, which robotic instruments then execute (\Cref{fig:exec}B). 

While \modelname{Python} scripts serve as the \textit{de facto} protocol language, specialized \glspl{dsl} provide more structured representations.\autocite{wang2022ulsa, ananthanarayanan2010biocoder, autoprotocol2023, Park2023CMDL} For example, \gls{chidl}\autocite{mehr2020universal, xdl2023spec} describes protocols using abstract commands (\modelname{Add}, \modelname{Stir}, \modelname{Filter}) and chemical objects (\modelname{Reagents}, \modelname{Vessels}). The \modelname{Chempiler} translates \gls{chidl} scripts into platform-specific instructions based on the physical laboratory configuration.

Writing protocols in formal languages requires coding expertise. Here, \glspl{gpm} translate natural-language protocols into machine-readable formats.\autocite{Lamas2024DSLXpert, jiang2024protocode, conrad2025lowering, inagaki2023robotic,Vaucher2020AutoExtraction}
\textcite{Pagel2024LLMChemputer} introduced a multi-agent workflow (based on \modelname{GPT-4}) that can convert unstructured chemistry papers into executable code. 
The first agent extracts all synthesis-relevant text, including supporting information; a procedure agent then sanitizes the data and tries to fill the gaps from chemical databases (using \gls{rag}); another agent translates procedures into \gls{chidl} and simulates them on virtual hardware; finally, a critique agent cross-checks the translation and fixes errors.

The example above shows one of the strengths of the compiled approach: it allows for pre-validation. 
The protocol can be simulated or checked for any errors before running on the actual hardware, ensuring safety. 
Another example of \gls{llm}-based validators for chemistry protocols is \modelname{CLAIRify}.\autocite{Yoshikawa2023CLAIRify} 
Leveraging an iterative prompting strategy, it uses \modelname{GPT‑3.5} to first translate the natural-language protocol into \gls{chidl} script, then automatically verifies its syntax and structure, identifies any errors, appends those errors to the prompt, and prompts the \gls{llm} again---iterating this process until a valid \gls{chidl} script is produced.

\subsubsection{Interpreted Automation}
Interpreted programming languages support higher abstraction levels through flexible command structures. Similarly, \glspl{gpm} can translate high-level goals into concrete steps,\autocite{ahn2022can, huang2022language} acting as \enquote{interpreters} between experimental intent and lab hardware.

For example, given \enquote{titrate the solution until it turns purple}, a \gls{gpm} agent (\Cref{sec:agents}) can break this into executable steps at runtime: add titrant incrementally, read color sensor, loop until condition met. This is \enquote{interpreted automation}---conversion happens during execution, not before.

The key advantage is real-time decision-making. After each action, the system analyzes sensor data (readings, spectra, errors) and selects the next step. This enables dynamic branching and conditional logic impossible in pre-compiled protocols.

\modelname{Coscientist}\autocite{boiko2023autonomous} demonstrates interpreted automation using \modelname{GPT-4} to control liquid-handling robots. The system searches the web for protocols, reads instrument documentation, writes \modelname{Python} code in real-time, and executes experiments on physical hardware. When errors occur, \modelname{GPT-4} debugs its own code. \modelname{Coscientist} successfully optimized palladium cross-coupling reactions, outperforming Bayesian optimization in finding high-yield conditions.

\modelname{ChemCrow}\autocite{bran2024augmenting} augments \modelname{GPT-4} with $18$ expert-designed tools for tasks including compound lookup, spectral analysis, and retrosynthesis. It planned and executed syntheses of \gls{deet} and three thiourea organocatalysts (example~\ref{example:chemcrow_organa}), and collaborated with chemists to discover a new chromophore.

\subsubsection{Hybrid Approaches}
Between fully compiled and fully interpreted automation lies a hybrid approach that combines the safety and reliability of compiled protocols with the AI-driven flexibility of interpreted systems. Each experiment run follows a fixed plan for safety and reproducibility, but between runs, the plan can adapt based on the GPM’s interpretation of results. This design provides a safeguard against interpreter errors, since every generated procedure passes through formal verification before execution---catching issues like a hallucinated instruction to add \SI{1000}{\milli\liter} of solvent to a \SI{100}{\milli\liter} flask.

\modelname{ORGANA} \autocite{darvish2025organa} is an \gls{llm}-based robotic assistant following this hybrid paradigm. 
It allows human chemists to describe their experimental goal in natural language. The system can converse with the user to clarify ambiguous requests (the agent would ask \enquote{do you mean X or Y?} if the instructions are unclear). Once the goal is understood, it uses \modelname{CLAIRify} \autocite{Yoshikawa2023CLAIRify} to convert and validate the natural-language description of a chemistry experiment into a \gls{chidl} script, which can be executed on a compatible platform. 
In one case, \modelname{ORGANA} carried out a multistep electrochemistry procedure---polishing electrodes, running an experiment, and analyzing the data---involving 19 substeps that it coordinated in parallel (see example~\ref{example:chemcrow_organa}).
If an unexpected observation occurred (e.g., a solution does not change color when expected), the system can notice via image analysis and modify the plan or alert the user. 
In user studies, \modelname{ORGANA} reduced the manual labor and frustration for chemists, who could offload tedious tasks and trust the agent to handle low-level decisions.

\begin{examplebox}[label={example:chemcrow_organa}]{Planning and executing experiments with \glspl{gpm}}

\textbf{ChemCrow:} When prompted with the task \enquote{Find and synthesize a thiourea organocatalyst which accelerates a Diels-Alder reaction}, \modelname{ChemCrow} sequentially queried tools to gather information and solve the task:

\par\vspace{0.5\baselineskip}

{\leftskip=2em \enquote{First, I need to find a thiourea organocatalyst that can accelerate the Diels-Alder reaction. I will perform a websearch to find a suitable catalyst.

\textbf{WebSearch tool:} Schreiner's thiourea catalyst

Now I will obtain the \gls{smiles}. 

\textbf{Name2Smiles tool:}

\texttt{FC(F)(F)c1cc(CN(=S)Nc2cc(C(F)(F)F)cc(C(F)(F)F)c2)cc(C(F)(F)F)c1}

I will plan a synthesis for Schreiner's thiourea catalyst. 

\textbf{SynthesisPlanner tool:} Detailed synthesis plan

\includegraphics[width=0.8\textwidth]{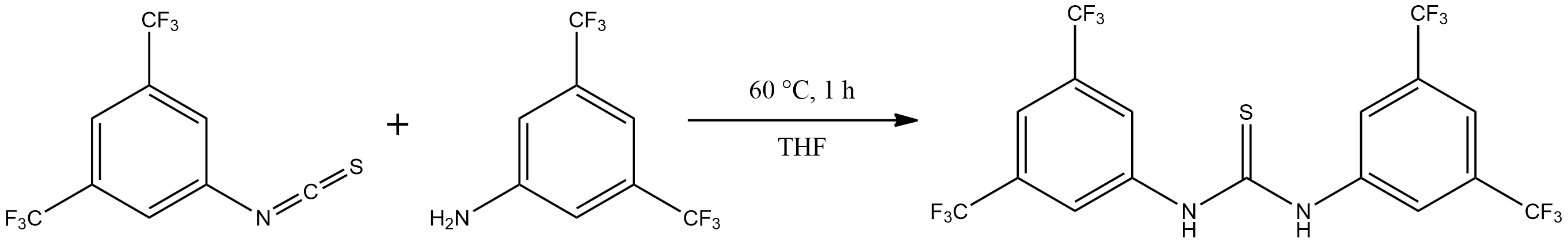}

I will execute the synthesis.

\textbf{SynthesisExecuter tool:} Successful
synthesis.}

}

\par\vspace{0.5\baselineskip}

The agent proposed to synthesize the target molecule from the corresponding isothiocyanate and aniline derivatives and executed the synthesis on the RoboRXN platform. \autocite{roborxn} The product was confirmed using HPLC/MS: \texttt{m/z 501 [M+H]} calculated, found \texttt{m/z 501.02}.

\par\vspace{\baselineskip}

\textbf{ORGANA}: \modelname{ORGANA} interacts with chemists using \glspl{llm} to derive experiment goals and makes decisions with chemists in the loop to control robots and lab devices.
\begin{itemize}
    \item \textbf{Start-up:} Users give their intention \enquote{I want to create a Pourbaix plot of a quinone, Sodium Anthraquinone-1-sulfonate.}, and start interact with \modelname{ORGANA}:
    
    \modelname{ORGANA}: What is the setup of the lab environment? What hardware and reagents are present?
    
    User: I have a \SI{2}{\milli\molar} solution of the quinone......
    
    \modelname{ORGANA}: Can you provide an example of how to run the experiment? 
    
    User: Add \SI{2}{\milli\liter} of quinone......

    \modelname{ORGANA}: What is the expected outcome of the example experiment? 
    
    User: The pH of the solution is expected ......
    
    \modelname{ORGANA}: How many experiments are we planning to run in total? 
    
    User: 6
    
    \modelname{ORGANA}: Ok, thank you! I will start the experiment......

    \item \textbf{Ambiguity and Uncertainty Mitigation:} 
    \modelname{ORGANA.Reasoner} evaluates whether experiment outcomes align with user expectations. If it can’t interpret the results, it alerts the user to check for environmental issues. The user provides feedback explaining the outcomes, which \modelname{ORGANA.Reasoner} then uses to refine its next plan:
    
    \enquote{I’m not sure what happened, but repeat the previous experiment just in case}
    
    \enquote{Nothing is wrong, carry on}

    \enquote{The pump was stuck, it’s ok again}

    \item \textbf{Experimental Reports:} 
    \modelname{ORGANA} reports the experimental logs and summaries:
    
    \enquote{A series of experiments was conducted to measure the potential of a quinone solution at various pH levels, specifically from $pH \space 7$ to $pH \space 9$......
    After performing the experiments, these are the results:
    The estimate for $pK_{a1}$ is $8.096$.
    The estimate for $pK_{a2}$ is $12.380$.
    The estimate for slope is $-60.958$.}

    \item \textbf{Results Comparison with Chemists:} Comparison of Pourbaix diagrams and first-region slope estimates from electrochemical experiments, with \modelname{ORGANA} using three measurement points per pH region and chemists using four. \modelname{ORGANA} yields results comparable to those of chemists: for $pK_{a1}$, \modelname{ORGANA} produces $8.03$ and chemists $8.02$ and for the estimated slope, \modelname{ORGANA} obtained \SI[per-mode=symbol]{-61.3}{\milli\volt\per\pH} while chemists obtained \SI[per-mode=symbol]{-62.7}{\milli\volt\per\pH}.

\end{itemize}

\end{examplebox}

\subsubsection{Limitations}
Current automation systems remain prototypes. 

Interpreted systems require frequent human intervention despite autonomy claims. They replicate known procedures but lack mechanistic understanding. Non-deterministic \gls{gpm} responses create reproducibility issues---small prompt changes yield different results, and closed-source models evolve unpredictably. Hallucinations risk incorrect planning for complex reactions. Hardware control introduces safety concerns: flexible \glspl{gpm} can devise unanticipated actions, requiring robust safeguards (\Cref{sec:safety}).

Compiled systems offer reliability but require extensive upfront formalization. The effort to translate protocols into formal languages often outweighs automation benefits for typical laboratory workflows.



\subsubsection{Open Challenges}
Self-driving laboratories orchestrated by \glspl{gpm} face technical challenges requiring research advances:\autocite{Tom2024SDL,Seifrid2022SDL}

\begin{itemize}
\item \textbf{Grounding Natural Language to Laboratory Actions.} Translating ambiguous natural-language instructions (\enquote{heat gently}) into precise, safe operations requires developing validation layers that detect physically impossible or hazardous actions before hardware execution.

\item \textbf{Universal Protocol Standards.} No widely adopted formalization standard exists. While languages like \gls{chidl} show promise, achieving interoperability across platforms requires community consensus on abstraction levels and device interfaces. \Glspl{mcp} offer a potential path forward by enforcing consistent interfaces between \glspl{gpm}, instruments, and verification layers.

\item \textbf{Autonomous Error Recovery.} Current systems cannot autonomously diagnose and recover from experimental failures. Developing general-purpose failure detection mechanisms and recovery strategies would enable truly autonomous operation.

\item \textbf{Multimodal Integration.} Chemists use diverse data types---spectra, chromatograms, TLC plates, and microscopy images. Integrating these modalities into \gls{gpm} decision-making loops remains technically challenging but essential for human-level experimental reasoning.

\item \textbf{Verification and Provenance.} Industrial and clinical applications require complete experimental provenance: every decision logged with reasoning traces, all parameters recorded, and outcomes traceable to specific model versions (\Cref{sec:safety}).
\end{itemize}

\subsection{Data Analysis}
The analysis of experimental data in chemistry remains a predominantly manual process. 

One key challenge that makes automation particularly difficult is the extreme heterogeneity of chemical data sources. 
Laboratories often rely on a wide variety of instruments, some of which are decades old, rarely standardized, or unique in configuration.\autocite{jablonka2022making} 
These devices output data in incompatible, non-standardized, or poorly documented formats, each requiring specialized processing pipelines. 
Despite efforts like \modelname{JCAMP-DX} \autocite{McDonald1988standard}, standardization attempts remain limited and have generally failed to gain widespread use. 
This diversity makes rule-based or hard-coded solutions largely infeasible, as they cannot generalize across the long tail of edge cases and exceptions found in real-world workflows.

\begin{figure}[!ht]
    \centering
\includegraphics[width=1\textwidth]{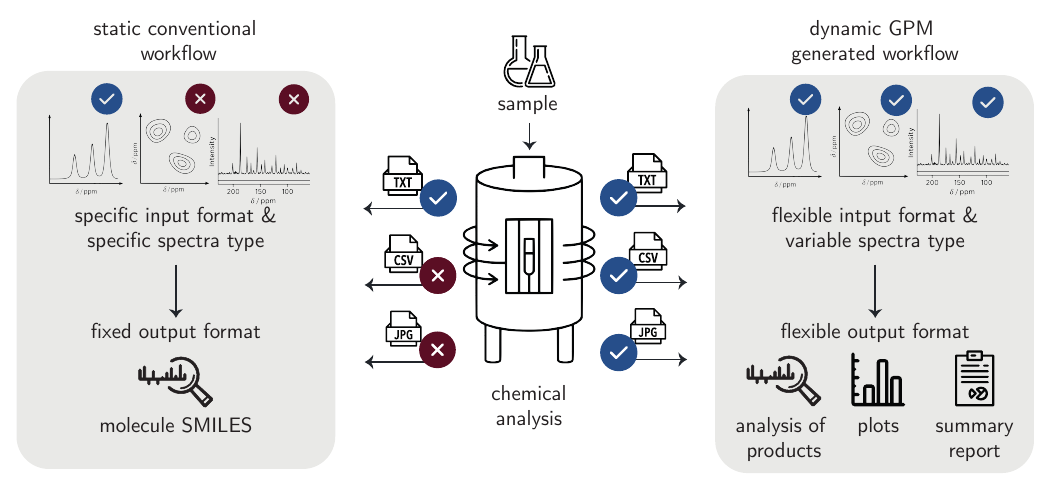}
    \caption{\textbf{Static conventional data analysis workflow vs.\ dynamic \gls{gpm} generated workflow}. Chemical analysis can be performed with a variety of instruments and techniques, resulting in many possible output data formats. The \gls{gpm} can use these diverse, raw data and process them into easy-to-understand plots, analysis, and reports. A hard-coded workflow, in contrast, is specifically made to analyze one specific data format and spectra and produces a fixed output format, e.g., the \gls{smiles} of the analyzed molecule.}
    \label{fig:anaylsis}
\end{figure}

This complexity makes chemical data analysis a promising application for \glspl{gpm} (\Cref{fig:anaylsis}). These models handle diverse tasks and formats using implicit knowledge from broad training data. In other domains, \glspl{llm} have successfully processed heterogeneous tabular data and performed classical data analysis without task-specific training.\autocite{narayan2022can, kayali2023chorus}

In chemistry, however, only a handful of studies have so far demonstrated similar capabilities. Early evaluations showed that \glspl{gpm} can support basic workflows such as the classification of \gls{xps} signals based on peak positions and intensities.\autocite{Fu2025large,decurt2024large} 
 
Spectroscopic data often appear as raw plots or images, making direct interpretation by \glspl{vlm} a more natural starting point for automated analysis. 
A broad assessment of \gls{vlm}-based spectral analysis was introduced with the \modelname{MaCBench} benchmark \autocite{alampara2024probing}, which systematically evaluates how \glspl{vlm} interpret experimental data in chemistry and materials science---including various types of spectra such as \gls{ir}, \gls{nmr}, and \gls{xrd}---directly from images. They showed that while \glspl{vlm} can correctly extract isolated features from plots, the performance substantially drops in tasks requiring deeper spatial reasoning.
To overcome these limitations, \textcite{kawchak2024high} explored two-step pipelines that decouple visual perception from chemical reasoning. First, the model interprets each spectrum individually (e.g., converting  \gls{ir}, \gls{nmr}, or \gls{ms} images into textual peak descriptions), and second, a \gls{llm} analyzes these outputs to propose a molecular structure based on the molecular formula.

More complex agentic systems extend beyond single-step analysis and attempt to orchestrate entire workflows. 
For example, \textcite{ghareeb2025robin0} developed a multi-agent system for assisting biological research with hypothesis generation (see \Cref{fig:hypothesis-generation}) and experimental analysis. Its data analysis agent \modelname{Finch} autonomously processes raw or preprocessed biological data, such as \gls{rna} sequencing or flow cytometry, by executing code in Jupyter notebooks and producing interpretable outputs. Currently, only these two data types are supported, and expert-designed prompts are still required to ensure reliable results. 

Similarly, \textcite{mandal2024autonomous} introduced \modelname{AILA}, which utilizes \gls{llm}-agents to plan, execute, and iteratively refine full \gls{afm} analysis pipelines. Compared to earlier prototypes, this systems emphasize transparency and reproducibility by producing both code and reports.
The system handles tasks such as image processing, defect detection, clustering, and the extraction of physical parameters. 

\subsubsection{Limitations}
While \glspl{gpm} offer promising capabilities for automating scientific data analysis, several concrete limitations remain. 
Recent evaluations such as \modelname{SciCode} \autocite{tian2024scicode} have shown that even \gls{sota} like \modelname{GPT-4-Turbo} frequently produce syntactically correct but semantically incorrect code, for instance, in common data analysis steps such as reading files, applying filters, or generating plots.

These technical shortcomings are further amplified by sensitivity to prompt formulation. As demonstrated by \textcite{Yan2020auto} and \textcite{alampara2024probing}, even minor changes in wording or structure can lead to drastically different results, highlighting a lack of robustness in prompt-based control. 

In practice, this means that robust prompting strategies, systematic validation, and human oversight remain essential components of any current deployment.

\subsubsection{Open Challenges}
Looking forward, several open challenges remain unresolved.

\begin{itemize}
    \item \textbf{True Chemical Reasoning}: It remains unclear whether current \glspl{llm} can perform genuine chemical analysis rather than relying on pattern-matching or shallow feature extraction. \autocite{alampara2024probing, alampara2025task}
    \item \textbf{Seamless Laboratory Integration}: No commercial systems yet provide robust, end-to-end interoperability with the diverse analytical instruments used in chemistry laboratories. Existing research prototypes support only limited data types and still depend heavily on expert curation. \autocite{ghareeb2025robin0, mandal2024autonomous}
    \item \textbf{Standardization and Real-World Validation}: Achieving production-ready systems requires progress not only in model robustness but also in data and protocol standardization, hardware integration, and thorough testing under realistic laboratory conditions. \autocite{testini2025measuring}
\end{itemize}

\subsection{Reporting}
To share insights obtained from data analysis, one often converts them into scientific publications or other forms of content, such as reports or blogs.
In this step, \glspl{gpm} can also take a central role. While writing assistance has been showcased in past works, it remains limited in scope and real-world impact.

\paragraph{From Data to Explanation} The lack of explainability of \gls{ml} predictions generates skepticism among experimental chemists\autocite{wellawatte2025human}, hindering the wider adoption of such models.\autocite{wellawatte2022model}
One promising approach to address this challenge is to convey explanations of model predictions in natural language. 
An approach proposed by \textcite{wellawatte2025human} involves coupling \glspl{llm} with feature importance analysis tools, such as \gls{shap} or \gls{lime}. 
In this framework, the \gls{llm} performs three key functions: First, it translates technical feature names into more accessible language. Second, using \gls{rag} over scientific literature, it retrieves relevant excerpts that explain the physicochemical relationships between identified molecular features and target properties. Third, it synthesizes these components into coherent natural language explanations that not only identify which structural features correlate with the property of interest, but also hypothesize \textit{why} these relationships exist based on established chemical principles from the literature. 

\paragraph{Writing Assistance} \glspl{llm} can assist with syntax improvement, redundancy identification,\autocite{khalifa2024using} figure and table captioning,\autocite{hsu2021scicap,selivanov2023medical} caption-figure matching,\autocite{hsu2023gpt04} and alt-text generation.\autocite{singh2024figura11y} Models can be personalized for specific audiences or writing styles.\autocite{li2023teach}

\glspl{llm} has also been shown to potentially help complete submission checklists. \textcite{goldberg2024usefulness} found $70\%$ of \gls{neurips} 2025 authors found \gls{llm} assistance useful for checklist completion, with the same fraction revising their submissions based on model feedback.

\subsubsection{Limitations}
\gls{llm} explanations appear credible but often lack faithfulness to underlying reasoning.\autocite{agarwal2024faithfulness} Models can reinforce existing biases through training data or prompting strategies.\autocite{kobak2025delving} While \glspl{llm} can process large datasets, they miss subtle artifacts and anomalies human researchers detect, and struggle distinguishing correlation from causation.\autocite{jin2023large}

\subsubsection{Open Challenges}
\begin{itemize}
    \item \textbf{Provenance Tracking Systems.} Developing methods to trace every claim back to specific training examples or retrieved sources. This requires architectures that maintain explicit links between generated text and source materials, enabling verification of attribution completeness.
    
    \item \textbf{Authorship Frameworks.} Defining contribution taxonomies that specify when \gls{llm} use constitutes co-authorship versus tool use. Journals and institutions need consensus guidelines for disclosure, attribution, and accountability when \glspl{llm} assist in research reporting.
\end{itemize}

\section{Accelerating Applications}
\subsection{Property Prediction} \label{sec:prediction}

\glspl{gpm} have emerged as a tool for predicting molecular and material properties. 
Current examples of \gls{gpm}-driven property prediction span both classification and regression from standardized benchmarks such as \modelname{MoleculeNet} \autocite{wu2018moleculenet}, to curated datasets targeting specific applications such as antibacterial activity \autocite{chithrananda2020chemberta} or photovoltaic efficiency \autocite{aneesh2025semantic}.

Three key methodologies have been explored to adapt these models for property prediction: prompting techniques (see \Cref{sec:prompting}), fine-tuning (see \Cref{sec:fine-tuning}) on domain-specific data, and \gls{rag} (see \Cref{sec:rag}) approaches that combine models with external knowledge bases. 

\begin{table}[htbp]
    \centering
    \caption{\textbf{Non-comprehensive list of \glspl{gpm} applied to property prediction tasks}. The table presents different models and their applications across different molecular and materials property prediction benchmarks, showing the diversity of properties (from molecular toxicology to crystal band gaps), datasets used for evaluation, modeling approaches (prompting, fine-tuning, or retrieval-augmented generation), and task types (classification or regression.)}
    \label{tab:property_prediction_models}
    \resizebox{\textwidth}{!}{%
    \begin{tabular}{lllcc}
        \toprule
        \textbf{Model} & \textbf{Property} & \textbf{Dataset} & \textbf{Approach}  & \textbf{Task}\\
        \cmidrule{1-5}
        \multirow{3}{*}{LLM-Prop\autocite{rubungo_llm-prop_2023}} & Band Gap & \multirow{3}{*}{CrystalFeatures-MP2022\autocite{rubungo_llm-prop_2023}} & \multirow{3}{*}{P} &  R  \\
        & Volume & & & R \\
        & Band gap &  & & C \\
        \midrule
        \multirow{12}{*}{LLM4SD\autocite{zheng2025large}} & Blood-brain barrier penetration & BBBP\autocite{sakiyama_prediction_2021} & \multirow{10}{*}{P} & C \\
        & FDA approval & ClinTox\autocite{wu2018moleculenet} & & C \\
        & Toxicology & Tox21\autocite{richard_tox21_2021} & & C \\
        & Drug-related side effects & SIDER\autocite{kuhn_sider_2016} & & C \\
        & HIV replication inhibition & HIV\autocite{wu2018moleculenet} & & C \\
        & $\beta$-secretase binding & BACE\autocite{wu2018moleculenet} & & C \\
        & Solubility & ESOL\autocite{wu2018moleculenet} & & R \\
        & Hydration Free Energy & FreeSolv\autocite{mobley_freesolv_2014} & & R \\
        & Lipophilicity & Lipophilicity\autocite{wu2018moleculenet} & & R \\
        & Quantum Mechanics & QM9\autocite{wu2018moleculenet} & & R \\
        \midrule
        \multirow{3}{*}{Domain Knowledge Prompt-Engineering\autocite{liu2025integrating}} & Crystal & Custom\autocite{liu2025integrating}& \multirow{3}{*}{P}   &\multirow{3}{*}{C,R}\\
        & Organic Small Molecules & PubChem\\
        & Enzymes & UniProt\\
        \midrule
        \multirow{10}{*}{MolecularGPT\autocite{liu2024moleculargpt}} & $\beta$-secretase binding & BACE \autocite{wu2018moleculenet} & \multirow{10}{*}{P} & C\\
        & HIV replication inhibition & HIV\autocite{wu2018moleculenet} & & C\\
        & Bioactivity & MUV \autocite{rohrer2009maximum} & & C\\
        & Toxicology & Tox21 \autocite{richard_tox21_2021} & & C \\
        & Toxicology & ToxCast \autocite{us2015toxicity} & & C\\
        & Blood-brain barrier penetration & BBBP\autocite{sakiyama_prediction_2021} & & C\\
        & Cytochrome P450 isozymes & CYP450 \autocite{ni2025curated} & & C\\
        & Solubility & ESOL \autocite{delaneyesol2004} & & R\\
        & Hydration Free Energy & FreeSolv \autocite{mobley_freesolv_2014}& & R\\
        & Lipophilicity& Lipophilicity\autocite{wu2018moleculenet} & & R\\
        \midrule
        \multirow{9}{*}{GPT-MolBERTa \autocite{balaji2023gptmolberta}} & Blood brain barrier penetration & BBBP\autocite{sakiyama_prediction_2021} & \multirow{9}{*}{P} & C\\
        & Toxicity & Tox21 \autocite{richard_tox21_2021} & & C\\
        & Toxicity & Toxcast \autocite{us2015toxicity} & & C\\
        & FDA Approval & Clintox \autocite{wu2018moleculenet} & & C \\
        & Solubility & ESOL \autocite{delaneyesol2004} & & R\\
        & Hydration Free Energy& FreeSolv\autocite{mobley_freesolv_2014} & &R \\
        & Lipophilicity & Lipophilicity \autocite{wu2018moleculenet}& & R\\
        & HIV replication inhibition& HIV\autocite{wu2018moleculenet} & & C\\
        &  $\beta$-secretase binding & BACE \autocite{wu2018moleculenet} & &  C\\
        \midrule
        \multirow{7}{*}{GPT-Chem\autocite{jablonka2024leveraging}} & HOMO/LUMO & QMUGs\autocite{isert_qmugs_2022} & \multirow{7}{*}{FT} & C, R \\
        & Solubility & DLS-100\autocite{mitchell_dls-100_2017} & & C, R \\
        & Lipophilicity & LipoData\autocite{jablonka2024leveraging} & & C, R \\
        & Hydration Free Energy & FreeSolv\autocite{mobley_freesolv_2014} & & C, R \\
        & Photoconversion Efficiency & OPV\autocite{jablonka2024leveraging} & & C, R  \\
        & Toxicology & Tox21\autocite{richard_tox21_2021} & & C, R  \\
        & $\text{CO}_2$ Henry Coefficients of MOFs & MOFSorb-H\autocite{lin_silico_2012} & & C, R \\        
        \midrule
        \multirow{4}{*}{\modelname{LLaMP\autocite{chiang2024llamp}}} & Bulk modulus & \multirow{4}{*}{Materials Project\autocite{riebesell2025framework}} & \multirow{4}{*}{RAG} & R \\
        & Formation energy & & & R \\
        & Electronic bandgap & & & R \\
        & Multi-element electronic bandgap & & & R \\
        \bottomrule
    \end{tabular}
    }
    \vspace{0.5em}
    \footnotesize
    \begin{minipage}{\linewidth}
        \textbf{Key:}  P = prompting; FT = fine-tuned model; RAG = retrieval-augmented generation; C = Classification; R = Regression
    \end{minipage}
\end{table}

\subsubsection{Prompting} 
Prompt engineering involves designing targeted instructions to guide \glspl{gpm} in performing specialized tasks without altering their underlying parameters.
In molecular and materials science, this strategy goes beyond simply asking a model to predict properties. Prompting strategies for molecular property prediction have evolved from foundational techniques like domain-knowledge and few-shot reasoning \autocite{liu2025integrating} to more advanced methods with multi-modal frameworks that extract interpretable rules \autocite{zheng2025large} (see \Cref{tab:property_prediction_models}).


Another emerging application of \gls{llm}-based \glspl{gpm} is their use as \enquote{feature extractors}, where they generate textual or embedded representations of molecules or materials. For instance, in materials science, \textcite{aneesh2025semantic} employed \glspl{llm} to generate text embeddings of perovskite solar cell compositions. 
These embeddings were subsequently used to train a \gls{gnn} for predicting power conversion efficiency, demonstrating the potential of \glspl{llm} to enhance feature representation in materials informatics. 
\begin{promptbox}[label={box: cot_prompting}]
    \textbf{\gls{cot} Prompting}\autocite{srinivas2024crossmodal}\\
    Prompt 1: What is the molecular structure of this chemical \gls{smiles} string? Could you describe its atoms, bonds, functional groups, and overall arrangement? \\
    Prompt 2: What are the physical properties of this molecule, such as its boiling point and melting point?\\
    \dots\\
    Prompt 14: Are there any environmental impacts associated with the production, use, or disposal of this molecule?
\end{promptbox}
Similarly, in the molecular domain, \textcite{srinivas2024crossmodal} used zero-shot \gls{llm} prompting (see \Cref{box: cot_prompting} for prompt examples) to generate detailed textual descriptions of molecular functional groups, which are used to train a small \gls{lm}. This \gls{lm} is used to compute text-level embeddings of molecules. Simultaneously, they generate molecular graph-level embeddings from \gls{smiles} string molecular graph inputs. They finally integrate the graph and text-level embeddings to produce a semantically enriched embedding.

\subsubsection{Fine-Tuning}\label{sec:prediction_FT}
\begin{figure}[H] 
    \centering
    \includegraphics[width=1\textwidth]{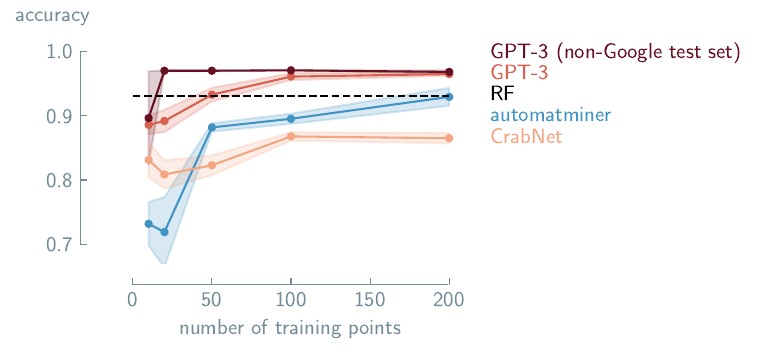}
    \caption{\textbf{Fine-tuned \modelname{GPT-3} for predicting solid-solution formation in high-entropy alloys} Performance comparison of different \gls{ml} approaches as a function of the number of training points. Results are shown for \modelname{Automatminer} (blue), \modelname{CrabNet} transformer (orange),  fine-tuned \modelname{GPT-3} (red), with error bars showing standard error of the mean. The non-Google test set shows the fine-tuned \modelname{GPT-3} model tested on compounds without an exact Google search match (dark red). The dashed line shows performance using random forest (RF). \modelname{GPT-3} achieves comparable accuracy to traditional approaches with fewer training examples. Data adapted from \textcite{jablonka2024leveraging}}
    \label{fig:gptchem}
\end{figure}

\paragraph{\gls{lift}} \textcite{dinh2022lift} showed that reformulating regression and classification as \gls{qa} tasks enables the use of an unmodified model architecture while improving performance (see \Cref{sec:fine-tuning} for a deeper discussion of \gls{lift}). 
In recognizing the scarcity of experimental data and acknowledging the persistence of this limitation, \textcite{jablonka2024leveraging} designed a \gls{lift}-based framework using \modelname{GPT-3} fine-tuned on task-specific small datasets (see \Cref{tab:property_prediction_models}). 
They demonstrated that fine-tuned \modelname{GPT-3} can match or surpass specialized \gls{ml} models in various chemistry tasks (as shown in \Cref{fig:gptchem}.

In a follow-up to \textcite{jablonka2024leveraging}'s work, \textcite{vanherck2025assessment} systematically evaluated this approach across 22 diverse real-world chemistry case studies using three open-source models. They demonstrate that fine-tuned \glspl{llm} can predict various material properties. For example, they achieved $96\%$ accuracy in predicting the adhesive free-energy of polymers, outperforming traditional \gls{ml} methods like random forest ($90\%$ accuracy). The \glspl{llm} can also work with non-standard inputs, like for predicting protein phase separation, where raw protein sequences could be directly input without pre-processing and achieve $95\%$ prediction accuracy. At the same time, when training datasets were very small (15 data points), the predictive accuracy of all fine-tuned models was lower than the random baseline (e.g., MOF synthesis). These case studies preliminarily demonstrate that these models can achieve predictive performance with some small datasets, work with various chemical representations (\gls{smiles}, \gls{mof}id, and \gls{iupac} names), and can outperform traditional \gls{ml} approaches for some material property prediction tasks.

In the materials domain, \modelname{LLMprop} fine-tunes \modelname{T5}\autocite{raffel2020exploring} to predict crystalline material properties from text descriptions generated by \modelname{Robocrystallographer}\autocite{ganose2019robocrystallographer}. By discarding \modelname{T5}’s decoder and adding task-specific prediction heads, the approach reduces computational overhead while leveraging the model’s ability to process structured crystal descriptions. 

Fine-tuning has also been applied to non-\gls{llm} architectures, specifically to \glspl{ssm} like Mamba (see \Cref{sec:example_architectures}). By pre-training on 91 million molecules, the Mamba-based model \modelname{$\text{O}_{SMI}-{\text{SSM}-}336\textit{M}$} outperformed transformer methods (\modelname{Yield-BERT}\autocite{krzyzanowski2025exploring}) in reaction yield prediction (e.g., Buchwald-Hartwig cross-coupling) and achieved competitive results in molecular property prediction benchmarks.\autocite{soares2025mamba-based} 

\subsubsection{Agents}

Caldas Ramos et al.\ introduced \modelname{MAPI-LLM}, a framework that processes natural-language queries about material properties using an \gls{llm} to decide which of the available tools, such as the Materials Project \gls{api}, the Reaction-Network package, or Google Search, to use to generate a response. \autocite{Jablonka2023} \modelname{MAPI-LLM} employs a \gls{react} prompt (see \Cref{sec:arch_agents} to read more about \gls{react}), to convert prompts such as  \textit{\enquote{Is $Fe_2O_3$ magnetic?}} or \textit{\enquote{What is the band gap of Mg(Fe3O3)2?}} into queries for Materials Project \gls{api}. 
The system processes multi-step prompts through logical reasoning, for example, when asked \textit{\enquote{If Mn2FeO3 is not metallic, what is its band gap?}}, the \gls{llm} system creates a two-step workflow to first verify metallicity before retrieving the band gap.

Building on this foundation of agent-based materials querying, \textcite{chiang2024llamp} advanced the approach with \modelname{LLaMP}, a framework that employs \enquote{hierarchical} \gls{react} agents to interact with computational and experimental data. This \enquote{hierarchical} framework employs a supervisor-assistant agent architecture where a complex problem is broken down and tasks are delegated to domain-specific agents.

\subsubsection{Limitations}\label{sec:property_core_limits}

\begin{figure}[H]
    \centering
    \includegraphics{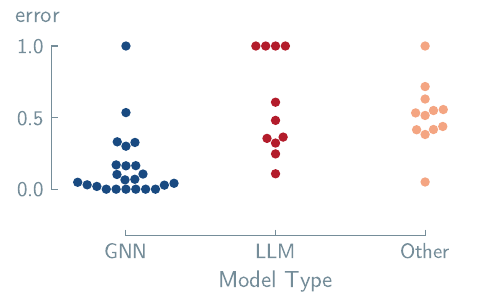}
    \caption{\textbf{Normalized error distributions for materials property prediction models across different architectures}. Each point represents the normalized error of a model on a specific property prediction task. Normalization was achieved with min/max values of each dataset to produce a range of errors between 0 and 1. The first column (blue) shows \gls{gnn} based models, the second column (red) displays \gls{llm} approaches, and the third column (orange) represents other baseline methods and \gls{sota} models, including \modelname{CrabNet}. \autocite{Wang_2021} Lower values indicate better predictive performance. Data adapted from \textcite{alampara2024mattext}}
    \label{fig:property_limitations}
\end{figure}

A significant challenge for \glspl{gpm} in chemistry lies in managing dataset limitations and selecting appropriate chemical representations. Practical applications often suffer from highly unbalanced datasets, where examples of optimal materials are vastly outnumbered by poor-performing ones, forcing difficult compromises that can diminish model performance \autocite{vanherck2025assessment}. The choice of how a material is represented (\gls{smiles} notation versus \gls{iupac}) also critically impacts performance, indicating that data preprocessing remains a crucial consideration.

Beyond data issues, architectural constraints present another barrier as illustrated in \Cref{fig:property_limitations}. \textcite{alampara2024mattext} reveal that while \glspl{llm} are effective for tasks relying on compositional information, they struggle to interpret geometric or spatial data when it is encoded in text. This suggests a fundamental limitation of transformer-based architectures for applications requiring spatial reasoning. Consequently, the conventional assumption that performance can be universally improved by scaling up model size or pre-training data is challenged \autocite{frey2023neural}. Such scaling may not overcome the inherent bias against geometric understanding.

\subsubsection{Open Challenges}
\begin{itemize}
    \item \textbf{Encoding 3D Structure}: Developing methods to effectively represent and integrate geometric, spatial, and structural information to overcome their inherent bias toward compositional data.
    \item \textbf{Dynamic Knowledge Integration}: Move beyond static fine-tuning to establish a reliable, real-time framework that can incorporate new, evolving scientific findings without catastrophic forgetting.
    \item \textbf{Fundamental Architectural Shifts}: Exploring whether many of the existing \gls{gpm} architectures are sufficient or if new, specialized ones are needed to capture complex relationships in molecular and materials science.
\end{itemize}

\subsection{Molecular and Material Generation} \label{sec:mol_generation}

\begin{figure}[htbp!]
    \centering
    \includegraphics[width=1\textwidth]{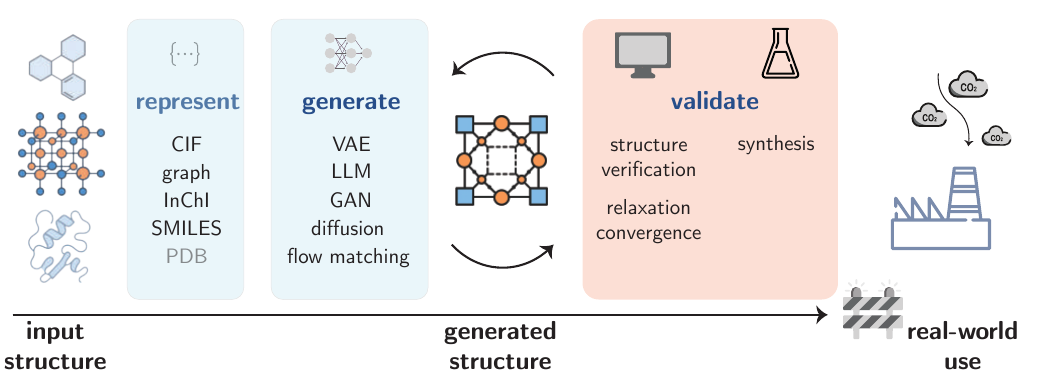}
    \caption{\textbf{Pipeline for molecular and materials generation}. The workflow begins with input structures represented in various formats, which are used to train \gls{ml} models to generate novel molecular and material structures. The generated structures should undergo a feedback loop through validation processes before being applied in the real world. Blue boxes indicate well-established areas of the pipeline with mature methodologies, while the red box represents critical bottlenecks. Currently, the largest gaps in the process are in the representation and validation steps.}
    \label{fig:generation}
\end{figure}

\noindent Early work in molecular and materials generation relied heavily on unconditional generation, where models produce novel structures without explicit guidance. These methods excel at exploring chemical space broadly but lack specific control. 
Conditional generation, in contrast, uses explicit prompts or constraints (e.g., property targets, structural fragments) to steer \glspl{gpm} toward meaningful molecule or material designs. 
Beyond the generation step, as \Cref{fig:generation} shows, critical bottlenecks persist in synthesizability and physical consistency at the validation stage.

\subsubsection{Generation}\label{sec:generation}

\paragraph{Prompting}
While zero-shot and few-shot prompting strategies offer a flexible approach to molecule generation, benchmark studies \autocite{guo2023large} reveal a significant performance gap compared to specialized models. \textcite{guo2023large} systematically evaluated \glspl{llm} like \modelname{GPT-4}, finding that while they could generate syntactically valid molecules ($89\%$ validity), their accuracy in meeting specific property targets was low ($< 20\%$). These performance gaps highlight the lack of chemical structure-property relationships possessed by \glspl{llm}.

\paragraph{Fine-Tuning} To overcome the limitations of prompting, fine-tuning has been adopted in molecular and materials generation, much like its use in property prediction with \gls{lift}-based frameworks (see \Cref{sec:fine-tuning} for a deeper explanation of \gls{lift} and \Cref{sec:prediction_FT} for a discussion of \gls{lift} applied to property prediction tasks). 

A representative example is \gls{icma}, which combines retrieval-augmented in-context learning with fine-tuning \autocite{li2025large}. This method avoids extensive domain-specific pre-training by retrieving relevant examples to guide the model. On standard benchmarks, \gls{icma} nearly doubled baseline performance for tasks like generating molecules from text descriptions (Cap2Mol). However, its strong dependence on retrieved examples raises questions about its ability to generalize to entirely novel molecular scaffolds. Other fine-tuning approaches have aimed to further improve accuracy on the  \enquote{Cap2Mol} task \autocite{li2024molreflect, lin2025property}. 

\paragraph{Diffusion and Flow Matching}
Diffusion and flow-based generative models provide a flexible framework for generating diverse structures by operating directly on latent representations \autocite{zhu20243m-diffusion}. A more complex challenge involves generating crystalline materials, which require modeling both discrete (atom type) and continuous (atomic position) variables. To address this, hybrid architectures like \modelname{FLowLLM} have emerged, combining the strengths of \glspl{llm} and flow matching. In this approach, a fine-tuned \gls{llm} learns a base distribution of crystals from text-based representations, which is then refined through rectified flow matching to optimize the atomic structure \autocite{sriram2024flowllm}.

\paragraph{Reinforcement Learning and Preference Optimization}

Translating \gls{gpm} generated outputs to the real world requires designing molecules and materials with specific target properties. \gls{rl} and preference optimization techniques\autocite{lee2024fine-tuning} have emerged as potential solutions for this challenge. 

\modelname{CrystalFormer-RL} uses \gls{rl} fine-tuning to optimize \modelname{CrystalFormer}\autocite{cao2024space}, a transformer-based crystal generator, with rewards from discriminative models (e.g., property predictors)\autocite{cao2025crystalformer-rl}. Here, \gls{rl} fine-tuning is shown to outperform supervised fine-tuning, enhancing both novel material discovery and retrieval of high-performing candidates from the pre-training dataset.

\gls{era} introduces a different optimization paradigm. \autocite{chennakesavalu2025aligning} Unlike \gls{ppo} or \gls{dpo}, \gls{era} uses gradient-based objectives to guide word-by-word generation with explicit reward functions, converging to a physics-inspired probability distribution that allows fine control over the generation process.

\paragraph{Agents} Agent-based frameworks leveraging \glspl{llm},  explained in \Cref{sec:agents}, have emerged as approaches for autonomous molecular and materials generation, demonstrating capabilities that extend beyond simple prompting or fine-tuning by incorporating iterative feedback loops, tool integration, and human-\gls{ai} collaboration. 
The \modelname{dZiner} framework exemplifies this approach for the inverse design of materials, where agents input initial \gls{smiles} strings with optimization task descriptions and generate validated candidate molecules by retrieving domain knowledge from the literature.\autocite{ansari2024dziner} 
It also uses domain-expert surrogate models to evaluate the required property in the new molecule/material. 

\subsubsection{Validation}
\paragraph{General validation} The most fundamental validation approaches use cheminformatics tools like \modelname{RDKit} to verify molecular validity. More sophisticated validation involves quantum mechanical calculations to compute molecular properties such as formation energies\autocite{kingsbury2022flexible}. 

The gold standard for validation is experimental synthesis, but significant gaps exist between computational generation and laboratory realization.
Retrosynthesis prediction algorithms attempt to bridge this gap by evaluating synthetic accessibility and proposing potential synthesis routes (see \Cref{sec:retrosynthesis}). 
However, these methods still face limitations in accurately predicting real-world synthesizability \autocite{zunger2019beware}.

\paragraph{Conditional Generation Validation} Beyond establishing the general validity of generated molecules, evaluation methods can assess both their novelty relative to training data and their ability to meet specific design goals. For inverse design tasks, such as optimizing binding affinity or solubility, the \textit{de novo} molecule generation benchmark GuacaMol differentiates between \textit{distribution-learning} (e.g., generating diverse, valid molecules) and \textit{goal-directed} optimization (e.g., rediscovering known drugs or meeting multi-objective constraints) \autocite{brown2019guacamol}. 
In the materials paradigm, frameworks such as \modelname{MatBench Discovery} evaluate analogous challenges such as stability, electronic properties, and synthesizability, but adapt metrics to periodic systems, such as energy above hull or band gap prediction accuracy\autocite{riebesell2025framework}. 

\subsubsection{Limitations}
Current generative pipelines for molecules and materials, as illustrated in \Cref{fig:generation}, face significant bottlenecks that limit their immediate real-world application. A primary limitation is the performance gap in conditional generation; while methods like fine-tuning and reinforcement learning have improved control, \glspl{llm} often lack the inherent chemical understanding to accurately meet specific property targets. Furthermore, the validation stage remains a critical challenge. Even when a structure is computationally valid, assessing its synthesizability and physical consistency relies heavily on approximations, with a significant gap remaining between in-silico prediction and experimental realization.

\subsubsection{Open Challenges}
\begin{itemize}
    \item \textbf{Robust Conditional Generation} Developing models that can reliably generate novel, diverse structures that satisfy complex, multi-objective constraints (e.g., high stability, specific electronic properties, and synthesizability) without over-relying on retrieved training data examples.
    \item \textbf{Bridging the Synthesis Gap} Creating new validation metrics that better predict the synthetic feasibility of generated materials and molecules, moving beyond structural similarity toward realistic pathway prediction.
    \item \textbf{Unified Multi-Scale Generation} Developing frameworks that are capable of simultaneously modeling different representations (e.g., text, graphs, 3D structures) and scales (e.g., from molecules to crystalline materials) within a single generative process
 \end{itemize}

\subsection{Retrosynthesis}\label{sec:retrosynthesis}

The practical utility of \glspl{gpm} for molecules and materials design is constrained by uncertainty in synthetic feasibility. 
Early work showed that attention-based models learn meaningful reaction representations, enabling accurate reaction outcome classification and prediction \autocite{schwaller2021mapping}.

Building on this, generation pipelines increasingly integrate synthesizability and retrosynthetic guidance via domain tools and \glspl{gpm} \autocite{liu2024multimodal}. 
For example, \textcite{sun2025synllama} adapted open \glspl{llm} to propose retrosynthetic routes and identify purchasable building blocks for experimentally validated SARS-CoV-2 Mpro inhibitors.

\glspl{llm} are also being fine-tuned as chemistry assistants for experimental guidance.
\textcite{zhang2025large} used a two-stage process, supervised fine-tuning with reaction and retrosynthesis \gls{qa} followed by \gls{rlhf}) to optimize reaction conditions, achieving an unreported Suzuki–Miyaura cross-coupling in 15 runs.

Predictive retrosynthesis has also extended to the inorganic domain. \textcite{kim2024large} demonstrated that fine-tuned \modelname{GPT-3.5}, and \modelname{GPT-4} can predict both the synthesizability of inorganic compounds from their chemical formulas and select appropriate precursors for synthesis, achieving performance comparable to specialized \gls{ml} models with minimal development time and cost. 

Toward autonomy, \textcite{bran2024augmenting} developed \modelname{ChemCrow}, an \gls{llm}-based system that autonomously plans and executes the synthesis of novel compounds by integrating specialized tools like a retrosynthesis planner (see \Cref{sec:planning} to read more about this capability of \modelname{ChemCrow} and its limitations) and reaction predictors. 
This approach mirrors the iterative experimental design cycle employed by human chemists, but is equipped with the scalability of automation. 

\subsubsection{Limitations}

\glspl{gpm} inherit the fundamental constraints associated with domain-specific models for retrosynthesis, which primarily concern data and evaluation methodologies. The prevailing reliance on patent-derived data introduces a significant bias, as these datasets are dominated by a limited set of reaction types and predominantly feature single-step examples. Furthermore, the absence of negative results and the frequent omission of experimental conditions can mislead model training \autocite{Lee2024noise, Voinarovska2023when, Saebi2023onthe, Maloney2023negative, Tadanki2025dissecting}. The evaluation paradigms suffer from a parallel shortcoming, as they are largely designed for single-step routes and fail to adequately capture the complexities of real-world, multi-step case studies \autocite{TorrenPeraire2024models, maziarz2025re, liu2024evaluating}.

A further significant challenge is data leakage. The vast scale of data used for training these models often leads to saturated benchmarks, where high performance may reflect memorization rather than genuine generalization.

Finally, while standalone \glspl{gpm} shows limited efficacy for multi-step retrosynthesis, its integration into agentic frameworks introduces a distinct set of limitations. Current evaluations are poorly suited to assess the critical components of such systems, such as planning, reasoning, and tool-calling capabilities. The focus remains predominantly on final answer correctness, which impedes large-scale optimization and correction of agents without extensive human oversight, typically resulting in the use of a fragile and error-prone LLM-as-Judge. 

These evaluation gaps, combined with the inherent limitations in long-term planning and chemical understanding of the underlying \glspl{llm}, present substantial obstacles to the development and practical application of agentic systems for this retrosynthetic task.

\subsubsection{Open Challenges}

\begin{itemize}
    \item \textbf{Data Quality and Bias}: To build more robust and generalizable models, the field must integrate higher-quality data from peer-reviewed scientific literature. This requires the development of standardized extraction schemas and data cleaning protocols to create a balanced and reliable training corpus \autocite{rios2025llm, mehr2020universal}.
    \item \textbf{Robust and Meaningful Evaluation}: There is a critical need for evaluation methodologies that prevent data leakage and test genuine model understanding. Benchmarks must be designed to go beyond memorization, instead probing a model's capacity for chemical reasoning, generalization to novel structures, and robust problem-solving.
    \item \textbf{Benchmarking Complex Planning}: Truly assessing a \gls{gpm}'s utility requires benchmarks that mirror real-world complexity. This involves evaluating performance in open-ended retrosynthetic planning and developing frameworks that assess a model's sequential decision-making.
\end{itemize}

\subsection{GPMs as Optimizers} \label{sec:llm-optimizers}

\begin{figure}[htb]
    \centering
    \includegraphics[width=1\textwidth]{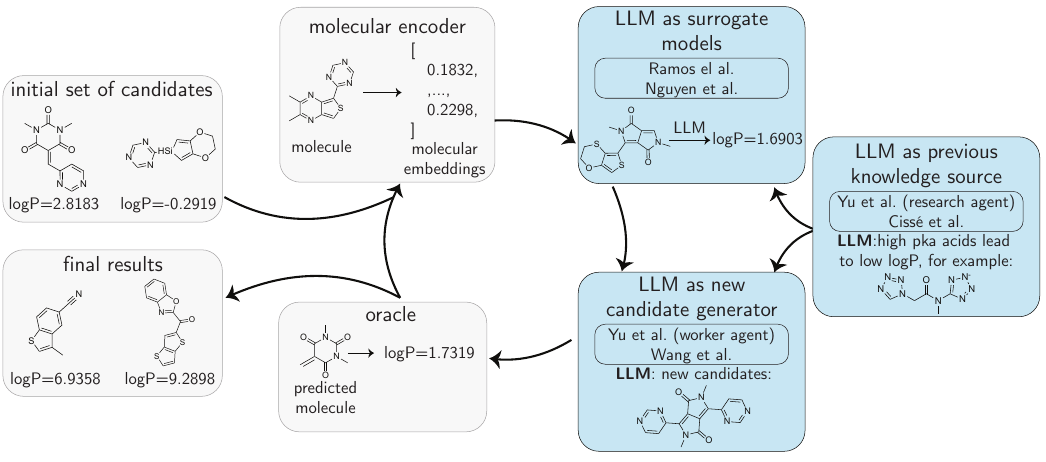}
    \caption{\textbf{Overview of the iterative optimization loop that mirrors the structure of the optimization section}. The blue boxes contain the different roles that the \glspl{llm} play in the loop, and which are described in the main text. References in which the use of \glspl{llm} for that step are detailed inside the small boxes inside each of the components of the loop. The example shown is about obtaining molecules with high \texttt{logP}.}
    \label{fig:optimization}
\end{figure}

Discovering novel compounds and reactions in chemistry and materials science has long relied on iterative trial-and-error processes rooted in existing domain knowledge \autocite{Taylor2023brief}. As noted in \Cref{sec:retrosynthesis}, these methods accelerate discovery, and optimization targets variables such as conditions or binding affinity. 
Nevertheless, the overall process remains slow and labor-intensive.
Traditional data-driven methods aim to address these limitations by combining predictive \gls{ml} models with optimization frameworks such as \gls{bo} or \glspl{ea}. 
These frameworks balance exploration of uncharted regions in chemical space with exploitation of known high-performing regions \autocite{Li2024sequential, Hse2021gryffin, Shields2021bayesian, Griffiths2020constrained, RajabiKochi2025adaptive}.

Recent advances in \glspl{llm} have been explored for targeting optimization challenges in chemistry and related domains \autocite{fernando2023promptbreeder0, yang2023large, chen2024instruct}. 
A commonly referenced advantage is that \glspl{llm} may process optimization tasks posed in natural language, which may facilitate knowledge incorporation, candidate comparison, and interpretability in some settings.
This can align with chemical problem-solving, where complex phenomena, such as reaction pathways or material behaviors, are often poorly captured by standard nomenclature; however, they can still be intuitively explained through natural language. 
Moreover, \glspl{gpm} can offer flexibility when problem definitions change, whereas many classical models require retraining.
Encoding domain-specific knowledge---including reaction rules, thermodynamic principles, and structure-property relationships---into structured prompts may allow \glspl{llm} to complement domain expertise with their ability to navigate complex chemical optimization problems.

Current \gls{llm} applications in chemistry optimization vary in scope and methodology. 
Many studies integrate \glspl{llm} into \gls{bo} frameworks, where models guide experimental design by predicting promising candidates \autocite{rankovic2023bochemian}. Others employ \glspl{ga} or hybrid strategies that combine \gls{llm}-generated hypotheses with computational screening \autocite{cisse2025language0based}.  

\subsubsection{LLMs as Surrogate Models}

A prominent \gls{llm}-driven strategy positions these models as surrogate models within optimization loops \autocite{yu2025collaborative}. 
Surrogate models---often implemented as \gls{gpr}---learn from prior data to approximate costly feature-outcome landscapes, which are often computationally expensive and time-consuming to evaluate, thereby guiding the acquisition.
A proposed advantage of the \glspl{llm} in this role is relatively better low-data performance compared to classical \gls{ml} optimization methods.
Their \gls{icl} capability enables task demonstration with minimal prompt examples while leveraging chemical knowledge from pre-training to generate accurate predictions. 
This allows \glspl{llm} to compensate for sparse experimental data effectively. 
However, \glspl{llm} are less robust than \gls{gpr} due to their tendency to hallucinate.

\textcite{ramos2023bayesian} demonstrated the use of this low-data regime through a framework that combines \gls{icl} using only one example in the prompt with a \gls{bo} workflow. 
Their \gls{bo}-\gls{icl} approach uses $k$-shot examples formatted as question-answer pairs, where the \gls{llm} generates candidate solutions conditioned on prior successful iterations. 
These candidates are ranked using an acquisition function, with top-$k$ selections integrated into subsequent prompts to refine predictions iteratively. 
Interestingly, in the reported experiments, the method performed competitively, matching top-1 accuracies on the evaluated benchmarks compared to classical \gls{bo} methods. This emphasizes the potential of \glspl{llm} as accessible, \gls{icl} optimizers when coupled with well-designed prompts.

Trying to address limitations in base \glspl{llm}’ inherent chemical knowledge---particularly their grasp of specialized representations like \gls{smiles} or structure-property mappings---\textcite{yu2025collaborative} introduced a hybrid architecture augmenting pre-trained \glspl{llm} with task-specific embedding and prediction layers. 
These layers, fine-tuned on domain data, aligned latent representations of input-output pairs (denoted as \texttt{<x>} and \texttt{<y>} in prompts), enabling the model to map chemical structures and properties into a unified, interpretable space. Crucially, the added layers were reported to improve chemical reasoning without sacrificing the flexibility of \gls{icl}, allowing the system to adapt to trends across iterations, similarly to what was done by \textcite{ramos2023bayesian}. 
In their evaluations of molecular optimization benchmarks, such as the \gls{pmo} \autocite{gao2022sample}, they revealed that \gls{llm}-based methods matched or improved over conventional methods, including \gls{bo}-\gls{gp}, \gls{rl} methods, and \gls{ga}.

\subsubsection{LLMs as Next Candidate Generators}

Recent studies explore the possibility of using \glspl{llm} in enhancing \glspl{ea} \autocite{lu2024generative} and \gls{bo} \autocite{amin2025towards} frameworks by leveraging their embedded chemical knowledge and ability to integrate prior information, thereby potentially reducing computational effort while improving output quality \autocite{lu2024generative}.
Within \glspl{ea}, \glspl{llm} refine molecular candidates through mutations (modifying molecular substructures) or crossovers (combining parent molecules). 
In \gls{bo} frameworks, they serve as acquisition policies, utilizing surrogate model predictions---both mean and uncertainty---to select optimal molecules or reaction conditions for evaluation.

For molecule optimization, \textcite{yu2025collaborative} introduced \modelname{MultiMol}, a dual-\gls{llm} system where one model proposes candidates and the other supplies domain knowledge (see \Cref{sec:opt-llm-know-source}). 
By fine-tuning the \enquote{worker} \gls{llm} to recognize molecular scaffolds and target properties, and expanding the training pool to include a pre-training dataset of $\sim1$ million samples, they report hit rates (percentage of generated molecules that meet the target properties under a certain threshold) exceeding $90\%$ on their evaluation set.

Similarly, \textcite{wang2024efficient} developed \modelname{MoLLEO}, integrating an \gls{llm} into an \gls{ea} to replace random mutations with \gls{llm}-guided modifications. Here, \modelname{GPT-4} generated optimized offspring from parent molecules that converged faster to high fitness scores in the reported experiments. 
Notably, while domain-specialized models (\modelname{BioT5}, \modelname{MoleculeSTM}) underperformed, the general-purpose \modelname{GPT-4} excelled---suggesting the utility of \glspl{llm} may be context-dependent.

\subsubsection{LLMs as Prior Knowledge Sources} \label{sec:opt-llm-know-source}

A key advantage of integrating \glspl{llm} into optimization frameworks is their ability to encode and deploy prior knowledge within the optimization loop. 
As illustrated in \Cref{fig:optimization}, this knowledge can be directed into either the surrogate model or candidate generation module, potentially reducing the number of optimization steps required through high-quality guidance if the feedback is useful.

For example, \textcite{cisse2025language0based} introduced \modelname{BORA}, which contextualizes conventional black-box \gls{bo} using an \gls{llm}. \modelname{BORA} maintains standard \gls{bo} as the core driver, but strategically activates the \gls{llm} when progress stalls. 
This leverages the model’s \gls{icl} capabilities to hypothesize promising search regions and propose new samples, regulated by a lightweight heuristic policy that manages costs and incorporates domain knowledge (or user input). 
Evaluations on synthetic benchmarks, such as optimizing the catalyst mixture for hydrogen generation, show that \modelname{BORA} was reported to accelerate exploration, and outperforms the \gls{llm}-\gls{bo} hybrid methods evaluated.

To potentially enhance the task-specific knowledge of the \gls{llm} generating feedback, \textcite{zhang2025large} fine-tuned a \modelname{Llama-2-7B} model using a multitask \gls{qa} dataset. This dataset was created with instructions from \modelname{GPT-4}. 
The resulting model served as a human assistant or operated within an active learning loop, thereby accelerating the exploration of new reaction spaces (see \Cref{sec:retrosynthesis}). 
However, as the authors note, even this task-specialized \gls{llm} produces suboptimal suggestions for optimization tasks. 
They remain prone to hallucinations and cannot assist with unreported reactions, but still improved upon pure classical methods in most of the evaluated applications.

\subsubsection{Approaching Optimization Problems}

Published works explore different ways of using \glspl{llm} for optimization problems in chemistry, from simple approaches, such as just prompting the model with some initial random set of experimental candidates and iterating \autocite{ramos2023bayesian}, to fine-tuning models in \gls{bo} fashion \autocite{rankovic2025gollum0}. 
A pragmatic initial step is to try a purely \gls{icl} approach, which allows one to obtain a first signal rapidly. 
Such results help determine whether a more complex, computationally intensive approach is necessary or whether prompt engineering is reliable for the application. 
Fine-tuning can be used as a way to enhance the chemical knowledge of the \glspl{llm} and can lead to improvements in optimization tasks where the model requires such knowledge to choose or generate better candidates. 
Fine-tuning might not be a game-changer for other approaches that rely more on sampling methods \autocite{wang2025llm0augmented}.

While some initial works showed that \glspl{llm} trained specifically on chemistry perform better for optimization tasks \autocite{kristiadi2024sober}, other works showed that a \glspl{gpm} such as \modelname{GPT-4} combined with an \gls{ea} outperformed all other models \autocite{wang2024efficient, macknight2025pre0trained}.

\subsubsection{Limitations}

Current \glspl{llm} for chemical optimization, to date, exhibit a pronounced volatility, rarely yielding stable performance gains. 
Their outcomes are highly sensitive to prompt phrasing, and a lack of calibrated uncertainty estimates prevents their use for principled acquisition strategies. \autocite{liu2024large}
Furthermore, these models frequently produce hallucinations and violate critical constraints, leading to invalid chemical structures or the proposal of unsafe and infeasible experimental conditions. 
Alleged improvements often hinge on narrow benchmarks, extensive pre- and post-processing, or access to downstream oracles, rather than on robust chemical reasoning. 
Within hybrid pipelines, the specific contribution of the \gls{llm} becomes difficult to isolate.

\subsubsection{Open Challenges}

\begin{itemize}
    \item \textbf{Correct Prompt Sensitivity}: A key challenge is achieving prompt invariance, where the ranking of candidates remains stable under controlled rephrasings. Promising research directions include developing rigorous invariance tests and moving beyond discrete tokens to leverage models' more stable hidden states \autocite{rankovic2025gollum0}.
    \item \textbf{More robust evaluations}: Future work should develop more robust frameworks that assess oracle realism, conduct thorough leakage audits to detect data contamination, and integrate uncertainty-aware metrics to better quantify risk and reliability.
    \item \textbf{Ablations for the specific LLM role}: There is a critical need for systematic ablation studies that isolate core \gls{llm} performance from enhancements like \gls{rag}, external scorers, or tool use. Such studies are essential for fairly comparing general and fine-tuned \glspl{llm} and for understanding the source of performance gains.
\end{itemize}

\section{Implications of GPMs: Education, Safety, and Ethics}
\label{sec:implications}

As \glspl{gpm} integrate into chemistry education and research, they bring transformative potential alongside critical challenges. Responsible deployment requires addressing pedagogical practices, chemical-specific safety risks, and ethical considerations unique to scientific knowledge systems.

\subsection{Education} \label{sec:education}

\glspl{gpm} open up potential for chemistry education, ranging from personalized tutoring and adaptive feedback to supporting educators in material preparation and assessment. \autocite{Mollick2024}
Specialized systems could tailor explanations to individual learning needs, help students rehearse concepts, and lower barriers to coding and data analysis.\autocite{Mollick2024, Sharma2025role, Du2024} Coupled with \gls{ar}, they could provide safe laboratory simulations before physical experiments. For educators, \glspl{gpm} promise to reduce workload through automated feedback on open-ended responses and individualized assignment generation.\autocite{Kortemeyer2024, gao2024towards}

Despite this potential, current implementations remain fragmented. Most uses rely on general-purpose interfaces without curricular alignment\autocite{Subasinghe2025, Tsai2023, shao2025unlocking}, and while first prototypes integrate \glspl{gpm} into structured systems with \gls{rag} \autocite{perez2025large, Jablonka2023}, learning tracking, or interactive tutoring modes \autocite{li2025tutorllm0, wang2025learnmate0, AnthropicEducation}, chemistry-focused platforms are rare. 
Studies further suggest that while models can handle simple recall or formatting tasks, they struggle with deeper reasoning, diagram interpretation, and robust grading.\autocite{handa2025education, baral2025drawedumath0, kharchenko2024advantages} 

\subsubsection{Limitations}

\glspl{gpm} lack transparency in sources and confidence, enabling hallucinations to mislead students.\autocite{marcus2025will, kosmyna2025your} Over-reliance risks deskilling: students complete assignments without developing chemical understanding or analytical thinking.\autocite{dung2025learning, Sharma2025role} Models are unreliable for grading chemistry work, particularly free-text reasoning and molecular diagrams.\autocite{baral2025drawedumath0, Kortemeyer2024}

\subsubsection{Open Challenges}

\begin{itemize}
    \item \textbf{Responsible Integration Strategies.} Developing pedagogical frameworks for thoughtful \gls{gpm} integration in chemistry curricula, adapting assessment to emphasize critical evaluation over rote completion.
    
    \item \textbf{Chemistry-Specific Systems.} Building platforms that integrate chemical structure recognition, spectroscopic data interpretation, and laboratory safety protocols into tutoring and assessment workflows.
    
    \item \textbf{Critical Competence Training.} Teaching students to evaluate chemical plausibility, identify hallucinated reactions, and creatively extend model outputs.\autocite{klein2025rethink}
    
    \item \textbf{Validated Assessment Tools.} Developing robust grading systems for chemistry-specific tasks: mechanism proposals, spectral interpretation, synthesis planning.
\end{itemize}

\subsection{Safety} \label{sec:safety}

\begin{figure}[!htbp]
    \centering
    \includegraphics[width=1\linewidth]{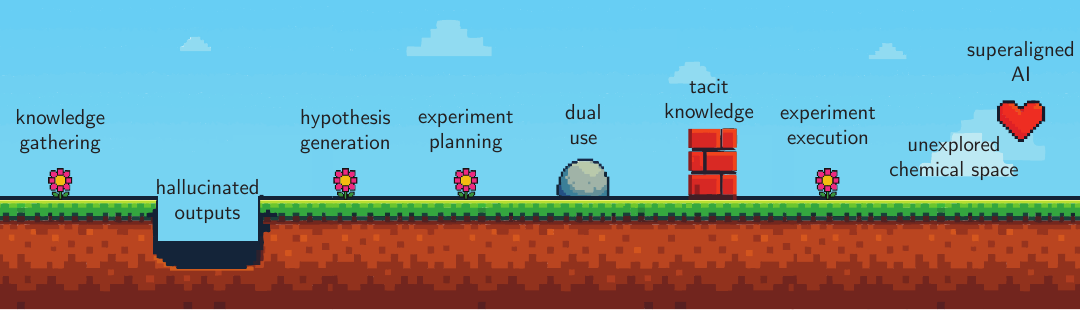}
    \caption{\textbf{A conceptual schematic depicting \gls{ai} risk factors in chemical science}. As one traverses through the scientific process, illustrated as a game, there are various obstacles to encountering \gls{ai} exacerbated risks. The path to superaligned chemical \gls{ai}-assistants is obfuscated by unexplored chemical space.}
    \label{fig:safety-overview}
\end{figure}

\noindent \glspl{gpm} in chemistry are double-edged: they accelerate discovery while potentially amplifying chemical safety risks. From democratizing hazardous synthesis knowledge to enabling autonomous dangerous compound production, these systems lower barriers to misuse (\Cref{fig:safety-overview}). Real-world constraints---specialized equipment, regulated reagents, tacit expertise---currently limit risks, but \glspl{gpm} are progressively overcoming these barriers.

\subsubsection{Chemical-Specific Risk Amplification}

\paragraph{Information Access and Synthesis Planning} 
\glspl{gpm} can lower cognitive barriers to accessing dangerous chemical knowledge. \textcite{urbina2022dual} demonstrated that molecular generators like \modelname{MegaSyn} could design toxic agents including VX when reward functions prioritize toxicity. Recent evaluations of \modelname{GPT-4} and \modelname{Claude 4 Opus} for biological threat creation found that while information is already accessible, \glspl{gpm} improve troubleshooting and help acquire tacit knowledge previously limiting non-experts.\autocite{openai2024building, anthropic2025system} \textcite{he2023control} showed \glspl{llm} generate pathways for explosives (\gls{petn}) and nerve agents (sarin). While precursor supply chains remain regulated, \glspl{gpm} that incrementally lower technical barriers act as force multipliers for malicious actors. Chemistry-specific agents like \modelname{ChemCrow}\autocite{bran2024augmenting} and \modelname{Coscientist}\autocite{boiko2023autonomous} demonstrate how \glspl{gpm} can plan and execute complex syntheses---capabilities extending to dangerous compounds.

\paragraph{Hallucinations in Chemical Contexts} 
\glspl{gpm} can hallucinate non-existent reactions, falsify safety protocols, and generate plausible-sounding but incorrect procedures.\autocite{pantha2024challenges, ji2023survey} In chemistry, these errors risk laboratory accidents or synthesis failures that waste resources. Temporal misalignment due to static training data means chemical knowledge becomes outdated as new reactions, hazards, and regulations emerge.

\paragraph{Autonomous Laboratory Risks} 
Autonomous laboratories controlled by \glspl{gpm} face cybersecurity vulnerabilities. Compromised systems could be manipulated to synthesize hazardous compounds. The timeline mismatch between \gls{ai} capabilities and infrastructure security creates risk windows where adversaries exploit inadequately protected systems.\autocite{rouleau2025risks, dean2025security}

\subsubsection{Existing Approaches to Safety} Red-teaming evaluations identify vulnerabilities but are reactive, not preventive. \modelname{ChemCrow}'s safeguards block known controlled substances, yet \textcite{he2023control} demonstrated these protections rely on post-query web searches rather than embedded constraints---easily circumventable. Machine unlearning attempts to remove hazardous knowledge face fundamental challenges in chemistry: defining \enquote{dangerous chemical knowledge} is context-dependent (bleach is benign alone, hazardous when combined), and removal risks degrading model utility for legitimate research. Alignment through \gls{rlhf} reduces harmful outputs but remains vulnerable to jailbreaks\autocite{kuntz2025os-harm, yona2024stealing, lynch2025agentic} and fails to generalize to novel threats (models might refuse known toxins while suggesting precursors).

\subsubsection{Solutions}

Chemical \gls{gpm} oversight should focus on \enquote{advanced chemical models} meeting specific risk thresholds: models trained on sensitive synthesis data, systems demonstrating autonomous synthesis planning capabilities, or agents with laboratory control interfaces. This targeted approach avoids burdening low-risk research while capturing concerning systems---analogous to proposed biological model regulations.\autocite{bloomfield2024ai}

Developing chemistry-aware guardrails requires systems that evaluate chemical plausibility, safety, and regulatory compliance before providing synthesis information. These technical safeguards should integrate real-time hazard assessment checking outputs against controlled substance databases, context-aware refusal mechanisms understanding legitimate versus dangerous use cases, precursor tracking identifying suspicious synthesis pathway queries, and audit logging enabling forensic analysis of concerning interactions. Establishing review processes for \gls{gpm}-assisted chemical research analogous to institutional biosafety committees provides essential institutional oversight. Researchers deploying chemical \glspl{gpm} with autonomous capabilities should undergo safety review, particularly for autonomous synthesis planning and execution, large-scale screening for bioactive compounds, or novel chemical class exploration without expert oversight.

International coordination remains critical for effective chemical \gls{gpm} governance. Unlike self-regulation, which risks conflicts of interest and competitive pressure toward laxity, an international oversight body---analogous to organizations governing nuclear materials or biological weapons---could harmonize safety standards across jurisdictions. Such an \gls{iaio} comprising \gls{ai} researchers, chemists, policymakers, and security experts could establish pre-approval requirements for high-risk chemical \gls{gpm} development, similar to institutional review boards in biomedical research.\autocite{trager2023international} Precedents exist in \gls{cern}'s approach to balancing civilian research with dual-use risk management and nuclear non-proliferation treaties tying market access to compliance.

Transparency requirements should mandate that chemical \gls{gpm} developers publicly report red-teaming results for dangerous synthesis queries, safety evaluation methodologies and thresholds, known vulnerabilities and mitigation strategies, and training data sources documenting chemical knowledge scope. The fundamental tension remains: \glspl{gpm} optimize reactions or design toxins with equal facility, but their black-box nature complicates accountability. Progress requires not just safeguards but deliberate constraints on chemical \gls{gpm} capabilities in high-risk domains.

\subsection{Ethics} \label{sec:ethics}

\gls{gpm} deployment in chemistry raises ethical concerns requiring careful consideration: from bias in chemical knowledge to environmental costs of computation.\autocite{crawford2021atlas}

\subsubsection{Environmental Impact of GPMs}

The computational requirements for training and deploying \glspl{gpm} contribute to environmental degradation through excessive energy consumption and carbon emissions. \autocite{spotte-smith2025considering, nature2023carbon} 
These computational resources are often powered by fossil fuel-based energy sources, which directly contribute to anthropogenic climate change. \autocite{strubell2019energy} 
The emphasis on \gls{ai} research has superseded some commitments made by big technological companies to carbon neutrality. 
For example, Google rescinded its commitment to carbon neutrality amid a surge in \gls{ai} usage (65$\%$ increase in carbon emissions between 2021--24) and funding.\autocite{bhuiyan2025google} 
Additionally, the water consumption for cooling data centers that support these models is another concern, particularly in regions facing water scarcity. \autocite{mytton2021data}

The irony is particularly stark when considering that in the chemical sciences, these models are used to address climate-related challenges, such as the development of sustainable materials or carbon capture technologies. 
As a scientific community, we must grapple with the questions about the sustainability of current \gls{ai} development trajectories and consider more efficient and renewable approaches to model development and deployment. \autocite{kolbert2024obscene}

\subsubsection{Copyright Infringement and Plagiarism Concerns}

\glspl{gpm} are typically trained on a vast corpora of copyrighted scientific literature, patents, and proprietary databases, often without explicit permission, a practice that has sparked legal disputes, such as \textit{Getty Images v.\ Stability \gls{ai}}, where plaintiffs allege unauthorized scraping of protected content. \autocite{kirchhubel2024intellectual} Developers at \modelname{OpenAI} claimed in a statement to the \gls{uk} House of Lords that training \gls{sota} models is \enquote{impossible} without copyrighted material, highlighting a fundamental tension between \gls{ip} law and \gls{ai} advancement. \autocite{openai2023written} 
In the chemical sciences, this challenge persists through the training of models on experimental results from pay-walled journals. 
A potential resolution to this in the scientific sphere lies in the expansion of open-access research frameworks. 
Initiatives like the \gls{cas} Common Chemistry database provide legally clear training data while maintaining attribution.
\glspl{llm} have shown a high propensity to regurgitate elements from their training data. 
When generating text, models may reproduce near-verbatim fragments of training data without citation, effectively obscuring intellectual contributions.\autocite{bender2021dangers} 
While some praise \glspl{gpm} for overcoming \enquote{blank-page syndrome} for early-career scientists \autocite{altmae2023artificial}, others warn that uncritical reliance on their outputs risks eroding scientific rigor.\autocite{donker2023dangers}

\subsubsection{Biases}

\glspl{gpm} inherit and amplify harmful prejudices and stereotypes present in their training data, which pose significant risks when applied translationally to medicinal chemistry and biochemistry. \autocite{spotte-smith2025considering, yang2024demographic, omiye2023large} 
These models can perpetuate inaccurate and harmful assumptions based on race and gender about drug efficacy, toxicity, and disease susceptibility, leading to misdiagnosis and mistreatment. \autocite{chen2023algorithmic} 
Historical medical literature contains biased representations of how different populations respond to treatments, and \glspl{gpm} trained on such data can reinforce these misconceptions. \autocite{mittermaier2023bias}
The problem extends to broader contexts in chemical research. 
Biased models can influence research priorities, funding decisions, and the development of chemical tools in ways that systematically disadvantage the most vulnerable populations \autocite{dotan2019value0laden}. 

\paragraph{Solutions} The problem of bias can be best addressed through top-down reform. 
The data necessary to train unbiased models can only exist if clinical studies of drug efficacy are conducted on diverse populations in the real world.\autocite{criado-perez2019invisible} 
To complement improved data collection, standard evaluations for bias testing must be developed and mandated prior to deployment of \glspl{gpm}.

\subsubsection{Access and Power Concentration} 

Although \glspl{gpm} have the potential to democratize access to advanced chemical research capabilities, they may also concentrate power in the hands of a few large companies that control the frontier models. 
This concentration raises concerns about equitable access to research tools, particularly for researchers in smaller institutions with limited resources.\autocite{satariano2025a1i1}

As a community, we should ensure that the benefits of \glspl{gpm} in chemistry remain broadly accessible via public compute, open-weight models, and portable tooling. We should also insist on fair access terms and transparent benchmarks so that no single provider can gatekeep core research.

\section{Outlook and Conclusions}
\begin{figure}[htb]
    \centering
    \includegraphics[width=1\linewidth]{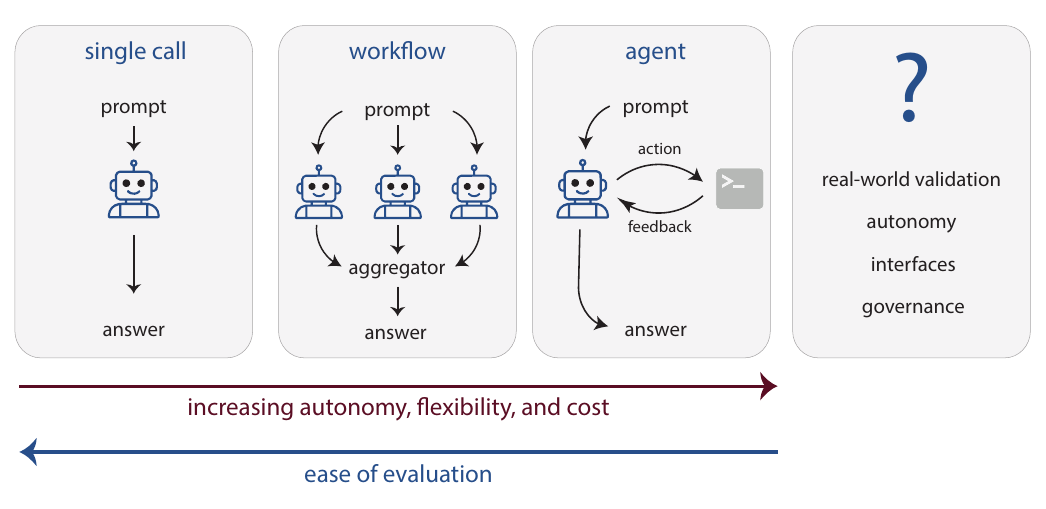}
    \caption{\textbf{Evolution of \gls{gpm}-powered systems.} First \gls{gpm} applications directly used zero- or few-shot prompted or fine-tuned \glspl{gpm}. More complex tasks could be solved by combining multiple \glspl{gpm} in workflows where the execution trajectory is pre-determined. In agents, \glspl{gpm} autonomously decide on the execution trajectory and, in this way, enable researchers to address open-ended tasks. Moving forward, coupling the validation closer to real-world objectives with further increased validation in better, custom user interfaces will enhance the impact of \glspl{gpm}. To ensure safe and ethical deployment, the community must engage with the broader public and policymakers to devise governance strategies.}
    \label{fig:conclusions}
\end{figure}

As we have explored in this review, \glspl{gpm}---especially \glspl{llm}---hold remarkable promise for the chemical sciences. 
The field has evolved from simple, single model calls to sophisticated workflows that chain multiple calls together, and further to the development of autonomous agents that determine their own problem-solving trajectories (see \Cref{fig:conclusions}).

\paragraph{Applicability of GPMs} With all this flexibility, it becomes tempting to deploy \glspl{gpm} across a wide range of problems in the chemical sciences. However, as discussed in the previous sections, they are not always a direct replacement for existing approaches. In some cases, using a \gls{gpm} may be unnecessary and inefficient. That said, there are clear scenarios where \glspl{gpm} offers distinct advantages. When data is unstructured or \enquote{fuzzy}---such as text-based descriptions or incomplete experimental records lacking explicit molecular structures---\glspl{gpm}, become valuable alternatives where specialized pipelines like \glspl{gnn} would be inapplicable. Similarly, in extremely low-data regimes where specialized pipelines or domain-specific foundation models would overfit, \gls{gpm} embeddings can provide more robust predictions than random baselines. Finally, for dynamic environments that require iterative reasoning, interpretation, and adjustment of the system states, \gls{gpm}-powered agents are well-suited. In contrast, when clean datasets are available and inductive biases are well-understood, domain-specific models typically offer superior performance and interpretability.

\paragraph{Open Questions} Several fundamental questions remain unresolved. We do not understand if there are fundamental limits to what can be predicted, given the inherent unpredictability of chemical systems and the reliance on tacit knowledge. It remains unclear whether generative models achieve a genuine understanding of chemistry or merely excel at pattern recognition, a distinction obscured by the lack of methods to quantify chemical reasoning \autocite{alampara2024probing}. This uncertainty challenges the need for human-interpretable output, implying that the latent knowledge within hidden embeddings may be more significant \autocite{hao2024training}. Moreover, we do not know what new datasets and techniques need to be developed, given the fact that the knowledge we extract from already published data is approaching a limit.\autocite{silver2025welcome} New data most likely will be generated by agents learning from their own experience.
To optimize systems, we need to better understand the underlying structure of chemical data.
In many other fields, data distributions have been shaped by special driving forces. 
For example, evolution led to a direct link between sequence and fitness in biological sequences, which makes such datasets special. In chemistry, it is unclear what the \enquote{driving force} that shapes datasets is. 

It is also unclear how quickly these innovations will permeate the average chemistry lab, where the adoption of new technology depends on more than just predictive prowess. 
And we also do not know yet how we should interface with those models for the greatest effectiveness.
In addition, it is also unclear how far acceleration can take us, as nature imposes some natural speed limits: Some experiments simply take their time. 

\paragraph{Looking Ahead} Overall, this landscape suggests a future rich with opportunity. And there are already some practical use cases for which we provide tutorials at \url{https://gpmbook.lamalab.org/tutorial-literature.html}.
But realizing the potential impact of \glspl{gpm} demands clear-eyed caution: while it is now deceptively easy to spin up prototypes, transforming them into robust, reliable tools is a far more arduous task. \autocite{sculley2014machine}
More crucial still is our need for rigorous measurement and feedback---whether in the construction of evaluation suites, the calibration of reward functions for reinforcement learning, or the design of sensible governance. 
No single discipline can shoulder this alone; chemists, policy experts, and computer scientists must broaden their ranks and collaborate. 
This is particularly true since science has always benefited from embracing a diversity of approaches. 
While \gls{gpm}-powered approaches for science, such as \enquote{AI scientists}, hold promise, a myopic focus on \enquote{AI scientists} might lead to \enquote{scientific monocultures}.\autocite{Savitsky_2025} 
We hope this review lowers the barrier to entry to the background and applications of \glspl{gpm} in the chemical sciences, inviting a wider spectrum of contributors to adopt a systems-science mindset---and, in doing so, to help harness the best of what \glspl{gpm} can offer for tackling the chemical sciences’ most persistent and pressing challenges.

\section*{Acknowledgments}

This work was supported by the Carl-Zeiss Foundation. 

\noindent A.A.\ acknowledges financial support for this research by the Fulbright U.S. Student Program, which is sponsored by the U.S. Department of State and the German-American Fulbright Commission. Its contents are solely the responsibility of the author and do not necessarily represent the official views of the Fulbright Program, the Government of the United States, or the German-American Fulbright Commission. 
 
\noindent M. S.-W.\ was supported by Intel and Merck via the AWASES research center. 

\noindent A.M.'s work was supported by the Helmholtz Association within the framework of the Helmholtz Foundation Model Initiative (SOL-AI project).

\noindent G.P.'s work was supported by the HPC Gateway measure of the Helmholtz Association.

\noindent K.M.J.\ is part of the NFDI consortium FAIRmat funded by the Deutsche Forschungsgemeinschaft (DFG, German Research Foundation) – project 460197019.

\noindent We thank Mimi Lavin and Maximilian Greiner for their feedback on a draft of this article.

\section*{Author contributions}
N. A.\ was the primary contributor for the \enquote{Building principles of GPMs} section. Including its writing and figures (excluding the \enquote{Model Level Adaptation} subsection). N. A.\ also reviewed the \enquote{Introduction,} \enquote{Shape and structure of chemical data,} \enquote{Evals,} \enquote{Implication of GPMs,} and \enquote{Property prediction and material generation in accelerating applications} sections. \\

\noindent A. A.\ was the primary contributor to the \enquote{existing \glspl{gpm} for chemical science}, \enquote{property prediction}, \enquote{molecule and materials generation}, \enquote{retrosynthesis}, \enquote{safety}, and \enquote{ethics} sections, including their writing and figures. Edited introduction, evaluations, architectures, knowledge gathering, experiment execution, and education sections. \\ 

\noindent M.R.-G. was the primary contributor to the \gls{ai} scientist overview, the hypothesis generation, and the \gls{llm} as optimizers sections, and helped in reviewing the entire manuscript. \\

\noindent A.M.\ was the primary contributor to the introduction and data sections, and the main contributor to the knowledge gathering and reporting sections within the applications section, with minor contributions to the architecture and the safety sections. Has drafted the initial outline of the article. Reviewed the architecture sections, the safety section, the hypothesis generation, the data analysis sections and contributed to the review of the LLM-as-optimizers section.\\

\noindent M.S.-W.\ was the primary contributor to the evaluation, education and data analysis sections. M.S.-W. also reviewed the shape of chemical data, hypothesis generation, experiment execution, reporting and safety section. Unified all figures. Kept track of upcoming deadlines. \\ 

\noindent M.\ S.\ was the primary contributor to the writing of experimental planning section and related figure. And also helped in reviewing \enquote{Knowledge gathering,} \enquote{Property prediction,} and \enquote{LLMs as optimizers} sections. \\

\noindent G.P.\ was the primary contributor to the \enquote{Adapting and Using General Purpose Models} section, including its writing and table. G.P.\ additionally reviewed the \enquote{The Shape and Structure of Chemical Data,} \enquote{Data Analysis,} \enquote{Reporting} and \enquote{Molecular and Material Generation} sections. \\

\noindent A.A.A.\ was the primary contributor to the \enquote{Experiment Execution} section, including its figure, and a minor contributor to the post-training subsection. A.A.A.\ reviewed the \enquote{Introduction,} \enquote{Experiment Planning,} \enquote{Molecule/Materials Generation} and \enquote{Education} sections, edited \enquote{Existing GPMs for Chemical Science}, \enquote{The Shape and Structure of Chemical Data} and part of \enquote{Building Principles of GPMs} sections, created the glossary, cleaned the references and ensured that most of them can be easily accessed via a DOI or a URL. \\

\noindent K.M.J.\ initiated and led the project. K.M.J.\ reviewed and edited all sections. 

\section*{Conflicts of Interest}
K.M.J.\ has been a paid contractor for OpenAI as part of the Red-Teaming Network.

\clearpage
\section*{Biographies}
\paragraph{Kevin Maik Jablonka}
Kevin Jablonka leads an independent research group at the Helmholtz Institute for Polymers in Energy Applications of the University of Jena and the Helmholtz Center Berlin, where he focuses on designing materials that work in the real world using data-driven techniques. He belongs to a new generation of scientists with a broad skill set, combining expertise in chemistry, materials science, and artificial intelligence. Recently, Kevin has been at the forefront of applying Large Language Models to chemistry and materials science.

\paragraph{Mara Schilling-Wilhelmi}
Mara Schilling-Wilhelmi obtained her Master's degree in Chemistry in 2024 from the Friedrich-Schiller University Jena, Germany. She then joined as a Ph.D.\ student under the supervision of Kevin M. Jablonka in the same year. Her research focuses on developing novel benchmarks for comparing state-of-the-art models in the chemistry domain, utilizing Vision-Language Models (VLMs) to extract information from published data, and applying this data to predict chemical reactions. 

\paragraph{Nawaf Alampara}
Nawaf Alampara is a PhD student in Kevin Maik Jablonka's group. Before this, he received a master's degree from IIT Bombay. He is interested in analyzing general-purpose AI systems to understand their limitations --- where they fail---and interpreting them to uncover why they fail. His goal is to use these insights to design AI systems that can advance science and research.

\paragraph{Adrian Mirza}
Adrian Mirza obtained his Master's degree in Chemical Engineering in 2023 from the Technical University of Technology, Delft, Netherlands. He is pursuing a doctorate at the HIPOLE Jena and Friedrich-Schiller University Jena, Germany, under the supervision of Kevin Maik Jablonka. His research focuses on applications of machine learning in the chemistry domain, primarily focused on data research and spectroscopy.

\paragraph{Meiling Sun}
Meiling Sun obtained her Master's degree from Jilin University, China. She is a PhD student in Kevin Maik Jablonka's group since 2025. Her current research focuses on understanding LLMs on chemical domain and scientific data extraction by LLMs.

\paragraph{Gordan Prastalo}
Gordan Prastalo obtained his Master's in Computer Science in 2025 at the Technical University of Munich. He is pursuing a doctorate at the Helmholtz Institute for Polymers in Energy Applications Jena and Friedrich-Schiller University Jena, under the supervision of Kevin Maik Jablonka. His research focuses on the application of machine learning in chemistry, with the primary focus on molecular representation.

\paragraph{Anagha Aneesh}
Anagha Aneesh obtained her Bachelor's in Chemistry and Physics in 2024 from Haverford College. She then completed a Fulbright Research Fellowship at the Friedrich-Schiller University Jena under the supervision of Kevin Maik Jablonka. Currently, she is a chemistry PhD student at Stanford University with a research focus on small molecule design using machine learning and LLM-based approaches.

\paragraph{Martiño Ríos-García}
Martiño Ríos-García is a PhD student at the Friedrich-Schiller-Universität Jena under the supervision of Kevin Maik Jablonka. His research focuses on the development of evaluation methods for General Purpose Models in the chemical and materials science fields. Before he obtained his master's and bachelor's degrees in Theoretical Chemistry and Computational Modeling, and Chemistry, respectively, from the University of Santiago de Compostela.

\paragraph{Ali Asghar Aghajani}
Ali Asghar Aghajani received his Bachelor's from Sharif University of Technology in Iran, studying Chemistry and Physics, and his Master's from Tehran University, in Computational Chemistry. Currently, he is a PhD student at Friedrich-Schiller University under the supervision of Kevin Maik Jablonka. His research is about combining active-learning techniques with laboratory automation to explore the chemical space.

\clearpage
\printbibliography

\clearpage
\begin{figure}
    \centering
    \includegraphics[]{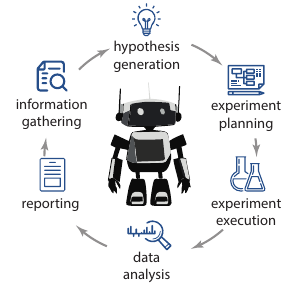}
    \caption{Table of contents figure.}
\end{figure}

\clearpage
\glsaddall
\printnoidxglossary[type=\acronymtype, sort=letter]
\begin{figure}[H]
    \centering
    \includegraphics[]{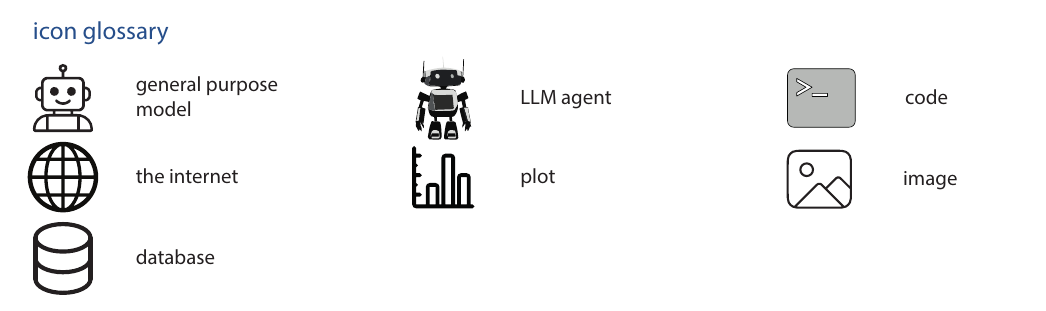}
    \caption{\textbf{Meaning of icons reused in our figures.}}
    \label{fig:icon-glossary}
\end{figure}
\printnoidxglossary[type=main, title=Glossary] \label{glossary}
\end{document}